\ifdefined\pdftexversion\pdfoutput=1\fi 
\documentclass[11pt]{article}

\newif\ifpreprint
\preprinttrue

\ifpreprint
  \usepackage[preprint]{acl}
\else
  \usepackage[review]{acl}
\fi

\usepackage{times}
\usepackage{latexsym}
\usepackage[T1]{fontenc}
\usepackage[utf8]{inputenc}
\usepackage{microtype}
\usepackage{inconsolata}
\usepackage{graphicx}
\usepackage{booktabs}
\usepackage{amsmath}
\usepackage{amssymb}
\usepackage{multirow}
\usepackage{siunitx}
\newcommand{\xdb}{XDecideBench}
\newcommand{\coretwelve}{Core12}
\newcommand{\fresh}{held-out cohort}
\newcommand{\Fresh}{Held-out cohort}
\newcommand{\choice}{\textsc{Choice}}
\newcommand{\noul}{\textsc{Noul}}
\newcommand{\score}{\textsc{Score}}
\newcommand{\lang}[1]{\texttt{#1}}

\newcommand{\Dhb}{\Delta_{\text{h}-\text{b}}}

\ifpreprint
  \newcommand{\Afterreview}{We}
  \newcommand{\afterreview}{}
  \newcommand{\releaseat}{ at \url{https://github.com/NaiveNeuron/XDecideBench}}
\else
  \newcommand{\Afterreview}{After review, we}
  \newcommand{\afterreview}{ after review}
  \newcommand{\releaseat}{}
\fi

\newcommand{\NumDecisionModels}{eleven}
\newcommand{\NumVendors}{ten}

\NewDocumentCommand{\minus}{}{\ensuremath{-}}
\NewDocumentCommand{\plus}{}{\ensuremath{+}}
\NewDocumentCommand{\tabminus}{}{\ifmmode-\else\textminus\fi}
\NewDocumentCommand{\tabplus}{}{\ifmmode+\else+\fi}
\AddToHook{env/tabular/begin}{\let\minus\tabminus\let\plus\tabplus}
\colorlet{tabgrey}{black!55} 
\newcommand{\tmark}[1]{\textcolor{tabgrey}{#1}}

\newcommand{\tci}[1]{\textcolor{tabgrey}{#1}}
\NewDocumentCommand{\tunc}{m}{\raisebox{-0.45ex}[0pt][0pt]{\fontsize{7}{7}\selectfont\textcolor{tabgrey}{\textpm#1}}}
\NewDocumentCommand{\tunder}{m}{\llap{\underline{\phantom{#1}}}}
\newcommand{\tgroup}[2]{\multicolumn{#1}{c}{\textit{#2}}}
\newcommand{\tabnote}[1]{\par\vspace{3pt}{\footnotesize\raggedright #1\par}}

\newsavebox{\fitwidthbox}
\newcommand{\fitwidth}[1]{\sbox{\fitwidthbox}{#1}%
  \ifdim\wd\fitwidthbox>\linewidth\resizebox{\linewidth}{!}{\usebox{\fitwidthbox}}\else\usebox{\fitwidthbox}\fi}

\newcommand{\NumHeadlineTracks}{nine} 
\newcommand{\NumTextsUnique}{88,425} 
\newcommand{\NumQuestionsUnique}{153,875}

\newcommand{\NumHeadlineLangs}{122}

\newcommand{\suiteMassiveVarieties}{52}

\newcommand{\suiteRtplxVarieties}{38}

\newcommand{\suiteInjongoVarieties}{16}

\newcommand{\suiteModerVarieties}{17}

\newcommand{\suiteDiagRtplxTexts}{400} 
\newcommand{\suiteDiagRtplxQuestions}{3,200} 
\newcommand{\NumCoreXnliTexts}{6,000} 
\newcommand{\coreItemsPerLangMassive}{500} 
\newcommand{\coreItemsPerLangXnli}{500} 
\newcommand{\coreItemsPerLangBelebele}{501} 
\newcommand{\coreItemsPerLangMmluProx}{500} 
\newcommand{\coreItemsPerLangSib}{204} 
\newcommand{\massiveFiftyTwoUtterances}{117} 
 
\newcommand{\skPplxCoreTwelve}{.712} 
\newcommand{\hwPplxCoreTwelve}{.019} 
\newcommand{\skPplxBelebele}{.756} 
\newcommand{\hwPplxBelebele}{.058} 
\newcommand{\skPplxMassive}{.718} 
\newcommand{\hwPplxMassive}{.075} 
\newcommand{\skPplxNli}{.397} 
\newcommand{\hwPplxNli}{.022} 
\newcommand{\skPplxRtplx}{.051} 
\newcommand{\hwPplxRtplx}{.041} 
\newcommand{\skPplxInclude}{.536} 
\newcommand{\hwPplxInclude}{.009} 
\newcommand{\skPplxInjongo}{.578} 
\newcommand{\hwPplxInjongo}{.026} 
\newcommand{\skPplxModer}{.673} 
\newcommand{\hwPplxModer}{.022} 
\newcommand{\skPplxBrighter}{.144} 
\newcommand{\hwPplxBrighter}{.020} 
\newcommand{\skJevCoreTwelve}{.688} 
\newcommand{\hwJevCoreTwelve}{.022} 
\newcommand{\skJevBelebele}{.785} 
\newcommand{\hwJevBelebele}{.060} 
\newcommand{\skJevMassive}{.619} 
\newcommand{\hwJevMassive}{.081} 
\newcommand{\skJevNli}{.128} 
\newcommand{\hwJevNli}{.030} 
\newcommand{\skJevRtplx}{\minus.101} 
\newcommand{\hwJevRtplx}{.051} 
\newcommand{\skJevInclude}{.531} 
\newcommand{\hwJevInclude}{.009} 
\newcommand{\skJevInjongo}{.499} 
\newcommand{\hwJevInjongo}{.024} 
\newcommand{\skJevModer}{.561} 
\newcommand{\hwJevModer}{.026} 
\newcommand{\skJevBrighter}{\minus.428} 
\newcommand{\hwJevBrighter}{.046} 
\newcommand{\skClefCoreTwelve}{.674} 
\newcommand{\hwClefCoreTwelve}{.020} 
\newcommand{\skClefBelebele}{.701} 
\newcommand{\hwClefBelebele}{.055} 
\newcommand{\skClefMassive}{.714} 
\newcommand{\hwClefMassive}{.069} 
\newcommand{\skClefNli}{.254} 
\newcommand{\hwClefNli}{.023} 
\newcommand{\skClefRtplx}{.045} 
\newcommand{\hwClefRtplx}{.046} 
\newcommand{\skClefInclude}{.459} 
\newcommand{\hwClefInclude}{.009} 
\newcommand{\skClefInjongo}{.492} 
\newcommand{\hwClefInjongo}{.026} 
\newcommand{\skClefModer}{.274} 
\newcommand{\hwClefModer}{.032} 
\newcommand{\skClefBrighter}{.167} 
\newcommand{\hwClefBrighter}{.016} 
\newcommand{\skLiquidCoreTwelve}{.663} 
\newcommand{\hwLiquidCoreTwelve}{.020} 
\newcommand{\skLiquidBelebele}{.711} 
\newcommand{\hwLiquidBelebele}{.061} 
\newcommand{\skLiquidMassive}{.634} 
\newcommand{\hwLiquidMassive}{.073} 
\newcommand{\skLiquidNli}{.315} 
\newcommand{\hwLiquidNli}{.023} 
\newcommand{\skLiquidRtplx}{.070} 
\newcommand{\hwLiquidRtplx}{.046} 
\newcommand{\skLiquidInclude}{.488} 
\newcommand{\hwLiquidInclude}{.009} 
\newcommand{\skLiquidInjongo}{.481} 
\newcommand{\hwLiquidInjongo}{.024} 
\newcommand{\skLiquidModer}{.449} 
\newcommand{\hwLiquidModer}{.027} 
\newcommand{\skLiquidBrighter}{.118} 
\newcommand{\hwLiquidBrighter}{.022} 
\newcommand{\skClefFlashCoreTwelve}{.631} 
\newcommand{\hwClefFlashCoreTwelve}{.020} 
\newcommand{\skClefFlashBelebele}{.597} 
\newcommand{\hwClefFlashBelebele}{.053} 
\newcommand{\skClefFlashMassive}{.745} 
\newcommand{\hwClefFlashMassive}{.061} 
\newcommand{\skClefFlashNli}{.127} 
\newcommand{\hwClefFlashNli}{.025} 
\newcommand{\skClefFlashRtplx}{.071} 
\newcommand{\hwClefFlashRtplx}{.047} 
\newcommand{\skClefFlashInclude}{.385} 
\newcommand{\hwClefFlashInclude}{.009} 
\newcommand{\skClefFlashInjongo}{.415} 
\newcommand{\hwClefFlashInjongo}{.026} 
\newcommand{\skClefFlashModer}{.379} 
\newcommand{\hwClefFlashModer}{.031} 
\newcommand{\skClefFlashBrighter}{.123} 
\newcommand{\hwClefFlashBrighter}{.016} 
\newcommand{\skMercuryCoreTwelve}{.611} 
\newcommand{\hwMercuryCoreTwelve}{.025} 
\newcommand{\skMercuryBelebele}{.677} 
\newcommand{\hwMercuryBelebele}{.076} 
\newcommand{\skMercuryMassive}{.651} 
\newcommand{\hwMercuryMassive}{.087} 
\newcommand{\skMercuryNli}{.188} 
\newcommand{\hwMercuryNli}{.030} 
\newcommand{\skMercuryRtplx}{\minus.022} 
\newcommand{\hwMercuryRtplx}{.052} 
\newcommand{\skMercuryInclude}{.411} 
\newcommand{\hwMercuryInclude}{.011} 
\newcommand{\skMercuryInjongo}{.416} 
\newcommand{\hwMercuryInjongo}{.031} 
\newcommand{\skMercuryModer}{.449} 
\newcommand{\hwMercuryModer}{.032} 
\newcommand{\skMercuryBrighter}{.013} 
\newcommand{\hwMercuryBrighter}{.027} 
\newcommand{\skLunaCoreTwelve}{.568} 
\newcommand{\hwLunaCoreTwelve}{.023} 
\newcommand{\skLunaBelebele}{.743} 
\newcommand{\hwLunaBelebele}{.073} 
\newcommand{\skLunaMassive}{.602} 
\newcommand{\hwLunaMassive}{.065} 
\newcommand{\skLunaNli}{\minus.058} 
\newcommand{\hwLunaNli}{.036} 
\newcommand{\skLunaRtplx}{\minus.018} 
\newcommand{\hwLunaRtplx}{.047} 
\newcommand{\skLunaInclude}{.600} 
\newcommand{\hwLunaInclude}{.009} 
\newcommand{\skLunaInjongo}{.580} 
\newcommand{\hwLunaInjongo}{.018} 
\newcommand{\skLunaModer}{.541} 
\newcommand{\hwLunaModer}{.031} 
\newcommand{\skLunaBrighter}{\minus.096} 
\newcommand{\hwLunaBrighter}{.031} 
\newcommand{\skKevCoreTwelve}{.449} 
\newcommand{\hwKevCoreTwelve}{.021} 
\newcommand{\skKevBelebele}{.389} 
\newcommand{\hwKevBelebele}{.091} 
\newcommand{\skKevMassive}{.371} 
\newcommand{\hwKevMassive}{.046} 
\newcommand{\skKevNli}{.070} 
\newcommand{\hwKevNli}{.030} 
\newcommand{\skKevRtplx}{\minus.169} 
\newcommand{\hwKevRtplx}{.051} 
\newcommand{\skKevInclude}{.273} 
\newcommand{\hwKevInclude}{.007} 
\newcommand{\skKevInjongo}{.225} 
\newcommand{\hwKevInjongo}{.014} 
\newcommand{\skKevModer}{.278} 
\newcommand{\hwKevModer}{.023} 
\newcommand{\skKevBrighter}{\minus.275} 
\newcommand{\hwKevBrighter}{.036} 
\newcommand{\skSolarCoreTwelve}{.398} 
\newcommand{\hwSolarCoreTwelve}{.024} 
\newcommand{\skSolarBelebele}{.334} 
\newcommand{\hwSolarBelebele}{.074}

\newcommand{\skSolarNli}{\minus.236} 
\newcommand{\hwSolarNli}{.029} 
\newcommand{\skSolarRtplx}{\minus.568} 
\newcommand{\hwSolarRtplx}{.078} 
\newcommand{\skSolarInclude}{.251} 
\newcommand{\hwSolarInclude}{.013}

\newcommand{\skSolarModer}{\minus.046} 
\newcommand{\hwSolarModer}{.058} 
\newcommand{\skSolarBrighter}{\minus1.396} 
\newcommand{\hwSolarBrighter}{.077} 
\newcommand{\skMsftCoreTwelve}{.651} 
\newcommand{\hwMsftCoreTwelve}{.018} 
\newcommand{\skMsftBelebele}{.716} 
\newcommand{\hwMsftBelebele}{.047} 
\newcommand{\skMsftMassive}{.721} 
\newcommand{\hwMsftMassive}{.059} 
\newcommand{\skMsftNli}{.108} 
\newcommand{\hwMsftNli}{.029} 
\newcommand{\skMsftRtplx}{\minus.229} 
\newcommand{\hwMsftRtplx}{.061} 
\newcommand{\skMsftInclude}{.400} 
\newcommand{\hwMsftInclude}{.011} 
\newcommand{\skMsftInjongo}{.417} 
\newcommand{\hwMsftInjongo}{.024} 
\newcommand{\skMsftModer}{.500} 
\newcommand{\hwMsftModer}{.029} 
\newcommand{\skMsftBrighter}{\minus.375} 
\newcommand{\hwMsftBrighter}{.040}

\newcommand{\skClefLocalCoreTwelve}{.674} 
\newcommand{\hwClefLocalCoreTwelve}{.020} 
\newcommand{\skClefFlashLocalCoreTwelve}{.631} 
\newcommand{\hwClefFlashLocalCoreTwelve}{.020} 
\newcommand{\skQwenTwentySevenHostedCoreTwelve}{.584} 
\newcommand{\hwQwenTwentySevenHostedCoreTwelve}{.024} 
\newcommand{\skQwenNineHostedCoreTwelve}{.477} 
\newcommand{\hwQwenNineHostedCoreTwelve}{.022} 
\newcommand{\skQwenTwentySevenLocalCoreTwelve}{.410} 
\newcommand{\hwQwenTwentySevenLocalCoreTwelve}{.021} 
\newcommand{\skQwenNineLocalCoreTwelve}{.403} 
\newcommand{\hwQwenNineLocalCoreTwelve}{.014}

\newcommand{\skHostedCoreTwelveLo}{.398} 
\newcommand{\skHostedCoreTwelveHi}{.712} 
\newcommand{\skQwenNineHostedMinusLocal}{.074} 
\newcommand{\skQwenTwentySevenHostedMinusLocal}{.174} 
\newcommand{\skMmbertThreeSets}{.673} 
\newcommand{\skMmbertThreeSetsCI}{[.639, .709]} 
\newcommand{\skLayaThreeSets}{.274} 
\newcommand{\skLayaThreeSetsCI}{[.234, .307]} 
\newcommand{\skLayaMassive}{.030} 
\newcommand{\nCompleteModels}{ten} 
\newcommand{\tauCoreVsRtplx}{.33} 
\newcommand{\tauCoreVsNli}{.64} 
\newcommand{\skQwenTwentySevenHostedBelebele}{.678} 
\newcommand{\hwQwenTwentySevenHostedBelebele}{.065} 
\newcommand{\skQwenTwentySevenHostedMassive}{.523} 
\newcommand{\hwQwenTwentySevenHostedMassive}{.088} 
\newcommand{\skQwenTwentySevenHostedNli}{.043} 
\newcommand{\hwQwenTwentySevenHostedNli}{.035} 
\newcommand{\skQwenTwentySevenHostedRtplx}{\minus.336} 
\newcommand{\hwQwenTwentySevenHostedRtplx}{.073} 
\newcommand{\skQwenTwentySevenHostedInclude}{.426} 
\newcommand{\hwQwenTwentySevenHostedInclude}{.010} 
\newcommand{\skQwenTwentySevenHostedInjongo}{.405} 
\newcommand{\hwQwenTwentySevenHostedInjongo}{.028} 
\newcommand{\skQwenTwentySevenHostedModer}{.397} 
\newcommand{\hwQwenTwentySevenHostedModer}{.031} 
\newcommand{\skQwenTwentySevenHostedBrighter}{\minus.056} 
\newcommand{\hwQwenTwentySevenHostedBrighter}{.028} 
\newcommand{\skQwenNineHostedBelebele}{.525} 
\newcommand{\hwQwenNineHostedBelebele}{.077} 
\newcommand{\skQwenNineHostedMassive}{.313} 
\newcommand{\hwQwenNineHostedMassive}{.091} 
\newcommand{\skQwenNineHostedNli}{\minus.036} 
\newcommand{\hwQwenNineHostedNli}{.029} 
\newcommand{\skQwenNineHostedRtplx}{\minus.562} 
\newcommand{\hwQwenNineHostedRtplx}{.086} 
\newcommand{\skQwenNineHostedInclude}{.310} 
\newcommand{\hwQwenNineHostedInclude}{.010} 
\newcommand{\skQwenNineHostedInjongo}{.167} 
\newcommand{\hwQwenNineHostedInjongo}{.029} 
\newcommand{\skQwenNineHostedModer}{.347} 
\newcommand{\hwQwenNineHostedModer}{.029} 
\newcommand{\skQwenNineHostedBrighter}{\minus.034} 
\newcommand{\hwQwenNineHostedBrighter}{.027} 
\newcommand{\skMediaPipeCoreTwelve}{\minus.084} 
\newcommand{\hwMediaPipeCoreTwelve}{.029} 
\newcommand{\skMediaPipeBelebele}{\minus.284} 
\newcommand{\hwMediaPipeBelebele}{.097} 
\newcommand{\skMediaPipeMassive}{.347} 
\newcommand{\hwMediaPipeMassive}{.058} 
\newcommand{\skMediaPipeNli}{\minus.997} 
\newcommand{\hwMediaPipeNli}{.039} 
\newcommand{\skMediaPipeRtplx}{\minus1.363} 
\newcommand{\hwMediaPipeRtplx}{.116} 
\newcommand{\skMediaPipeInclude}{\minus.242} 
\newcommand{\hwMediaPipeInclude}{.010} 
\newcommand{\skMediaPipeInjongo}{.054} 
\newcommand{\hwMediaPipeInjongo}{.010} 
\newcommand{\skMediaPipeModer}{\minus.004} 
\newcommand{\hwMediaPipeModer}{.021} 
\newcommand{\skMediaPipeBrighter}{\minus1.166} 
\newcommand{\hwMediaPipeBrighter}{.068} 
\newcommand{\skGlinerCoreTwelve}{.057} 
\newcommand{\hwGlinerCoreTwelve}{.015} 
\newcommand{\skGlinerBelebele}{\minus.066} 
\newcommand{\hwGlinerBelebele}{.048} 
\newcommand{\skGlinerMassive}{.029} 
\newcommand{\hwGlinerMassive}{.037} 
\newcommand{\skGlinerNli}{.010} 
\newcommand{\hwGlinerNli}{.007} 
\newcommand{\skGlinerRtplx}{\minus.984} 
\newcommand{\hwGlinerRtplx}{.081} 
\newcommand{\skGlinerInclude}{\minus.034} 
\newcommand{\hwGlinerInclude}{.005} 
\newcommand{\skGlinerInjongo}{.032} 
\newcommand{\hwGlinerInjongo}{.008} 
\newcommand{\skGlinerModer}{.076} 
\newcommand{\hwGlinerModer}{.021} 
\newcommand{\skGlinerBrighter}{\minus.807} 
\newcommand{\hwGlinerBrighter}{.044} 
\newcommand{\skLayaSuiteCoreTwelve}{.119} 
\newcommand{\hwLayaSuiteCoreTwelve}{.022} 
 
\newcommand{\skLayaSuiteBelebele}{\minus.167} 
\newcommand{\hwLayaSuiteBelebele}{.057} 
 
\newcommand{\skLayaSuiteMassive}{\minus.120} 
\newcommand{\hwLayaSuiteMassive}{.069} 
\newcommand{\skLayaSuiteNli}{\minus.206} 
\newcommand{\hwLayaSuiteNli}{.033} 
\newcommand{\skLayaSuiteRtplx}{\minus1.647} 
\newcommand{\hwLayaSuiteRtplx}{.118} 
\newcommand{\skLayaSuiteInclude}{\minus.085} 
\newcommand{\hwLayaSuiteInclude}{.006} 
 
\newcommand{\skLayaSuiteInjongo}{\minus.096} 
\newcommand{\hwLayaSuiteInjongo}{.014} 
\newcommand{\skLayaSuiteModer}{\minus.415} 
\newcommand{\hwLayaSuiteModer}{.057} 
\newcommand{\skLayaSuiteBrighter}{\minus2.907} 
\newcommand{\hwLayaSuiteBrighter}{.128}

\newcommand{\skClefLocalBelebele}{.701} 
\newcommand{\hwClefLocalBelebele}{.055} 
\newcommand{\skClefFlashLocalBelebele}{.597} 
\newcommand{\hwClefFlashLocalBelebele}{.053} 
\newcommand{\skQwenTwentySevenLocalBelebele}{.409} 
\newcommand{\hwQwenTwentySevenLocalBelebele}{.046} 
\newcommand{\skQwenNineLocalBelebele}{.308} 
\newcommand{\hwQwenNineLocalBelebele}{.040} 
\newcommand{\solarCappedMassive}{52} 
\newcommand{\solarCappedInjongo}{16}

\newcommand{\glinerCappedBelebele}{3} 
\newcommand{\tauAccuracyVsBrierHosted}{.96} 
\newcommand{\nHostedDecisionModels}{ten} 
\newcommand{\tauRpsVsBrierRtplx}{.64} 
\newcommand{\tauRpsVsBrierBrighter}{.87} 
\newcommand{\ordRtplxSwappedPairs}{8} 
\newcommand{\ordRtplxPairs}{45} 
\newcommand{\ordModels}{ten} 
\newcommand{\aoneLocalClefAccInteraction}{2.2} 
\newcommand{\aoneLocalClefBrierGainInteraction}{.013} 
\newcommand{\aoneLocalClefFlashAccInteraction}{4.3} 
\newcommand{\aoneLocalClefFlashBrierGainInteraction}{\minus.011} 
\newcommand{\aoneLocalClefFlashBrierGainInteractionCI}{[\minus.022, \minus.000]} 
\newcommand{\aoneLocalPplxAccInteraction}{4.7} 
\newcommand{\aoneLocalPplxBrierGainInteraction}{.042} 
\newcommand{\aoneHostedClefAccInteraction}{1.6}

\newcommand{\aoneHostedClefFlashAccInteraction}{2.0}

\newcommand{\aoneHostedPplxAccInteraction}{4.1}

\newcommand{\aoneLocalAccGainAllLo}{7.6} 
\newcommand{\aoneLocalAccGainAllHi}{18.2} 
\newcommand{\aoneLocalBrierGainAllLo}{.15} 
\newcommand{\aoneLocalBrierGainAllHi}{.29} 
\newcommand{\aoneLocalEceGainAllLo}{.06} 
\newcommand{\aoneLocalEceGainAllHi}{.19} 
\newcommand{\esLocalClefLow}{.33} 
\newcommand{\esLocalClefHigh}{.36} 
\newcommand{\esLocalClefFlashLow}{.30} 
\newcommand{\esLocalClefFlashHigh}{.31} 
\newcommand{\esLocalPplxLow}{.44} 
\newcommand{\esLocalPplxHigh}{.42} 
\newcommand{\esLocalClefDiff}{\minus.032} 
\newcommand{\esLocalClefDiffCI}{[\minus.061, \minus.003]} 
\newcommand{\esHostedClefDiff}{\minus.008} 
\newcommand{\esHostedClefFlashDiff}{\minus.065} 
\newcommand{\esHostedClefFlashDiffCI}{[\minus.096, \minus.034]} 
\newcommand{\esHostedPplxDiff}{\plus.052} 
\newcommand{\esHostedPplxDiffCI}{[.017, .090]} 
\newcommand{\webShareHighMin}{4.2} 
\newcommand{\webShareLowMax}{0.36} 
\newcommand{\bbLanguages}{106} 
\newcommand{\bbAllBeat}{88} 
\newcommand{\bbAllBeatCI}{twelve} 
\newcommand{\bbWebN}{91} 
\newcommand{\bbUnmapped}{five} 
\newcommand{\bbMerged}{ten} 
\newcommand{\bbWebNSens}{101} 
\newcommand{\bbPosClef}{95} 
\newcommand{\bbCiPosClef}{28} 
\newcommand{\bbAccMedianClef}{4.0} 
\newcommand{\bbEsMedianClef}{.26} 
\newcommand{\bbBrierCiNegClef}{102} 
\newcommand{\bbBrierCiPosClef}{zero} 
\newcommand{\bbRhoClef}{\minus.41} 
\newcommand{\bbRhoSensClef}{\minus.39} 
\newcommand{\bbPClef}{$p<.001$} 
\newcommand{\bbSlopeClef}{\minus1.0} 
\newcommand{\bbBrierRhoClef}{\minus.03} 
\newcommand{\bbBrierPClef}{$p=.774$} 
\newcommand{\bbPosClefFlash}{98} 
\newcommand{\bbCiPosClefFlash}{44} 
\newcommand{\bbAccMedianClefFlash}{7.0} 
\newcommand{\bbEsMedianClefFlash}{.31} 
\newcommand{\bbBrierCiNegClefFlash}{94} 
\newcommand{\bbBrierCiPosClefFlash}{one} 
\newcommand{\bbRhoClefFlash}{\plus.02} 
\newcommand{\bbRhoSensClefFlash}{\plus.03} 
\newcommand{\bbPClefFlash}{$p=.824$} 
\newcommand{\bbSlopeClefFlash}{\plus0.3} 
\newcommand{\bbBrierRhoClefFlash}{\minus.53} 
\newcommand{\bbBrierPClefFlash}{$p<.001$} 
\newcommand{\bbPosPplx}{103} 
\newcommand{\bbCiPosPplx}{50} 
\newcommand{\bbAccMedianPplx}{5.5} 
\newcommand{\bbEsMedianPplx}{.40} 
\newcommand{\bbBrierCiNegPplx}{100} 
\newcommand{\bbBrierCiPosPplx}{one} 
\newcommand{\bbRhoPplx}{\minus.54} 
\newcommand{\bbRhoSensPplx}{\minus.54} 
\newcommand{\bbPPplx}{$p<.001$} 
\newcommand{\bbSlopePplx}{\minus2.4} 
\newcommand{\bbBrierRhoPplx}{\plus.14} 
\newcommand{\bbBrierPPplx}{$p=.172$}

\newcommand{\tempCells}{36} 
\newcommand{\tempEceLower}{36} 
 
\newcommand{\tempEceExcl}{29} 
\newcommand{\aThreeDevMassive}{500} 
\newcommand{\aThreeDevXnli}{500} 
\newcommand{\aThreeThresholdN}{250} 
\newcommand{\aThreeXnliEnRiskLo}{2.3} 
\newcommand{\aThreeXnliEnRiskHi}{5.1} 
\newcommand{\aThreeMassiveEnRiskLo}{1.7} 
\newcommand{\aThreeMassiveEnRiskHi}{6.6} 
\newcommand{\aThreeSolarXnliEnAdmitted}{3} 
\newcommand{\aThreeXnliDeFrRiseLo}{3.0} 
\newcommand{\aThreeXnliDeFrRiseHi}{8.0} 
\newcommand{\aThreeXnliDeFrRiskLo}{7.8} 
\newcommand{\aThreeXnliDeFrRiskHi}{12.4} 
\newcommand{\aThreeXnliKevLowRisk}{22.4} 
\newcommand{\aThreeXnliEnCovLo}{.90} 
\newcommand{\aThreeXnliEnCovHi}{.92} 
\newcommand{\aThreeXnliHighCovLo}{.80} 
\newcommand{\aThreeXnliHighCovHi}{.83} 
\newcommand{\aThreeXnliLowCovLo}{.65} 
\newcommand{\aThreeXnliLowCovHi}{.76} 
\newcommand{\aThreeMassiveLiquidFrRise}{2.2} 
\newcommand{\aThreeMassiveLiquidFrRiseCI}{[0.3, 4.4]}

\newcommand{\aThreeMassiveExcessSmallMax}{1.2} 
\newcommand{\aThreeMassiveExcessLuna}{2.6} 
\newcommand{\aThreeMassiveExcessLiquid}{5.6} 
\newcommand{\aThreeMassiveExcessKev}{9.9} 
\newcommand{\aThreeMassiveLowAdmDropLuna}{7.3} 
\newcommand{\aThreeMassiveLowAdmDropKev}{25.1} 
\newcommand{\aThreeMassiveJevSwAdmitted}{10.8} 
\newcommand{\aThreeMassiveJevEnAdmitted}{46.8} 
\newcommand{\aThreeFloorFourRiskMax}{2.2} 
\newcommand{\aThreeFloorFourCovMax}{2.6} 
\newcommand{\aThreeFloorEightRiskMax}{0.6} 
\newcommand{\aThreeFloorEightCovMax}{4.6} 
\newcommand{\aThreeRoundedRiskMax}{0.6} 
\newcommand{\aThreeRoundedCovMax}{1.2}

\newcommand{\aThreeXnliEmptySetOtherMax}{13} 
\newcommand{\aThreeXnliEmptySetEnMax}{4} 
\newcommand{\retestCases}{8,448} 
\newcommand{\retestTopOneLo}{96.5} 
\newcommand{\retestTopOneHi}{100} 
\newcommand{\retestMaxBrierChange}{.0024} 
\newcommand{\retestSentinelMaxDp}{.032} 
\newcommand{\hostedLocalTopOneClef}{99.1} 
\newcommand{\hostedLocalTopOneClefFlash}{99.5} 
\newcommand{\retestTopOneMsft}{98.9} 
\newcommand{\retestIdenticalMsft}{10} 
\newcommand{\retestBrierMsft}{.0002} 
\newcommand{\msftRepeatDp}{.03} 
\newcommand{\hTwoMargin}{.08} 
\newcommand{\hTwoMarginPointsLo}{2.8} 
\newcommand{\hTwoMarginPointsHi}{3.3} 
\newcommand{\freshItemsPerLang}{405} 
\newcommand{\freshParentGroups}{282} 
\newcommand{\hOnePower}{98} 
\newcommand{\hTwoPower}{70} 
\newcommand{\hTwoMarginEightyPower}{.087} 
\newcommand{\hTwoAlphaPerSide}{.05} 
\newcommand{\hTwoStandardError}{.030} 
\newcommand{\hTwoPowerAtCoreEstimate}{62} 
\newcommand{\hTwoPowerHolmFirst}{53} 
\newcommand{\hOnePowerHolm}{95} 
\newcommand{\hTwoPowerPerPairLo}{23} 
\newcommand{\hTwoPowerPerPairHi}{35} 
\newcommand{\hTwoBackboneErrorTwentySeven}{.354} 
\newcommand{\hTwoMarginPointsTwentySeven}{2.8} 
\newcommand{\hTwoBackboneErrorNine}{.410} 
\newcommand{\hTwoMarginPointsNine}{3.3} 
\newcommand{\regEnUnits}{232} 
\newcommand{\regEnWrong}{109} 
\newcommand{\regEnWrongPaired}{56} 
\newcommand{\regEnWrongSep}{22} 
\newcommand{\regEnGeTen}{47} 
\newcommand{\regEnThreshold}{.10} 
 
\newcommand{\regWorstPick}{.336} 
\newcommand{\regWorstBest}{.814} 
\newcommand{\regCoreOtherSuites}{eight} 
\newcommand{\regCoreRightSuites}{two} 
\newcommand{\regFreshUnits}{eleven} 
\newcommand{\regBbUnits}{105} 
\newcommand{\regBbPaired}{37} 
\newcommand{\regBbShownPaired}{35} 
\newcommand{\freshAlpha}{.05} 
\newcommand{\freshHolmFirstLevel}{.025} 
\newcommand{\freshHolmSecondLevel}{.05} 
\newcommand{\freshClusters}{282} 
\newcommand{\freshReps}{10,000} 
\newcommand{\freshDOne}{4.8} 
\newcommand{\freshDOneBound}{2.8} 
\newcommand{\freshHOneP}{.0001} 
\newcommand{\freshDTwo}{\minus.034} 
\newcommand{\freshDTwoCI}{[\minus.073, .006]} 
\newcommand{\freshHTwoP}{.0252} 
\newcommand{\freshHostedDOne}{2.7} 
\newcommand{\freshHostedDOneBound}{0.9} 
\newcommand{\freshHostedHOneP}{.0022} 
\newcommand{\freshHostedDTwo}{\minus.039} 
\newcommand{\freshHostedDTwoCI}{[\minus.084, .006]} 
\newcommand{\freshHostedHTwoP}{.0655} 
\newcommand{\freshDTwoLevel}{90} 
\newcommand{\freshDOneClef}{4.8} 
\newcommand{\freshDOneClefCI}{[2.3, 7.3]} 
\newcommand{\freshDTwoClef}{\minus.006} 
\newcommand{\freshDTwoClefCI}{[\minus.069, .060]} 
\newcommand{\freshBrierGainInteractionClef}{\plus.010} 
\newcommand{\freshBrierGainInteractionClefCI}{[\minus.013, .032]} 
\newcommand{\freshEceGainInteractionClef}{\minus.045} 
\newcommand{\freshEceGainInteractionClefCI}{[\minus.071, \minus.016]} 
\newcommand{\freshWebRhoClef}{\minus.77} 
\newcommand{\freshWebPClef}{.005} 
\newcommand{\freshDOneClefFlash}{4.7} 
\newcommand{\freshDOneClefFlashCI}{[1.7, 7.7]} 
\newcommand{\freshDTwoClefFlash}{\minus.045} 
\newcommand{\freshDTwoClefFlashCI}{[\minus.117, .028]} 
\newcommand{\freshBrierGainInteractionClefFlash}{\minus.028} 
\newcommand{\freshBrierGainInteractionClefFlashCI}{[\minus.057, .000]} 
\newcommand{\freshEceGainInteractionClefFlash}{\minus.106} 
\newcommand{\freshEceGainInteractionClefFlashCI}{[\minus.132, \minus.070]} 
\newcommand{\freshWebRhoClefFlash}{\minus.68} 
\newcommand{\freshWebPClefFlash}{.018} 
\newcommand{\freshDOnePplx}{4.9} 
\newcommand{\freshDOnePplxCI}{[2.0, 7.8]} 
\newcommand{\freshDTwoPplx}{\minus.050} 
\newcommand{\freshDTwoPplxCI}{[\minus.114, .016]} 
\newcommand{\freshBrierGainInteractionPplx}{\plus.039} 
\newcommand{\freshBrierGainInteractionPplxCI}{[.016, .063]} 
\newcommand{\freshEceGainInteractionPplx}{\minus.025} 
\newcommand{\freshEceGainInteractionPplxCI}{[\minus.057, \minus.001]} 
\newcommand{\freshWebRhoPplx}{\minus.51} 
\newcommand{\freshWebPPplx}{.093}

\newcommand{\nFreshRanked}{ten} 
\newcommand{\tauFreshVsCore}{.91} 
 
\newcommand{\skFreshLo}{\minus.214} 
\newcommand{\skFreshHi}{.730} 
\newcommand{\skMediaPipeFresh}{\minus.214} 
\newcommand{\skGlinerFresh}{.010} 
\newcommand{\covGlinerFresh}{91} 
\newcommand{\freshAThreeXnliDeFrRiskLo}{8.4} 
\newcommand{\freshAThreeXnliDeFrRiskHi}{13.2} 
\newcommand{\freshAThreeXnliEnRiskLo}{1.3} 
\newcommand{\freshAThreeXnliEnRiskHi}{3.6} 
\newcommand{\freshAThreeXnliLowCovLo}{.65} 
\newcommand{\freshAThreeXnliLowCovHi}{.75} 
\newcommand{\freshCases}{4,860} 
\newcommand{\rtplxToxDiffer}{4,016} 
\newcommand{\rtplxPairs}{7,400} 
\newcommand{\rtplxToxDifferPct}{54} 
\newcommand{\rtplxAllDifferPct}{22} 
\newcommand{\sibLinkSeRatioLo}{0.96} 
\newcommand{\sibLinkSeRatioHi}{1.02} 
\newcommand{\sibLinkModels}{twelve} 
\newcommand{\natLangs}{nine} 
\newcommand{\natLunaNativeOverTranslated}{.210} 
\newcommand{\natJevTranslatedOverNative}{.082} 
\newcommand{\descMassiveUpModels}{four} 
\newcommand{\descMassiveUpLo}{.035} 
\newcommand{\descMassiveUpHi}{.053} 
\newcommand{\layaMassiveAccuracy}{.320} 
\newcommand{\srLocalDOne}{6.8} 
\newcommand{\srLocalDOneCI}{[5.7, 8.0]} 
\newcommand{\srLocalDTwo}{\plus.037} 
\newcommand{\srLocalDTwoCI}{[\plus.006, \plus.067]} 
\newcommand{\srLocalRho}{\minus.83} 
\newcommand{\srHostedDOne}{4.5} 
\newcommand{\srHostedDOneCI}{[3.4, 5.6]} 
\newcommand{\srHostedDTwo}{\minus.013} 
\newcommand{\srHostedDTwoCI}{[\minus.043, \plus.017]} 
 
\newcommand{\srFreshLocalDOne}{7.1} 
\newcommand{\srFreshLocalDOneCI}{[4.2, 10.1]} 
\newcommand{\srFreshLocalDTwo}{\minus.001} 
\newcommand{\srFreshLocalDTwoCI}{[\minus.063, \plus.065]} 
 
\newcommand{\srFreshHostedDOne}{5.0} 
\newcommand{\srFreshHostedDOneCI}{[2.3, 7.8]} 
\newcommand{\srFreshHostedDTwo}{\minus.025} 
\newcommand{\srFreshHostedDTwoCI}{[\minus.090, \plus.044]} 
 
\newcommand{\srBbLocalAbove}{90} 
\newcommand{\srBbLocalMedian}{11.0} 
\newcommand{\srBbHostedAbove}{72} 
 
\newcommand{\srBbLanguages}{106} 
\newcommand{\niLangs}{eight} 
\newcommand{\niXnli}{.022} 
\newcommand{\niXnliCI}{[.017, .026]} 
\newcommand{\niXnliLow}{.039} 
\newcommand{\niXnliHigh}{.002} 
\newcommand{\niXnliLowHigh}{.037} 
\newcommand{\niXnliLowHighCI}{[.029, .046]} 
\newcommand{\niMassive}{\minus.003} 
\newcommand{\niMassiveCI}{[\minus.010, \plus.003]} 
\newcommand{\niMassiveFull}{.055} 
 
\newcommand{\niMassiveIds}{.018} 
\newcommand{\niMassiveIdsCI}{[.014, .022]} 
\newcommand{\niFullModels}{six} 
\newcommand{\niHeadShrink}{.099} 
\newcommand{\niHeadShrinkCI}{[.067, .129]} 
\newcommand{\niHeadShrinkMassive}{.039} 
\newcommand{\niPickWrong}{five} 
\newcommand{\niBestChanges}{three} 
\newcommand{\niRegEn}{.014} 
\newcommand{\niRegMaxEn}{.081} 
\newcommand{\niRegNative}{.023} 
\newcommand{\niRegMaxNative}{.106} 
\newcommand{\niTauMassive}{.78} 
\newcommand{\niTauXnli}{.87} 
\newcommand{\niJevXnliNat}{\plus.009$^{*}$} 
\newcommand{\niJevXnliFull}{\plus.025$^{*}$} 
\newcommand{\niJevMassiveNat}{\minus.001} 
\newcommand{\niJevMassiveFull}{\minus.027$^{*}$} 
\newcommand{\niPplxXnliNat}{\minus.002} 
\newcommand{\niPplxXnliFull}{\plus.006} 
\newcommand{\niPplxMassiveNat}{\plus.001} 
\newcommand{\niPplxMassiveFull}{\minus.024$^{*}$} 
\newcommand{\niLiquidXnliNat}{\minus.001} 
\newcommand{\niLiquidXnliFull}{\minus.001} 
\newcommand{\niLiquidMassiveNat}{\minus.016$^{*}$} 
\newcommand{\niLiquidMassiveFull}{\minus.016$^{*}$} 
\newcommand{\niLunaXnliNat}{\plus.101$^{*}$} 
\newcommand{\niLunaXnliFull}{\plus.065$^{*}$} 
\newcommand{\niLunaMassiveNat}{\minus.000} 
\newcommand{\niLunaMassiveFull}{\minus.171$^{*}$} 
\newcommand{\niMercuryXnliNat}{\minus.005} 
\newcommand{\niMercuryXnliFull}{\minus.012$^{*}$} 
\newcommand{\niMercuryMassiveNat}{\minus.001} 
\newcommand{\niMercuryMassiveFull}{\minus.041$^{*}$} 
\newcommand{\niKevXnliNat}{\minus.015$^{*}$} 
\newcommand{\niKevMassiveNat}{\plus.023$^{*}$} 
\newcommand{\niSolarXnliNat}{\minus.059$^{*}$} 
\newcommand{\niClefXnliNat}{\minus.078$^{*}$} 
\newcommand{\niClefMassiveNat}{\minus.010$^{*}$} 
\newcommand{\niClefFlashXnliNat}{\minus.119$^{*}$} 
\newcommand{\niClefFlashXnliFull}{\minus.088$^{*}$} 
\newcommand{\niClefFlashMassiveNat}{\minus.021$^{*}$} 
\newcommand{\niClefFlashMassiveFull}{\minus.049$^{*}$} 
\newcommand{\niMsftXnliNat}{\minus.046$^{*}$} 
\newcommand{\niMsftMassiveNat}{\minus.004} 
\newcommand{\niQwenNineXnliNat}{\minus.021} 
\newcommand{\niQwenNineMassiveNat}{\plus.018} 
\newcommand{\atJevXnliEnRisk}{.035} 
\newcommand{\atJevXnliEnAdm}{63} 
\newcommand{\atJevXnliEnCov}{.92} 
\newcommand{\atJevXnliEnSize}{1.06} 
\newcommand{\atJevXnliHighRisk}{.098} 
\newcommand{\atJevXnliHighAdm}{55} 
\newcommand{\atJevXnliHighCov}{.81} 
\newcommand{\atJevXnliHighSize}{1.07} 
\newcommand{\atJevXnliLowRisk}{.099} 
\newcommand{\atJevXnliLowAdm}{37} 
\newcommand{\atJevXnliLowCov}{.73} 
\newcommand{\atJevXnliLowSize}{1.03} 
\newcommand{\atPplxXnliEnRisk}{.045} 
\newcommand{\atPplxXnliEnAdm}{93} 
\newcommand{\atPplxXnliEnCov}{.91} 
\newcommand{\atPplxXnliEnSize}{0.96} 
\newcommand{\atPplxXnliHighRisk}{.104} 
\newcommand{\atPplxXnliHighAdm}{89} 
\newcommand{\atPplxXnliHighCov}{.83} 
\newcommand{\atPplxXnliHighSize}{0.95} 
\newcommand{\atPplxXnliLowRisk}{.133} 
\newcommand{\atPplxXnliLowAdm}{82} 
\newcommand{\atPplxXnliLowCov}{.75} 
\newcommand{\atPplxXnliLowSize}{0.90} 
\newcommand{\atLiquidXnliEnRisk}{.047} 
\newcommand{\atLiquidXnliEnAdm}{89} 
\newcommand{\atLiquidXnliEnCov}{.92} 
\newcommand{\atLiquidXnliEnSize}{0.99} 
\newcommand{\atLiquidXnliHighRisk}{.100} 
\newcommand{\atLiquidXnliHighAdm}{81} 
\newcommand{\atLiquidXnliHighCov}{.82} 
\newcommand{\atLiquidXnliHighSize}{0.97} 
\newcommand{\atLiquidXnliLowRisk}{.127} 
\newcommand{\atLiquidXnliLowAdm}{67} 
\newcommand{\atLiquidXnliLowCov}{.74} 
\newcommand{\atLiquidXnliLowSize}{0.93} 
\newcommand{\atMercuryXnliEnRisk}{.049} 
\newcommand{\atMercuryXnliEnAdm}{85} 
\newcommand{\atMercuryXnliEnCov}{.91} 
\newcommand{\atMercuryXnliEnSize}{1.02} 
\newcommand{\atMercuryXnliHighRisk}{.096} 
\newcommand{\atMercuryXnliHighAdm}{74} 
\newcommand{\atMercuryXnliHighCov}{.82} 
\newcommand{\atMercuryXnliHighSize}{1.03} 
\newcommand{\atMercuryXnliLowRisk}{.105} 
\newcommand{\atMercuryXnliLowAdm}{58} 
\newcommand{\atMercuryXnliLowCov}{.76} 
\newcommand{\atMercuryXnliLowSize}{1.04} 
\newcommand{\atKevXnliEnRisk}{.051} 
\newcommand{\atKevXnliEnAdm}{78} 
\newcommand{\atKevXnliEnCov}{.90} 
\newcommand{\atKevXnliEnSize}{1.04} 
\newcommand{\atKevXnliHighRisk}{.126} 
\newcommand{\atKevXnliHighAdm}{72} 
\newcommand{\atKevXnliHighCov}{.80} 
\newcommand{\atKevXnliHighSize}{1.04} 
\newcommand{\atKevXnliLowRisk}{.224} 
\newcommand{\atKevXnliLowAdm}{52} 
\newcommand{\atKevXnliLowCov}{.65} 
\newcommand{\atKevXnliLowSize}{1.07} 
\newcommand{\atLunaXnliEnAdm}{0} 
\newcommand{\atLunaXnliEnCov}{.93} 
\newcommand{\atLunaXnliEnSize}{1.46} 
\newcommand{\atLunaXnliHighAdm}{0} 
\newcommand{\atLunaXnliHighCov}{.86} 
\newcommand{\atLunaXnliHighSize}{1.52} 
\newcommand{\atLunaXnliLowAdm}{0} 
\newcommand{\atLunaXnliLowCov}{.85} 
\newcommand{\atLunaXnliLowSize}{1.67} 
\newcommand{\atSolarXnliEnRisk}{.143} 
\newcommand{\atSolarXnliEnAdm}{3} 
\newcommand{\atSolarXnliEnCov}{.91} 
\newcommand{\atSolarXnliEnSize}{1.38} 
\newcommand{\atSolarXnliHighRisk}{.225} 
\newcommand{\atSolarXnliHighAdm}{3} 
\newcommand{\atSolarXnliHighCov}{.83} 
\newcommand{\atSolarXnliHighSize}{1.44} 
\newcommand{\atSolarXnliLowRisk}{.322} 
\newcommand{\atSolarXnliLowAdm}{3} 
\newcommand{\atSolarXnliLowCov}{.73} 
\newcommand{\atSolarXnliLowSize}{1.49} 
\newcommand{\atMsftXnliEnRisk}{.023} 
\newcommand{\atMsftXnliEnAdm}{70} 
\newcommand{\atMsftXnliEnCov}{.92} 
\newcommand{\atMsftXnliEnSize}{1.04} 
\newcommand{\atMsftXnliHighRisk}{.110} 
\newcommand{\atMsftXnliHighAdm}{64} 
\newcommand{\atMsftXnliHighCov}{.81} 
\newcommand{\atMsftXnliHighSize}{1.06} 
\newcommand{\atMsftXnliLowRisk}{.174} 
\newcommand{\atMsftXnliLowAdm}{41} 
\newcommand{\atMsftXnliLowCov}{.70} 
\newcommand{\atMsftXnliLowSize}{1.09} 
\newcommand{\atJevMassiveEnRisk}{.030} 
\newcommand{\atJevMassiveEnAdm}{47} 
\newcommand{\atJevMassiveEnCov}{.94} 
\newcommand{\atJevMassiveEnSize}{1.86} 
\newcommand{\atJevMassiveHighRisk}{.029} 
\newcommand{\atJevMassiveHighAdm}{39} 
\newcommand{\atJevMassiveHighCov}{.93} 
\newcommand{\atJevMassiveHighSize}{2.00} 
\newcommand{\atJevMassiveLowRisk}{.022} 
\newcommand{\atJevMassiveLowAdm}{29} 
\newcommand{\atJevMassiveLowCov}{.88} 
\newcommand{\atJevMassiveLowSize}{2.38} 
\newcommand{\atPplxMassiveEnRisk}{.017} 
\newcommand{\atPplxMassiveEnAdm}{59} 
\newcommand{\atPplxMassiveEnCov}{.93} 
\newcommand{\atPplxMassiveEnSize}{1.20} 
\newcommand{\atPplxMassiveHighRisk}{.019} 
\newcommand{\atPplxMassiveHighAdm}{54} 
\newcommand{\atPplxMassiveHighCov}{.92} 
\newcommand{\atPplxMassiveHighSize}{1.23} 
\newcommand{\atPplxMassiveLowRisk}{.023} 
\newcommand{\atPplxMassiveLowAdm}{43} 
\newcommand{\atPplxMassiveLowCov}{.87} 
\newcommand{\atPplxMassiveLowSize}{1.27} 
\newcommand{\atLiquidMassiveEnRisk}{.041} 
\newcommand{\atLiquidMassiveEnAdm}{68} 
\newcommand{\atLiquidMassiveEnCov}{.94} 
\newcommand{\atLiquidMassiveEnSize}{1.40} 
\newcommand{\atLiquidMassiveHighRisk}{.050} 
\newcommand{\atLiquidMassiveHighAdm}{64} 
\newcommand{\atLiquidMassiveHighCov}{.93} 
\newcommand{\atLiquidMassiveHighSize}{1.51} 
\newcommand{\atLiquidMassiveLowRisk}{.060} 
\newcommand{\atLiquidMassiveLowAdm}{50} 
\newcommand{\atLiquidMassiveLowCov}{.87} 
\newcommand{\atLiquidMassiveLowSize}{1.72} 
\newcommand{\atMercuryMassiveEnRisk}{.040} 
\newcommand{\atMercuryMassiveEnAdm}{60} 
\newcommand{\atMercuryMassiveEnCov}{.95} 
\newcommand{\atMercuryMassiveEnSize}{1.94} 
\newcommand{\atMercuryMassiveHighRisk}{.034} 
\newcommand{\atMercuryMassiveHighAdm}{57} 
\newcommand{\atMercuryMassiveHighCov}{.93} 
\newcommand{\atMercuryMassiveHighSize}{1.99} 
\newcommand{\atMercuryMassiveLowRisk}{.044} 
\newcommand{\atMercuryMassiveLowAdm}{46} 
\newcommand{\atMercuryMassiveLowCov}{.90} 
\newcommand{\atMercuryMassiveLowSize}{2.76} 
\newcommand{\atKevMassiveEnRisk}{.066} 
\newcommand{\atKevMassiveEnAdm}{49} 
\newcommand{\atKevMassiveEnCov}{.94} 
\newcommand{\atKevMassiveEnSize}{2.51} 
\newcommand{\atKevMassiveHighRisk}{.080} 
\newcommand{\atKevMassiveHighAdm}{43} 
\newcommand{\atKevMassiveHighCov}{.91} 
\newcommand{\atKevMassiveHighSize}{2.76} 
\newcommand{\atKevMassiveLowRisk}{.113} 
\newcommand{\atKevMassiveLowAdm}{23} 
\newcommand{\atKevMassiveLowCov}{.80} 
\newcommand{\atKevMassiveLowSize}{3.76} 
\newcommand{\atLunaMassiveEnRisk}{.046} 
\newcommand{\atLunaMassiveEnAdm}{48} 
\newcommand{\atLunaMassiveEnCov}{.90} 
\newcommand{\atLunaMassiveEnSize}{2.07} 
\newcommand{\atLunaMassiveHighRisk}{.060} 
\newcommand{\atLunaMassiveHighAdm}{47} 
\newcommand{\atLunaMassiveHighCov}{.89} 
\newcommand{\atLunaMassiveHighSize}{2.12} 
\newcommand{\atLunaMassiveLowRisk}{.066} 
\newcommand{\atLunaMassiveLowAdm}{40} 
\newcommand{\atLunaMassiveLowCov}{.86} 
\newcommand{\atLunaMassiveLowSize}{2.41} 
\newcommand{\atMsftMassiveEnRisk}{.037} 
\newcommand{\atMsftMassiveEnAdm}{81} 
\newcommand{\atMsftMassiveEnCov}{.94} 
\newcommand{\atMsftMassiveEnSize}{1.16} 
\newcommand{\atMsftMassiveHighRisk}{.045} 
\newcommand{\atMsftMassiveHighAdm}{78} 
\newcommand{\atMsftMassiveHighCov}{.91} 
\newcommand{\atMsftMassiveHighSize}{1.16} 
\newcommand{\atMsftMassiveLowRisk}{.052} 
\newcommand{\atMsftMassiveLowAdm}{62} 
\newcommand{\atMsftMassiveLowCov}{.86} 
\newcommand{\atMsftMassiveLowSize}{1.23} 
\newcommand{\ctPplxCI}{[.693, .731]} 
\newcommand{\ctPplxWorstLang}{sw} 
\newcommand{\ctPplxWorst}{.62} 
\newcommand{\ctPplxGap}{\minus.075} 
\newcommand{\ctPplxCov}{1.000} 
\newcommand{\ctJevCI}{[.667, .710]} 
\newcommand{\ctJevWorstLang}{sw} 
\newcommand{\ctJevWorst}{.58} 
\newcommand{\ctJevGap}{\minus.096} 
\newcommand{\ctJevCov}{1.000} 
\newcommand{\ctClefCI}{[.655, .695]} 
\newcommand{\ctClefWorstLang}{sw} 
\newcommand{\ctClefWorst}{.56} 
\newcommand{\ctClefGap}{\minus.103} 
\newcommand{\ctClefCov}{1.000} 
\newcommand{\ctLiquidCI}{[.642, .682]} 
\newcommand{\ctLiquidWorstLang}{sw} 
\newcommand{\ctLiquidWorst}{.55} 
\newcommand{\ctLiquidGap}{\minus.102} 
\newcommand{\ctLiquidCov}{1.000} 
\newcommand{\ctClefFlashCI}{[.611, .651]} 
\newcommand{\ctClefFlashWorstLang}{sw} 
\newcommand{\ctClefFlashWorst}{.47} 
\newcommand{\ctClefFlashGap}{\minus.133} 
\newcommand{\ctClefFlashCov}{1.000} 
\newcommand{\ctMercuryCI}{[.588, .638]} 
\newcommand{\ctMercuryWorstLang}{sw} 
\newcommand{\ctMercuryWorst}{.45} 
\newcommand{\ctMercuryGap}{\minus.110} 
\newcommand{\ctMercuryCov}{1.000} 
\newcommand{\ctLunaCI}{[.545, .591]} 
\newcommand{\ctLunaWorstLang}{sw} 
\newcommand{\ctLunaWorst}{.50} 
\newcommand{\ctLunaGap}{\minus.084} 
\newcommand{\ctLunaCov}{.998} 
\newcommand{\ctKevCI}{[.428, .470]} 
\newcommand{\ctKevWorstLang}{sw} 
\newcommand{\ctKevWorst}{.28} 
\newcommand{\ctKevGap}{\minus.150} 
\newcommand{\ctKevCov}{1.000} 
\newcommand{\ctSolarCI}{[.373, .422]} 
\newcommand{\ctSolarWorstLang}{sw} 
\newcommand{\ctSolarWorst}{.14} 
\newcommand{\ctSolarGap}{\minus.141} 
\newcommand{\ctSolarCov}{.773} 
\newcommand{\ctMsftCI}{[.633, .669]} 
\newcommand{\ctMsftWorstLang}{sw} 
\newcommand{\ctMsftWorst}{.52} 
\newcommand{\ctMsftGap}{\minus.131} 
\newcommand{\ctMsftCov}{1.000} 
\newcommand{\ctClefLocalCI}{[.655, .695]} 
\newcommand{\ctClefLocalWorstLang}{sw} 
\newcommand{\ctClefLocalWorst}{.56} 
\newcommand{\ctClefLocalGap}{\minus.103} 
\newcommand{\ctClefLocalCov}{1.000} 
\newcommand{\ctClefFlashLocalCI}{[.611, .651]} 
\newcommand{\ctClefFlashLocalWorstLang}{sw} 
\newcommand{\ctClefFlashLocalWorst}{.47} 
\newcommand{\ctClefFlashLocalGap}{\minus.133} 
\newcommand{\ctClefFlashLocalCov}{1.000} 
\newcommand{\ctQwenTwentySevenHostedCI}{[.562, .610]} 
\newcommand{\ctQwenTwentySevenHostedWorstLang}{sw} 
\newcommand{\ctQwenTwentySevenHostedWorst}{.45} 
\newcommand{\ctQwenTwentySevenHostedGap}{\minus.134} 
\newcommand{\ctQwenTwentySevenHostedCov}{1.000} 
\newcommand{\ctQwenNineHostedCI}{[.455, .499]} 
\newcommand{\ctQwenNineHostedWorstLang}{sw} 
\newcommand{\ctQwenNineHostedWorst}{.29} 
\newcommand{\ctQwenNineHostedGap}{\minus.166} 
\newcommand{\ctQwenNineHostedCov}{1.000} 
\newcommand{\ctQwenTwentySevenLocalCI}{[.390, .431]} 
\newcommand{\ctQwenTwentySevenLocalWorstLang}{hi} 
\newcommand{\ctQwenTwentySevenLocalWorst}{.30} 
\newcommand{\ctQwenTwentySevenLocalGap}{\minus.129} 
\newcommand{\ctQwenTwentySevenLocalCov}{1.000} 
\newcommand{\ctQwenNineLocalCI}{[.388, .417]} 
\newcommand{\ctQwenNineLocalWorstLang}{hi} 
\newcommand{\ctQwenNineLocalWorst}{.29} 
\newcommand{\ctQwenNineLocalGap}{\minus.110} 
\newcommand{\ctQwenNineLocalCov}{1.000} 
\newcommand{\skSolarCoreTwelveUncapped}{.497} 
\newcommand{\ctSolarCappedCells}{12} 
\newcommand{\nRankedCoreTwelve}{eleven} 
\newcommand{\rkPplxCoreTwelve}{1--3} 
\newcommand{\rkJevCoreTwelve}{1--7} 
\newcommand{\rkClefCoreTwelve}{2--5} 
\newcommand{\rkLiquidCoreTwelve}{2--6} 
\newcommand{\rkClefFlashCoreTwelve}{3--8} 
\newcommand{\rkMercuryCoreTwelve}{5--9} 
\newcommand{\rkLunaCoreTwelve}{5--9} 
\newcommand{\rkKevCoreTwelve}{8--11} 
\newcommand{\rkSolarCoreTwelve}{6--11} 
\newcommand{\rkMsftCoreTwelve}{1--7} 
\newcommand{\rkQwenNineLocalCoreTwelve}{9--11} 
\newcommand{\lodoTauLo}{0.82} 
\newcommand{\lodoTauHi}{0.96} 
\newcommand{\costApiTotal}{78.26} 
\newcommand{\costApiBenchmark}{69.94} 
\newcommand{\costApiFresh}{2.44} 
\newcommand{\costApiExploratory}{5.53} 
\newcommand{\costApiPilot}{0.35} 
\newcommand{\costApiDecision}{31.82} 
\newcommand{\costApiBackbones}{46.44} 
\newcommand{\costApiDecisionMax}{6.66} 
\newcommand{\costTokensInM}{797} 
\newcommand{\costTokensOutM}{73} 
\newcommand{\costHHundredHours}{41.4} 
\newcommand{\costRtxHours}{14.1} 
\newcommand{\costMmbertTrainHours}{2.4} 
\newcommand{\costCpuHours}{57.2}

\title{XDecideBench: Benchmarking Typed Decision Models\\ and Confidence Thresholds Across \NumHeadlineLangs{} Languages}

\hypersetup{
  pdftitle={XDecideBench: Benchmarking Typed Decision Models and Confidence Thresholds Across \NumHeadlineLangs{} Languages},
  pdfsubject={A multilingual benchmark of \NumHeadlineTracks{} suites in \NumHeadlineLangs{} languages for Typed
    Decision Models (System One models) and their confidence thresholds},
  pdfkeywords={multilingual benchmark, Typed Decision Models, Jev, confidence thresholds, calibration,
    selective prediction, cross-lingual transfer},
  pdflang={en},
}

\ifpreprint
  \author{Marek \v{S}uppa \\
  Comenius University in Bratislava \\
  Cisco Systems
  \And
  Jaroslav Kop\v{c}an \\
  Kempelen Institute of \\
  Intelligent Technologies}
\hypersetup{pdfauthor={Marek \v{S}uppa, Jaroslav Kop\v{c}an}}

\else
  \author{Anonymous submission}
\fi

\begin{document}

\maketitle

\begin{abstract}
Typed Decision Models such as Jev, Clef and pplx-decider return a probability for each answer an application provides.
Applications use them in many languages, but are compared mostly in English.
We introduce \xdb{}, a multilingual benchmark of \NumHeadlineTracks{} suites in \NumHeadlineLangs{} languages.
On our twelve-language suite, three open models fine-tuned for typed decisions beat their Qwen base models in every language.
On new test data with local base models, their accuracy gain is larger in languages with less web text than in those with more.
They fix a similar share of base-model errors in both, so the larger gain may reflect weaker base models.
The best model on the twelve-language suite is best on only \regCoreRightSuites{} other suites.
The best English model on each suite, chosen in hindsight, is significantly beaten in about a quarter of suite and language combinations.
On Belebele, Decision-1 is best in English but falls far behind Jev in Burmese.
A confidence threshold fitted on English to allow 5\% errors gives a higher error rate on XNLI, even in German and French.
\xdb{} shows that results from one suite or from English are a poor guide to how a model and its threshold behave in other languages.
\end{abstract}

\begin{figure}[t]
  \centering
  \includegraphics[width=\columnwidth]{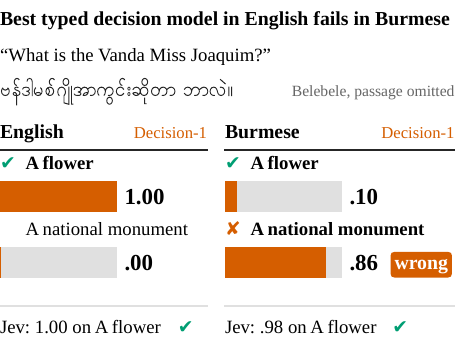}
  \caption{\textbf{The best model in English is confidently wrong on this Burmese item.} A real Belebele item, selected after the runs, with the Burmese options glossed in English. Bars show the probabilities of Microsoft's Decision-1, the best model on English Belebele. Two options that both Decision-1 and Jev put near zero are omitted. Over all Burmese items its Brier skill is \regWorstPick{}, against \regWorstBest{} for Jev (\S\ref{sec:res-matrix}).}
  \label{fig:teaser}
\end{figure}

\section{Introduction}
\label{sec:intro}

Many calls that applications make to language models are decisions, such as routing a request to one of sixty intents, flagging a message as abusive or rating its anger on a four-level scale.
Typed Decision Models \citep{lu2026rejection,gan2026sansi}, also called System One models \citep{typesafe2026jev}, answer such calls directly (Figure~\ref{fig:teaser}).
With each request the application declares the possible answers (named options, yes or no, or a level on an ordered rubric), and the model returns a probability for each.
We call these \choice{}, \noul{} and \score{} decisions \citep{typesafe2026jev}.
Some are \emph{decision heads}, built by fine-tuning a general language model (the \emph{backbone}) to return such probabilities.
A \emph{backbone readout} asks the untuned backbone the same question and reads its option probabilities, locally or through a hosted API.
An application that acts only on confident decisions also needs a threshold, which is cheapest to fit on English development data.

Model rankings change from one test suite to another.
On \coretwelve{}, our development cohort of five datasets in twelve languages, GPT-6 Luna ranks eighth of the \nCompleteModels{} models with complete runs, but it is first on the native INCLUDE exams, at \skLunaInclude{} against \skPplxInclude{} for the runner-up.
On the other suites, Kendall $\tau$ with the \coretwelve{} ranking is only \tauCoreVsRtplx--\tauCoreVsNli.
The model that wins in English is not a safe choice elsewhere. Judged in hindsight, it loses significantly in \regEnWrongPaired{} of the \regEnUnits{} suite--language cells and reaches only \regWorstPick{} on Burmese Belebele.
A threshold set on English data for 5\% errors lets through \aThreeXnliDeFrRiskLo--\aThreeXnliDeFrRiskHi\% on German and French XNLI.

Earlier studies evaluate either one Typed Decision Model across languages or several models in English (\S\ref{sec:related}), whereas \xdb{} evaluates several models on the same multilingual items.
RQ1 asks how the models compare across suites and label regimes, RQ2 how much open heads gain over their own backbones, and RQ3 whether confidence policies fitted in English hold in other languages.
Our first contribution is a common scoring contract (\S\ref{sec:protocol}). It fixes what counts as a valid answer, how refusals are scored, which items form a cell and how we resample. Within a suite, Jev in Thai and Clef in German are then scored alike.

Our second contribution is the benchmark itself (\S\ref{sec:benchmark}), with \NumTextsUnique{} texts and \NumQuestionsUnique{} questions in \NumHeadlineLangs{} languages, grouped into \NumHeadlineTracks{} aligned, locally labelled or native suites.
We score \NumDecisionModels{} Typed Decision Models from \NumVendors{} developers, the stated Qwen backbones of three open heads (two of them locally verified) and a supervised mmBERT comparator (\S\ref{sec:panel}).
Each open head beats its backbone in all \coretwelve{} languages under both readouts and gains more accuracy where web text is scarce.
Under the local readout, a pre-registered held-out cohort confirms this larger gain and shows that the heads remove a similar share of backbone errors in low- and high-web-share languages (\S\ref{sec:res-a1}).

\section{Related Work}
\label{sec:related}

\begin{table}[t]
  \centering
  \footnotesize
  \setlength{\tabcolsep}{1.5pt}
  \begin{tabular}{@{}>{\raggedright\arraybackslash}p{6.6em}S[table-format=3]S[table-format=2]lccccc@{}}
    \toprule
    Study & {Langs} & {DMs} & Types & Same & P/N & H--B & Transf. \\
    \midrule
    \citeauthor{deusser2026jev}     & 200\rlap{+} & 1  & CNS & \tmark{--} & \tmark{--} & \tmark{--} & \tmark{--} \\
    \citeauthor{yanng2026systemone} & 8      & 5  & C   & \checkmark & \tmark{--} & \tmark{--} & \tmark{--} \\
    \citeauthor{convai2026laya}     & 51     & 1  & CNS & \tmark{--} & \tmark{--} & \tmark{--} & \tmark{--} \\
    \citeauthor{tateno2026s1mb}     & 1      & 23 & CNS & \checkmark & \tmark{--} & \tmark{--} & \tmark{--} \\
    \citeauthor{rafe2026decision}   & 1      & 8  & CN  & \checkmark & \tmark{--} & \checkmark & \tmark{--} \\
    \citeauthor{lu2026rejection}    & 1      & 2  & C   & \checkmark & \tmark{--} & \tmark{--} & task \\
    \midrule
    \textbf{\xdb{}}                 & 122    & 11 & CNS & \checkmark & \checkmark & \checkmark & lang. \\
    \bottomrule
  \end{tabular}
  \tabnote{\emph{Langs}: languages evaluated (ours: headline suites). \emph{DMs}: Typed Decision Models or checkpoints. \emph{Types}: \choice{}, \noul{}, \score{}. \emph{Same}: models on identical items. \emph{P/N}: aligned and native evidence reported separately. \emph{H--B}: heads against own backbones. \emph{Transf.}: a confidence policy reused on another task or language. \textbf{\xdb{}}: this paper.}
  \caption{\textbf{No prior study covers more than two of the last four columns.}}
  \label{tab:related}
\end{table}

\paragraph{Typed Decision Models.}
\citet{deusser2026jev} evaluate Jev on 37 datasets, including Belebele in 122 varieties and SIB-200 in 205 configurations.
They report calibration and selective prediction, but compare Jev only with ordinary language models read through their option probabilities.
Community benchmarks compare several such models on shared items, either in English with all three decision types \citep{tateno2026s1mb,decisionindex2026,jevbench2026} or in eight languages with \choice{} only \citep{yanng2026systemone}.
Laya's report covers 51 MASSIVE languages with a 20-intent subset \citep{convai2026laya}.
\citet{rafe2026decision} and OpenDecider \citep{opendecider2026} compare decision checkpoints with their untuned backbones in English.
Other English-centred preprints benchmark calibration \citep{sharma2026typedbench}, audit agent decisions \citep{li2026fastmodels} and train a looped model \citep{gan2026sansi}.
\citet{lu2026rejection} find that a rejection policy calibrated on one task loses its operating point on another (Table~\ref{tab:related}).

\paragraph{Multilingual calibration and benchmark design.}
Ordinary multilingual models are miscalibrated outside English, and temperatures fitted in English transfer poorly \citep{ahuja2022calibration,jiang2022calibrating,yang2024calibration,zhou2025multilingual}.
English-heavy instruction tuning can raise low-resource confidence without raising accuracy \citep{huang2026eacl}.
The closest study is MlingConf \citep{xue2025mlingconf}, which finds that English dominates LLM confidence estimation across languages.
We ask whether an English-fitted policy, either a risk threshold \citep{geifman2017selective} or a split conformal set \citep{sadinle2019least,angelopoulos2023conformal}, keeps its operating point in other languages for several Typed Decision Models.
AfroBench and BenchMAX rank general LLMs on African languages and post-edited translations \citep{ojo2025afrobench,huang2025benchmax}, and MuBench aligns items for generative models \citep{han2025mubench}.
\xdb{} adds typed decisions, matched head--backbone pairs and a pre-registered held-out confirmation.
Like Belebele and XTREME-S, we fix parallel items \citep{bandarkar2024belebele,conneau2022xtremes}. Because translation adds its own artefacts \citep{artetxe2020translation}, we report native material separately, following INCLUDE and Global-MMLU \citep{romanou2025include,singh2025globalmmlu}.
Our chance-normalised skills follow SEA-HELM and MMTEB \citep{susanto2025seahelm,enevoldsen2025mmteb}, and as in HELM, failures stay in the score \citep{liang2023helm}.
We resample clusters \citep{miller2024errorbars} and give each score a sacreBLEU-style signature \citep{post2018sacrebleu}.

\section{The Benchmark}
\label{sec:benchmark}

\begin{figure*}[t]
  \centering
  \includegraphics[width=\textwidth]{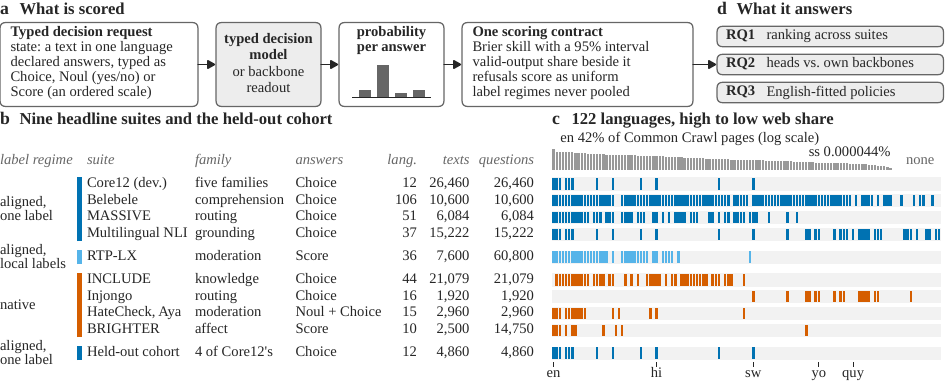}
  \caption{\textbf{Overview of \xdb{}.} (a)~What we score (\S\ref{sec:protocol}). (b)~Headline suites by label regime, and the held-out cohort. A question is one typed decision about a text. Multilingual NLI reuses the \NumCoreXnliTexts{} \coretwelve{} XNLI texts, so the suites hold \NumTextsUnique{} distinct texts and \NumQuestionsUnique{} questions. (c)~One tick per covered language, ordered by share of Common Crawl pages (log scale). Languages missing from the crawl use their macrolanguage's share, and languages with neither are at the right. (d)~Research questions (\S\ref{sec:results}).}
  \label{fig:overview}
\end{figure*}

\paragraph{Three label regimes.}
\xdb{} scores only the probabilities that a Typed Decision Model returns for the declared answers (Figure~\ref{fig:overview}a).
In four suites the same source items appear in all languages with one label per item, so a gap between languages reflects language and translation rather than content.
RTP-LX also aligns its items across \suiteRtplxVarieties{} locales, but three annotators from each locale label them, and the gold label is their median.
The gold toxicity label differs from the English one for \rtplxToxDiffer{} of \rtplxPairs{} aligned locale--item pairs (\rtplxToxDifferPct\%), and across all eight rubrics \rtplxAllDifferPct\% of labels differ (Appendix~\ref{app:rtplx}).
A language gap on RTP-LX therefore measures how well a model predicts local labels.
The four native suites contain items written in each language for its speakers and test local validity. Since topic, culture and difficulty differ, we never compare their raw scores across languages \citep{romanou2025include,singh2025globalmmlu,hupkes2025multiloko}.

\paragraph{Suites.}
Besides \coretwelve{}, the aligned suites are Wide-Belebele \citep{bandarkar2024belebele}, MASSIVE, an NLI family \citep{conneau2018xnli,adelani2025irokobench,ebrahimi2022americasnli} and RTP-LX toxicity \citep{dewynter2025rtplx} (Figure~\ref{fig:overview}b).
MASSIVE covers all \suiteMassiveVarieties{} locales on \massiveFiftyTwoUtterances{} utterances unused elsewhere \citep{fitzgerald2023massive}.
The native suites are INCLUDE \citep{romanou2025include}, InjongoIntent in \suiteInjongoVarieties{} African languages \citep{yu2025injongo}, Multilingual HateCheck with Aya red-teaming prompts \citep{rottger2022mhc,aakanksha2024prism} and BRIGHTER emotion intensity \citep{muhammad2025brighter} (Appendix~\ref{app:tracks}).
Belebele, NLI and Injongo reach the languages with the least web text (Figure~\ref{fig:overview}c).

\paragraph{Scoring contract.}
We score a suite only when a submission covers every case in every cell \citep{conneau2022xtremes}.
Refusals, interface limits and invalid outputs count as a uniform distribution, and we print the share of valid outputs beside each score \citep{liang2023helm,himmi2023missing}.
A cell's score is skill relative to a declared reference $r$.
For \choice{}, $r$ is uniform chance, and for the skewed \noul{} and \score{} datasets it is the cell's constant base-rate predictor.
With $\mathrm{BS}$ the mean multiclass Brier score, the primary metric is $\mathrm{BSS} = 1 - \mathrm{BS}(\text{model}) / \mathrm{BS}(r)$.
The Brier score is strictly proper and bounded, though BSS is not bounded below. Unlike log loss, it stays finite when rounded vendor probabilities contain exact zeros \citep{gneiting2007strictly}.
We average with equal weights at each level from items up to suites, but never across suites \citep{enevoldsen2025mmteb}.
Each mean has a cluster-bootstrap interval (\S\ref{sec:protocol}) and a signature, and we never rank scores with different signatures together.

\paragraph{\coretwelve{}.}
\coretwelve{} covers twelve languages from tiers 2 to 5 of \citet{joshi2020state} (Arabic, German, English, Spanish, French, Chinese, Hindi, Russian, Vietnamese, Thai, Urdu and Swahili) and keeps the audited items of five labelled sources that exist in all twelve.
The sources are MASSIVE intents over all 60 classes \citep{bastianelli2020slurp}, XNLI \citep{conneau2018xnli}, Belebele, full MMLU-ProX \citep{xuan2025mmluprox} and SIB-200 topics \citep{adelani2024sib200}, with \coreItemsPerLangMassive{}, \coreItemsPerLangXnli{}, \coreItemsPerLangBelebele{}, \coreItemsPerLangMmluProx{} and \coreItemsPerLangSib{} items per language.
For the headline \coretwelve{} requests we reuse the source labels, translations and distractors.
Each request pairs target-language content with an English instruction and English option texts, as in an application with an English schema.
MASSIVE is localized, XNLI and SIB-200 are translated and MMLU-ProX is machine-translated with expert review.

\paragraph{Development and held-out cohorts.}
\coretwelve{} is the development cohort, and the \fresh{} has \freshCases{} cases (\freshItemsPerLang{} source items per language) from MASSIVE, XNLI, Belebele and MMLU-ProX in the \coretwelve{} languages.
None of its cases shares a source item, near-duplicate text or input hash with any other manifest.
We froze it before any inference and ran it only after pre-registration.
\Afterreview{} will release pinned manifests that rebuild identical cases\releaseat, but not the request texts (Appendices~\ref{app:data} and~\ref{app:compute}).

\section{Panel and Readouts}
\label{sec:panel}

\begin{table}[t]
  \centering
  \footnotesize
  \setlength{\tabcolsep}{1.6pt}
  \begin{tabular}{@{}llll@{}}
    \toprule
    Typed Decision Model & Access & Backbone & Precision \\
    \midrule
    \textbf{Clef}                  & local, API & Qwen3.8-27B$^\dagger$  & fp32/4 dp \\
    \textbf{Clef-flash}            & local, API & Qwen3.5-9B$^\dagger$   & fp32/4 dp \\
    \textbf{pplx-decider}-v1.1     & API        & Qwen3.8-27B$^\ddagger$ & full \\
    Microsoft-\textbf{Decision-1}  & API        & Qwen3.5-9B$^\ddagger$  & full \\
    \textbf{Kev-4B}                & API        & Qwen3.5-4B$^\ddagger$  & 4 dp \\
    \textbf{Laya}-multilingual     & local      & mmBERT-base            & full \\
    \addlinespace[3pt]
    \textbf{Jev} 1.13              & API        & \tmark{undisclosed}    & 2 dp \\
    GPT-6 \textbf{Luna} Decisions  & API        & \tmark{undisclosed}    & 2 dp \\
    Liquid \textbf{d1}             & API        & \tmark{undisclosed}    & full \\
    \textbf{Mercury} Decide        & API        & \tmark{undisclosed}    & full \\
    \textbf{Solar} Decide          & API        & \tmark{undisclosed}    & 6 dp \\
    \bottomrule
  \end{tabular}
  \tabnote{\textbf{Bold}: the short name used in the text and all tables. API models are dated snapshots, except the undated hosted Clef routes. $^\dagger$Locally verified pair: the head's pinned local weights and this exact backbone revision. $^\ddagger$Base stated by the vendor and not verified by us. \emph{Precision}: of the returned probabilities (dp: decimal places).}
  \caption{\textbf{The eleven Typed Decision Models.} Three are paired with their own Qwen backbone, and a fourth pair, Decision-1, is in Appendix~\ref{app:second}.}
  \label{tab:panel}
\end{table}

\begin{table*}[t]
  \centering
  \fitwidth{\footnotesize
\setlength{\tabcolsep}{1.5pt}
\begin{tabular}{@{}lS[table-format=-0.3]@{}lS[table-format=-0.3]@{}lS[table-format=-0.3]@{}lS[table-format=-0.3]@{}lS[table-format=-1.3]@{}lS[table-format=-0.3]@{}lS[table-format=-0.3]@{}lS[table-format=-0.3]@{}lS[table-format=-1.3]@{}l@{}}
  \toprule
   & \multicolumn{8}{c}{Aligned, shared labels} & \multicolumn{2}{c}{Locale labels} & \multicolumn{8}{c}{Native-authored} \\
  \cmidrule(lr){2-9}\cmidrule(lr){10-11}\cmidrule(lr){12-19}
  Model & \multicolumn{2}{c}{\coretwelve{}} & \multicolumn{2}{c}{Belebele} & \multicolumn{2}{c}{MASSIVE} & \multicolumn{2}{c}{NLI} & \multicolumn{2}{c}{RTP-LX} & \multicolumn{2}{c}{INCLUDE} & \multicolumn{2}{c}{Injongo} & \multicolumn{2}{c}{Moder.} & \multicolumn{2}{c}{BRIGHTER} \\
  \textit{Reference} & \multicolumn{2}{c}{\textit{chance}} & \multicolumn{2}{c}{\textit{chance}} & \multicolumn{2}{c}{\textit{chance}} & \multicolumn{2}{c}{\textit{chance}} & \multicolumn{2}{c}{\textit{base rate}} & \multicolumn{2}{c}{\textit{chance}} & \multicolumn{2}{c}{\textit{chance}} & \multicolumn{2}{c}{\textit{mixed}} & \multicolumn{2}{c}{\textit{base rate}} \\
  \midrule
  \tgroup{19}{Hosted Typed Decision Models, by \coretwelve{} skill} \\
  pplx-decider & \bfseries \skPplxCoreTwelve & \tunc{\hwPplxCoreTwelve} & \skPplxBelebele\tunder{\skPplxBelebele} & \tunc{\hwPplxBelebele} & \skPplxMassive & \tunc{\hwPplxMassive} & \bfseries \skPplxNli & \tunc{\hwPplxNli} & \skPplxRtplx & \tunc{\hwPplxRtplx} & \skPplxInclude\tunder{\skPplxInclude} & \tunc{\hwPplxInclude} & \skPplxInjongo\tunder{\skPplxInjongo} & \tunc{\hwPplxInjongo} & \bfseries \skPplxModer & \tunc{\hwPplxModer} & \skPplxBrighter\tunder{\skPplxBrighter} & \tunc{\hwPplxBrighter} \\
  Jev & \skJevCoreTwelve\tunder{\skJevCoreTwelve} & \tunc{\hwJevCoreTwelve} & \bfseries \skJevBelebele & \tunc{\hwJevBelebele} & \skJevMassive & \tunc{\hwJevMassive} & \skJevNli & \tunc{\hwJevNli} & \skJevRtplx & \tunc{\hwJevRtplx} & \skJevInclude & \tunc{\hwJevInclude} & \skJevInjongo & \tunc{\hwJevInjongo} & \skJevModer\tunder{\skJevModer} & \tunc{\hwJevModer} & \skJevBrighter & \tunc{\hwJevBrighter} \\
  Clef & \skClefCoreTwelve & \tunc{\hwClefCoreTwelve} & \skClefBelebele & \tunc{\hwClefBelebele} & \skClefMassive & \tunc{\hwClefMassive} & \skClefNli & \tunc{\hwClefNli} & \skClefRtplx & \tunc{\hwClefRtplx}\rlap{$^\dagger$} & \skClefInclude & \tunc{\hwClefInclude} & \skClefInjongo & \tunc{\hwClefInjongo} & \skClefModer & \tunc{\hwClefModer} & \bfseries \skClefBrighter & \tunc{\hwClefBrighter} \\
  d1 & \skLiquidCoreTwelve & \tunc{\hwLiquidCoreTwelve} & \skLiquidBelebele & \tunc{\hwLiquidBelebele} & \skLiquidMassive & \tunc{\hwLiquidMassive} & \skLiquidNli\tunder{\skLiquidNli} & \tunc{\hwLiquidNli} & \skLiquidRtplx\tunder{\skLiquidRtplx} & \tunc{\hwLiquidRtplx} & \skLiquidInclude & \tunc{\hwLiquidInclude} & \skLiquidInjongo & \tunc{\hwLiquidInjongo} & \skLiquidModer & \tunc{\hwLiquidModer} & \skLiquidBrighter & \tunc{\hwLiquidBrighter} \\
  Decision-1 & \skMsftCoreTwelve & \tunc{\hwMsftCoreTwelve} & \skMsftBelebele & \tunc{\hwMsftBelebele} & \skMsftMassive\tunder{\skMsftMassive} & \tunc{\hwMsftMassive} & \skMsftNli & \tunc{\hwMsftNli} & \skMsftRtplx & \tunc{\hwMsftRtplx} & \skMsftInclude & \tunc{\hwMsftInclude} & \skMsftInjongo & \tunc{\hwMsftInjongo} & \skMsftModer & \tunc{\hwMsftModer} & \skMsftBrighter & \tunc{\hwMsftBrighter} \\
  Clef-flash & \skClefFlashCoreTwelve & \tunc{\hwClefFlashCoreTwelve} & \skClefFlashBelebele & \tunc{\hwClefFlashBelebele} & \bfseries \skClefFlashMassive & \tunc{\hwClefFlashMassive} & \skClefFlashNli & \tunc{\hwClefFlashNli} & \bfseries \skClefFlashRtplx & \tunc{\hwClefFlashRtplx}\rlap{$^\dagger$} & \skClefFlashInclude & \tunc{\hwClefFlashInclude} & \skClefFlashInjongo & \tunc{\hwClefFlashInjongo} & \skClefFlashModer & \tunc{\hwClefFlashModer} & \skClefFlashBrighter & \tunc{\hwClefFlashBrighter} \\
  Mercury & \skMercuryCoreTwelve & \tunc{\hwMercuryCoreTwelve} & \skMercuryBelebele & \tunc{\hwMercuryBelebele} & \skMercuryMassive & \tunc{\hwMercuryMassive} & \skMercuryNli & \tunc{\hwMercuryNli} & \skMercuryRtplx & \tunc{\hwMercuryRtplx} & \skMercuryInclude & \tunc{\hwMercuryInclude} & \skMercuryInjongo & \tunc{\hwMercuryInjongo} & \skMercuryModer & \tunc{\hwMercuryModer} & \skMercuryBrighter & \tunc{\hwMercuryBrighter} \\
  Luna & \skLunaCoreTwelve & \tunc{\hwLunaCoreTwelve}\rlap{$^\dagger$} & \skLunaBelebele & \tunc{\hwLunaBelebele}\rlap{$^\dagger$} & \skLunaMassive & \tunc{\hwLunaMassive}\rlap{$^\dagger$} & \skLunaNli & \tunc{\hwLunaNli} & \skLunaRtplx & \tunc{\hwLunaRtplx}\rlap{$^\dagger$} & \bfseries \skLunaInclude & \tunc{\hwLunaInclude}\rlap{$^\dagger$} & \bfseries \skLunaInjongo & \tunc{\hwLunaInjongo}\rlap{$^\dagger$} & \skLunaModer & \tunc{\hwLunaModer}\rlap{$^\dagger$} & \skLunaBrighter & \tunc{\hwLunaBrighter} \\
  Kev-4B & \skKevCoreTwelve & \tunc{\hwKevCoreTwelve} & \skKevBelebele & \tunc{\hwKevBelebele} & \skKevMassive & \tunc{\hwKevMassive} & \skKevNli & \tunc{\hwKevNli} & \skKevRtplx & \tunc{\hwKevRtplx} & \skKevInclude & \tunc{\hwKevInclude} & \skKevInjongo & \tunc{\hwKevInjongo} & \skKevModer & \tunc{\hwKevModer} & \skKevBrighter & \tunc{\hwKevBrighter} \\
  Solar & \skSolarCoreTwelve & \tunc{\hwSolarCoreTwelve}\rlap{$^\ddagger$} & \skSolarBelebele & \tunc{\hwSolarBelebele} & \multicolumn{2}{c}{\tmark{capped}} & \skSolarNli & \tunc{\hwSolarNli} & \skSolarRtplx & \tunc{\hwSolarRtplx} & \skSolarInclude & \tunc{\hwSolarInclude} & \multicolumn{2}{c}{\tmark{capped}} & \skSolarModer & \tunc{\hwSolarModer} & \skSolarBrighter & \tunc{\hwSolarBrighter} \\
  \addlinespace[2pt]
  \tgroup{19}{Local Typed Decision Model} \\
  Laya & \skLayaSuiteCoreTwelve & \tunc{\hwLayaSuiteCoreTwelve}\rlap{$^\dagger$} & \skLayaSuiteBelebele & \tunc{\hwLayaSuiteBelebele}\rlap{$^\dagger$} & \skLayaSuiteMassive & \tunc{\hwLayaSuiteMassive} & \skLayaSuiteNli & \tunc{\hwLayaSuiteNli} & \skLayaSuiteRtplx & \tunc{\hwLayaSuiteRtplx} & \skLayaSuiteInclude & \tunc{\hwLayaSuiteInclude}\rlap{$^\dagger$} & \skLayaSuiteInjongo & \tunc{\hwLayaSuiteInjongo} & \skLayaSuiteModer & \tunc{\hwLayaSuiteModer} & \skLayaSuiteBrighter & \tunc{\hwLayaSuiteBrighter} \\
  \addlinespace[2pt]
  \tgroup{19}{Backbone readouts, hosted} \\
  Qwen3.8-27B & \skQwenTwentySevenHostedCoreTwelve & \tunc{\hwQwenTwentySevenHostedCoreTwelve} & \skQwenTwentySevenHostedBelebele & \tunc{\hwQwenTwentySevenHostedBelebele} & \skQwenTwentySevenHostedMassive & \tunc{\hwQwenTwentySevenHostedMassive} & \skQwenTwentySevenHostedNli & \tunc{\hwQwenTwentySevenHostedNli} & \skQwenTwentySevenHostedRtplx & \tunc{\hwQwenTwentySevenHostedRtplx} & \skQwenTwentySevenHostedInclude & \tunc{\hwQwenTwentySevenHostedInclude} & \skQwenTwentySevenHostedInjongo & \tunc{\hwQwenTwentySevenHostedInjongo} & \skQwenTwentySevenHostedModer & \tunc{\hwQwenTwentySevenHostedModer} & \skQwenTwentySevenHostedBrighter & \tunc{\hwQwenTwentySevenHostedBrighter} \\
  Qwen3.5-9B & \skQwenNineHostedCoreTwelve & \tunc{\hwQwenNineHostedCoreTwelve} & \skQwenNineHostedBelebele & \tunc{\hwQwenNineHostedBelebele} & \skQwenNineHostedMassive & \tunc{\hwQwenNineHostedMassive} & \skQwenNineHostedNli & \tunc{\hwQwenNineHostedNli} & \skQwenNineHostedRtplx & \tunc{\hwQwenNineHostedRtplx} & \skQwenNineHostedInclude & \tunc{\hwQwenNineHostedInclude} & \skQwenNineHostedInjongo & \tunc{\hwQwenNineHostedInjongo} & \skQwenNineHostedModer & \tunc{\hwQwenNineHostedModer} & \skQwenNineHostedBrighter & \tunc{\hwQwenNineHostedBrighter} \\
  \addlinespace[2pt]
  \tgroup{19}{Comparators} \\
  MP-EG2 & \skMediaPipeCoreTwelve & \tunc{\hwMediaPipeCoreTwelve} & \skMediaPipeBelebele & \tunc{\hwMediaPipeBelebele} & \skMediaPipeMassive & \tunc{\hwMediaPipeMassive} & \skMediaPipeNli & \tunc{\hwMediaPipeNli} & \skMediaPipeRtplx & \tunc{\hwMediaPipeRtplx} & \skMediaPipeInclude & \tunc{\hwMediaPipeInclude} & \skMediaPipeInjongo & \tunc{\hwMediaPipeInjongo} & \skMediaPipeModer & \tunc{\hwMediaPipeModer} & \skMediaPipeBrighter & \tunc{\hwMediaPipeBrighter} \\
  MP-GLiNER & \skGlinerCoreTwelve & \tunc{\hwGlinerCoreTwelve}\rlap{$^\dagger$} & \skGlinerBelebele & \tunc{\hwGlinerBelebele}\rlap{$^\ddagger$} & \skGlinerMassive & \tunc{\hwGlinerMassive} & \skGlinerNli & \tunc{\hwGlinerNli} & \skGlinerRtplx & \tunc{\hwGlinerRtplx} & \skGlinerInclude & \tunc{\hwGlinerInclude}\rlap{$^\dagger$} & \skGlinerInjongo & \tunc{\hwGlinerInjongo}\rlap{$^\dagger$} & \skGlinerModer & \tunc{\hwGlinerModer} & \skGlinerBrighter & \tunc{\hwGlinerBrighter} \\
  \bottomrule
\end{tabular}
}
  \tabnote{Brier skill against each suite's reference, with failures scored as uniform. Grey subscripts: half-width of the 95\% joint cluster-bootstrap interval. \textbf{Bold}: best hosted Typed Decision Model per column. \underline{Underlined}: runner-up. $^\dagger$Valid-output coverage below 1. $^\ddagger$Partly over an interface cap (Table~\ref{tab:failzero}). \tmark{capped}: all items exceed Solar's option cap. Laya ran on its own copies of the suite manifests, and its model card lists MASSIVE and XNLI as evaluated in its development. Local runs (\coretwelve{} / Belebele only): Clef \skClefLocalCoreTwelve\tunc{\hwClefLocalCoreTwelve} / \skClefLocalBelebele\tunc{\hwClefLocalBelebele}, Clef-flash \skClefFlashLocalCoreTwelve\tunc{\hwClefFlashLocalCoreTwelve} / \skClefFlashLocalBelebele\tunc{\hwClefFlashLocalBelebele}. Local backbone readouts: Qwen3.8-27B \skQwenTwentySevenLocalCoreTwelve\tunc{\hwQwenTwentySevenLocalCoreTwelve} / \skQwenTwentySevenLocalBelebele\tunc{\hwQwenTwentySevenLocalBelebele}, Qwen3.5-9B \skQwenNineLocalCoreTwelve\tunc{\hwQwenNineLocalCoreTwelve} / \skQwenNineLocalBelebele\tunc{\hwQwenNineLocalBelebele}. Names: Table~\ref{tab:panel}.}
  \caption{\textbf{Rankings change across suites.} Suites differ in reference and population, so scores should not be compared across columns.}
  \label{tab:matrix}
\end{table*}

\paragraph{Typed Decision Models.}
Table~\ref{tab:panel} lists the \NumDecisionModels{} models that we run on identical requests.
They are TypeSafe's Jev \citep{typesafe2026jev}, the open-weight heads Clef, Clef-flash, pplx-decider and Kev \citep{cloudflare2026clef,perplexity2026pplx,palmer2026kev}, hosted models from Liquid, Upstage, Inception, OpenAI and Microsoft \citep{liquid2026d1,upstage2026solar,inception2026mercury,openai2026decisions,microsoft2026decision} and the small encoder Laya \citep{convai2026laya}.
Where head training is documented, it is mostly in English.
We also run Clef and Clef-flash locally on an H100 from pinned public weights, and Laya from a pinned checkpoint.
The hosted Clef routes name no checkpoint, but their top answer matches the local one on \hostedLocalTopOneClef\% of \coretwelve{} cases for Clef and on \hostedLocalTopOneClefFlash\% for Clef-flash (Appendix~\ref{app:compute}).

\paragraph{Backbones and readouts.}
For the three primary head--backbone pairs we also score the declared base \citep{qwen2026qwen38,qwen2026qwen35}.
Two pairs, Clef with Qwen3.8-27B and Clef-flash with Qwen3.5-9B, are \emph{locally verified}: we run both head and backbone from pinned local weights. For pplx-decider we rely on the vendor's statement of the base.
Our provider offers no log-probabilities for Kev-4B's stated base, Qwen3.5-4B, so we did not run it.
The \emph{local} readout gives the backbone the same input, instruction and option texts through one frozen prompt (Appendix~\ref{app:prompt}) in its own chat template with thinking disabled.
We teacher-force each option text and take the softmax over mean per-token log-likelihoods, which gives an exact but uncalibrated distribution.
The \emph{hosted} readout sends the prompt with lettered options to a pinned bf16 (9B) or fp16 (27B) endpoint.
It reads the label distribution from the top-20 log-probabilities. It is our only backbone readout outside \coretwelve{} and Wide-Belebele.
The pre-registered hypotheses use the local readout, which scores the heads' own option texts exactly, and the hosted readout is a sensitivity check.

\paragraph{Comparators and limits.}
As a supervised comparator, we fine-tune Laya's backbone, mmBERT-base, on the training data of MASSIVE, SIB-200 and English MultiNLI \citep{marone2025mmbert,williams2018mnli}.
Laya uses the same encoder and has near-zero Brier skill on MASSIVE (\skLayaMassive{}) despite \layaMassiveAccuracy{} accuracy.
We also score the MediaPipe Decision Maker with EmbeddingGemma~2 and with GLiNER2.5-Decide as off-the-shelf comparators (MP-EG2 and MP-GLiNER).
Vendors return probabilities at different precisions, and each model has a fixed acceptance rule that matches its rounding (Appendix~\ref{app:precision}).
Because two-decimal APIs return exact zeros, we floor log loss at $10^{-6}$.
Interface limits count as failures rather than missing data.
Solar Decide rejects \choice{} questions with more than 26 options, so all 60-intent MASSIVE cases score as uniform. Laya refuses inputs that its description cap would truncate.

\section{Protocol and Statistics}
\label{sec:protocol}

\paragraph{Evidence status.}
We use \coretwelve{} for development, and the \fresh{} below is the prospective test.
We fixed the \coretwelve{} manifests and an accuracy-based protocol before any benchmark inference, and registered each model with its dated snapshot before its first benchmark request (Table~\ref{tab:chrono}).
We made Brier skill the primary suite score after most \coretwelve{} runs, but before any request on the other eight suites, and pinned those suites' manifests at the same time.
On \coretwelve{}, the Brier-skill and chance-corrected-accuracy rankings of the \nHostedDecisionModels{} hosted models agree at Kendall $\tau_b=\tauAccuracyVsBrierHosted$.

\paragraph{Head--backbone comparison.}
For a pair, $\Dhb(d,\ell)$ is the head's metric minus its backbone's metric on the same cases of dataset $d$ and language $\ell$.
Accuracy uses all cases, while Brier score and adaptive expected calibration error (ECE; \citealp{nixon2019measuring}) use the cases where both models return a valid answer.
The per-language effect averages the five datasets with equal weight.
The interaction contrast is the mean $\Dhb$ over a low-web-share group \{Hindi, Swahili, Thai, Urdu\} minus the mean over a high-web-share group \{English, German, Spanish, French, Russian, Chinese\}.
Web share, our rough proxy for pretraining exposure, is at least \webShareHighMin\% of Common Crawl pages in the high group and at most \webShareLowMax\% in the low group \citep{commoncrawl2026languages}, and Arabic and Vietnamese fall in between.
An absolute gain cannot exceed the backbone's error rate, so we report the share of backbone errors removed, $(a_h-a_b)/(1-a_b)$ for accuracies $a_h$ and $a_b$.

\paragraph{Calibration with an English temperature.}
For each head and its backbone's local readout, we fit one temperature per dataset (MASSIVE, XNLI) on English development data and apply it unchanged to all \coretwelve{} languages.
We report ECE and Brier score before and after scaling, which leaves accuracy unchanged.

\paragraph{Transfer of English-fitted confidence policies.}
We split the English development data (\aThreeDevMassive{} MASSIVE and \aThreeDevXnli{} XNLI decisions) by source cluster and fit a temperature on one half.
On the other half we fit thresholds at 1\%, 5\% and 10\% empirical risk and a split-conformal set with score $1-p(y)$ at $\alpha=0.1$ \citep{sadinle2019least,angelopoulos2023conformal}.
We apply these policies unchanged to all \coretwelve{} languages on MASSIVE and XNLI, for each hosted model with English development runs, and report risk among admitted decisions, admitted share, conformal coverage and set size.
The targets are nominal, with thresholds estimated on \aThreeThresholdN{} English decisions, and conformal coverage assumes exchangeability with the English data.

\begin{figure}[!t]
  \centering
  \includegraphics[width=\columnwidth]{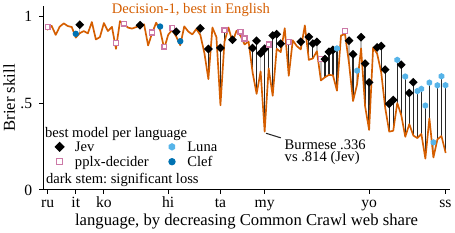}
  \caption{\textbf{The best model in English is often beaten in other languages.} Belebele Brier skill of Decision-1, which is best in English (line), and of the best model per language, chosen in hindsight (markers). The \bbUnmapped{} languages without a web share are left out. Dark stems: significant paired losses (\regBbPaired{} of \regBbUnits{} languages, \regBbShownPaired{} shown).}
  \label{fig:rq1}
\end{figure}

\paragraph{Uncertainty.}
Translations of one source item, questions on one Belebele passage and XNLI hypotheses with a shared premise are dependent, so all intervals resample whole source clusters.
Contrasts use a paired percentile bootstrap in which both models and all target languages share one cluster resample.
For transfer, we recompute admitted sets, risk and coverage in each replicate.
For suite scores, all languages and models of a parallel dataset share one draw of clusters, while native datasets resample within each language \citep{miller2024errorbars}.
The intervals reflect only item sampling, for this fixed set of suites and languages.
Development-cohort intervals are not adjusted for multiplicity, and we call a contrast \emph{significant} when its 95\% interval excludes zero.

\paragraph{Confirmation on the \fresh{}.}
Before any request on the \fresh{} (Table~\ref{tab:chrono}), we pre-registered the panel, the analysis code and two hypotheses on the local readout, pooled over the three head--backbone pairs.
Holm correction controls the familywise error at $\alpha=\freshAlpha$ \citep{holm1979simple}.
H1 states that the heads gain more accuracy in the low-web-share group than in the high one.
H2 states that the share of backbone errors removed is the same in both groups. We test equivalence with two one-sided tests (TOST) and a margin of $\pm\hTwoMargin$, about \hTwoMarginPointsLo{} to \hTwoMarginPointsHi{} accuracy points at low-web-share backbone error rates.
Simulated power is \hOnePower\% for H1 at the \coretwelve{} effect and \hTwoPower\% for H2 at a true difference of zero (Appendix~\ref{app:compute}).
We will release the registration\afterreview.

\section{Results}
\label{sec:results}

\subsection{Decision quality across suites (RQ1)}
\label{sec:res-matrix}

Table~\ref{tab:matrix} gives all model--suite scores.
On \coretwelve{}, the hosted Typed Decision Models range from \skHostedCoreTwelveLo{} to \skHostedCoreTwelveHi{}, and all except Solar beat both local backbone readouts.
The hosted readout of plain Qwen3.8-27B (\skQwenTwentySevenHostedCoreTwelve{}) scores above GPT-6 Luna, Kev-4B and Solar.
Only coarse tiers are stable, and the rank intervals of the top models overlap (pplx-decider \rkPplxCoreTwelve, Jev \rkJevCoreTwelve, Decision-1 \rkMsftCoreTwelve, d1 \rkLiquidCoreTwelve{} in Table~\ref{tab:core12}).
On the three \coretwelve{} datasets it was trained for, the supervised mmBERT comparator reaches \skMmbertThreeSets{} \skMmbertThreeSetsCI{}, and Laya, which uses the same encoder, reaches \skLayaThreeSets{} \skLayaThreeSetsCI{} (\skLayaSuiteCoreTwelve{} on all of \coretwelve{}).
Both MediaPipe arms trail all hosted models on \coretwelve{}, Belebele, RTP-LX and INCLUDE, and Laya beats its reference on only one of its nine suites, \coretwelve{}.

\paragraph{Rankings depend on the task.}
On the other suites, Kendall $\tau$ with the \coretwelve{} ranking of the \nCompleteModels{} models with complete runs is only \tauCoreVsRtplx{} (RTP-LX) to \tauCoreVsNli{} (NLI, Injongo).
GPT-6 Luna is eighth on \coretwelve{} but first on the native INCLUDE exams (\skLunaInclude{}, against \skPplxInclude{} for pplx-decider).
It also shares the Injongo lead with pplx-decider (\skLunaInjongo{} against \skPplxInjongo{}) but falls below chance on NLI (\skLunaNli{}).
On the two score-type suites, several models do no better than a base-rate predictor.
On RTP-LX, skill runs from \skSolarRtplx{} (Solar) to \skClefFlashRtplx{} (Clef-flash) and is significantly below zero for Jev, Kev-4B, Decision-1 and Solar.
Jev, second on \coretwelve{}, scores \skJevBrighter{} on BRIGHTER and beats only Solar. Kev-4B, Luna and Decision-1 also score below the base rate.
Ordinal scoring also reorders models.
On RTP-LX, Brier-skill and ranked-probability-skill rankings agree at $\tau_b=\tauRpsVsBrierRtplx$, and Luna moves from fifth to second (Appendix~\ref{app:full}).

\paragraph{Choosing a model on English.}
If we pick the best English model on each parallel suite from the same test data, another model beats it in \regEnWrong{} of \regEnUnits{} suite--language cells, significantly so in \regEnWrongPaired{} (Appendix~\ref{app:full}).
On Belebele, Decision-1 is best in English and loses significantly in \regBbPaired{} of \regBbUnits{} languages (Figure~\ref{fig:rq1}), with the largest loss in Burmese, where it ranks seventh.
Picking the best model on \coretwelve{} gives pplx-decider, which is best on only \regCoreRightSuites{} of the \regCoreOtherSuites{} other suites and significantly beaten on Belebele, INCLUDE and BRIGHTER.
On the \fresh{}, the English pick is best in all \regFreshUnits{} languages.
There, the ranking of the \nFreshRanked{} Typed Decision Models agrees with \coretwelve{} at Kendall $\tau=\tauFreshVsCore$ (exploratory, Appendix~\ref{app:fresh}), so new languages on the same datasets reorder the models much less than new suites do.

\paragraph{Label descriptions and native items.}
One-line English label descriptions raise MASSIVE skill by \descMassiveUpLo{} to \descMassiveUpHi{} for \descMassiveUpModels{} models but barely change SIB-200.
On the \natLangs{} languages INCLUDE shares with MMLU-ProX, Luna's native-exam skill exceeds its translated-exam skill by \natLunaNativeOverTranslated{}, while Jev's is \natJevTranslatedOverNative{} lower (Appendix~\ref{app:nat}).
Neither diagnostic explains the ranking changes.

\paragraph{Interface language.}
In an exploratory ablation (Appendix~\ref{app:native}), \texttt{claude-opus-5.5} translated the instructions and option texts of \coretwelve{} XNLI and MASSIVE into \niLangs{} languages, and \texttt{gpt-6.1-sol} back-translated them blind, with no native-speaker check.
Native instructions lower pooled XNLI Brier skill by \niXnli{} \niXnliCI{}, by \niXnliLow{} in the low-web-share group and \niXnliHigh{} in the high one.
On MASSIVE, translating the texts changes nothing (\niMassive{} \niMassiveCI{}) if hosted models still see English option IDs.
With a fully native interface and neutral IDs, skill falls by \niMassiveFull{} across \niFullModels{} models.
Clef-flash's XNLI lead over the hosted Qwen3.5-9B readout shrinks by \niHeadShrink{} \niHeadShrinkCI{}, so the \S\ref{sec:res-a1} head gains hold under English schemas.

\subsection{Heads versus their own backbones (RQ2)}
\label{sec:res-a1}

\begin{figure}[t]
  \centering
  \includegraphics[width=\columnwidth]{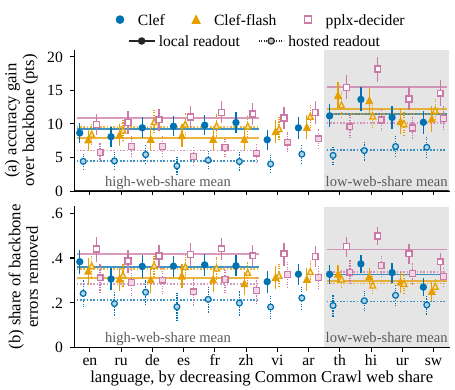}
  \caption{\textbf{Accuracy gains grow as web share falls.} Heads against their own backbones per \coretwelve{} language, by decreasing web share. (a)~Accuracy gain. (b)~The gain as a share of the backbone's errors, which barely changes under the local readout. Solid: local readout. Pale, dotted: hosted readout. Bars: 95\% paired intervals. Segments: group means. Shaded: low-web-share group.}
  \label{fig:a1}
\end{figure}

\paragraph{Accuracy and Brier score.}
Under the local readout, which H1 and H2 use, the heads win in all twelve languages on both metrics (Figures~\ref{fig:a1}a and~\ref{fig:a1-metrics}), by \aoneLocalAccGainAllLo{} to \aoneLocalAccGainAllHi{} accuracy points and \aoneLocalBrierGainAllLo{} to \aoneLocalBrierGainAllHi{} in Brier score.
The accuracy gain is significantly larger in the low-web-share group, by \aoneLocalClefAccInteraction{} points for Clef, \aoneLocalClefFlashAccInteraction{} for Clef-flash and \aoneLocalPplxAccInteraction{} for pplx-decider.
In Brier score the interaction is significantly positive for Clef (\aoneLocalClefBrierGainInteraction{}) and pplx-decider (\aoneLocalPplxBrierGainInteraction{}) but significantly negative for Clef-flash, \aoneLocalClefFlashBrierGainInteraction{} \aoneLocalClefFlashBrierGainInteractionCI{}.
On \bbLanguages{} Wide-Belebele languages, no head is significantly less accurate than its backbone (Appendix~\ref{app:bb}).
On \coretwelve{}, the heads remove similar shares of backbone errors in both groups (Figure~\ref{fig:a1}b), with \esLocalClefLow{} against \esLocalClefHigh{} for Clef, \esLocalClefFlashLow{} against \esLocalClefFlashHigh{} for Clef-flash and \esLocalPplxLow{} against \esLocalPplxHigh{} for pplx-decider.
Only Clef's difference is significant, and it favours the high-web-share group (\esLocalClefDiff{} \esLocalClefDiffCI{}).

\paragraph{Calibration.}
The heads have lower raw ECE than their backbones in all languages, by \aoneLocalEceGainAllLo{} to \aoneLocalEceGainAllHi{}.
With English-fitted temperatures for both arms (Appendix~\ref{app:temp}), the heads keep the lower Brier score in all \tempCells{} MASSIVE and XNLI pair--language cells and the lower ECE in \tempEceLower{}, significantly in \tempEceExcl{}.

\paragraph{Hosted backbone readout.}
Hosted readouts raise the backbones' \coretwelve{} skill by \skQwenNineHostedMinusLocal{} (9B) and \skQwenTwentySevenHostedMinusLocal{} (27B), but the heads still win in all twelve languages on both metrics.
The low-web-share group still gains significantly more accuracy, by \aoneHostedClefAccInteraction{} points for Clef, \aoneHostedClefFlashAccInteraction{} for Clef-flash and \aoneHostedPplxAccInteraction{} for pplx-decider.
The difference in errors removed is \esHostedClefDiff{} for Clef, \esHostedClefFlashDiff{} \esHostedClefFlashDiffCI{} for Clef-flash and \esHostedPplxDiff{} \esHostedPplxDiffCI{} for pplx-decider.
In an exploratory comparison outside H1 and H2, Decision-1 and its stated base Qwen3.5-9B also gain more accuracy in the low-web-share group under both readouts (Appendix~\ref{app:second}).
On Wide-Belebele, Decision-1's accuracy gain has an interval above zero in \srBbLocalAbove{} of \srBbLanguages{} languages under the local readout and below zero in none, with a median gain of \srBbLocalMedian{} points.

\paragraph{Held-out confirmation.}
On the \fresh{}, under the local readout, the pooled accuracy gain is \freshDOne{} points larger in the low-web-share group, confirming H1 (Holm-level bound \freshDOneBound{}, $p=\freshHOneP$).
The pooled difference in errors removed is \freshDTwo{}, with its \freshDTwoLevel\% interval \freshDTwoCI{} inside the $\pm\hTwoMargin$ margin.
H2 holds at Holm's second step ($\alpha=\freshHolmSecondLevel$, TOST $p=\freshHTwoP$) but not at the stricter first-step level of \freshHolmFirstLevel{}.
Under the exploratory hosted readout, H1 holds with a smaller contrast (\freshHostedDOne{} points, $p=\freshHostedHOneP$), but H2 does not (TOST $p=\freshHostedHTwoP$).

\subsection{Transfer of English-fitted confidence policies (RQ3)}
\label{sec:res-a3}

\begin{figure}[t]
  \centering
  \includegraphics[width=\columnwidth]{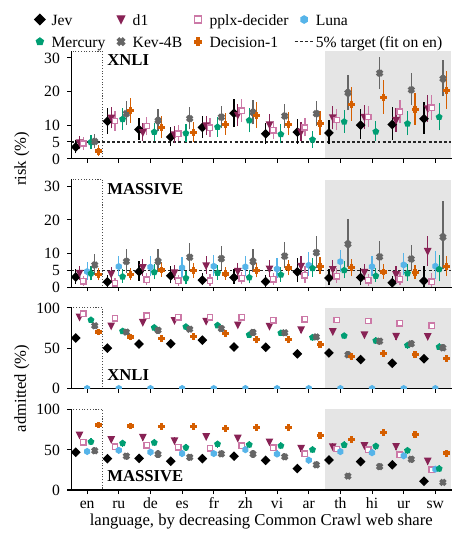}
  \caption{\textbf{English-fitted thresholds let through more errors on XNLI and fewer decisions on MASSIVE.} Error rate among admitted decisions (top) and admitted share (bottom) per \coretwelve{} language under an English-fitted 5\% threshold, for the hosted models we could assess. Luna admits no English XNLI decisions, so it has no XNLI risk. Dotted box: English. Shaded: low-web-share group. Bars: 95\% paired source-cluster intervals. Appendix~\ref{app:a3} lists the omitted cells.}
  \label{fig:a3}
\end{figure}

On English test data, the 5\% threshold keeps risk near target (Table~\ref{tab:a3}), at \aThreeXnliEnRiskLo--\aThreeXnliEnRiskHi\% on XNLI and \aThreeMassiveEnRiskLo--\aThreeMassiveEnRiskHi\% on MASSIVE.
Only Kev-4B exceeds 5\% on both, and Solar admits only \aThreeSolarXnliEnAdmitted\% on XNLI.

\paragraph{More admitted errors on XNLI.}
For all six assessable models, risk rises significantly in German and French, by \aThreeXnliDeFrRiseLo--\aThreeXnliDeFrRiseHi{} points to \aThreeXnliDeFrRiskLo--\aThreeXnliDeFrRiskHi\% (Figure~\ref{fig:a3}).
Risk is higher still in the low-web-share group for all six models, and Kev-4B reaches a group mean of \aThreeXnliKevLowRisk\%.
Conformal coverage for these six models is \aThreeXnliEnCovLo--\aThreeXnliEnCovHi{} in English and drops to \aThreeXnliHighCovLo--\aThreeXnliHighCovHi{} in the high-web-share group and \aThreeXnliLowCovLo--\aThreeXnliLowCovHi{} in the low-web-share one, significantly below target.
Luna keeps coverage at \atLunaXnliLowCov--\atLunaXnliEnCov{} by returning larger sets.
On the \fresh{}, the same thresholds give \freshAThreeXnliDeFrRiskLo--\freshAThreeXnliDeFrRiskHi\% risk in German and French in an exploratory analysis (Appendix~\ref{app:fresh}).

\paragraph{Lower admitted share on MASSIVE.}
On MASSIVE, risk stays near the target, but the models admit fewer decisions.
No model's mean risk changes significantly in high-web-share languages, though d1 in French (\aThreeMassiveLiquidFrRise{} points \aThreeMassiveLiquidFrRiseCI{}) and Decision-1 in Spanish rise significantly.
Outside English, risk exceeds the target by at most \aThreeMassiveExcessSmallMax{} points for Jev, pplx-decider, Mercury and Decision-1, and by \aThreeMassiveExcessLuna{} for Luna.
The largest excesses are for d1 and Kev-4B in Swahili, at \aThreeMassiveExcessLiquid{} and \aThreeMassiveExcessKev{} points.
In the low-web-share group, admitted share is \aThreeMassiveLowAdmDropLuna{} (Luna) to \aThreeMassiveLowAdmDropKev{} (Kev-4B) points below English, and Jev admits \aThreeMassiveJevSwAdmitted\% of Swahili decisions against \aThreeMassiveJevEnAdmitted\% of English ones.

\section{Conclusion}
\label{sec:conclusion}
On \xdb{}, rankings change from suite to suite. In hindsight, the best model in English loses significantly in one cell of four, so neither a \coretwelve{} nor an English leaderboard reliably ranks models for deployment.
Three open heads beat their Qwen backbones in all development languages and gain more where web text is scarce. Under the local readout, the held-out cohort confirms the larger gain, though the heads fix about the same fraction of their backbones' errors in both groups.
English-fitted thresholds admit more errors on XNLI, even in German and French, and fewer decisions on MASSIVE.
Practitioners should pick a model on suites that match their decision type and setting, and fit its confidence policy on target-language data.
\Afterreview{} will release the manifests, scorer and run receipts\releaseat{} so that new models can be scored on the same cases.

\section*{Limitations}
\label{sec:limitations}

\paragraph{Scope of the evidence.}
The head--backbone and transfer analyses use \coretwelve{}, the development cohort.
We repeat both on the \fresh{}, but the repeated transfer analysis is exploratory.
The low-web-share group of the \fresh{} has four languages, and only Swahili lies below Joshi tier 3.
The \fresh{} and the other suites are held out from our development analyses. But all sources are public, and some are well-known evaluation sets, so they may appear on the web or in vendors' training data.

\paragraph{Head--backbone contrast.}
Head training data are undisclosed, so we cannot say which part of a gain comes from decision training.
The local backbone readout is a length-normalised likelihood projection under one frozen prompt, and the hosted readout reads label tokens from a truncated top-20 list.
Other prompts or readouts could make backbones stronger or weaker and change both the absolute gains and the shares of backbone errors removed.
We have only three pre-registered pairs, plus the exploratory Decision-1--Qwen3.5-9B pair. Two of the three share the 27B backbone, and pplx-decider's base is stated by its vendor.
The comparison of shares concerns group averages rather than individual languages.

\paragraph{Translation, schema and web share.}
Most aligned suites translate or localize English material, and \coretwelve{} requests pair target-language content with English instructions and options.
Our language effects therefore combine the effect of target-language content under an English schema with translation artefacts.
In an exploratory ablation (Appendix~\ref{app:native}), native instructions lower XNLI skill most in low-web-share languages, so the English schema may understate those gaps.
Web share is confounded with script, language family, source quality and translation process.
The XNLI--MASSIVE contrast cannot separate task from translation.
RTP-LX gaps also include differences between local annotator pools.

\paragraph{Hosted models are snapshots.}
Headline API scores come from one pass over the cases, against snapshots that vendors may retire.
We repeated the \retestCases{} \coretwelve{} MASSIVE and SIB-200 cases on the same day.
The repeat keeps the top answer for \retestTopOneLo{} to \retestTopOneHi\% of cases per model and moves dataset Brier skill by at most \retestMaxBrierChange{}.
Where we bracketed a run with probe requests, their probabilities drift by at most \retestSentinelMaxDp{} between its start and end (Appendix~\ref{app:compute}).
We did not repeat the other eight suites, and the hosted Clef routes name no checkpoint.
Vendors also round probabilities to different precisions. Brier skill limits the damage, but rounding still discards information.

\paragraph{HateCheck measures specific weaknesses.}
HateCheck is a functional test suite rather than a sample of real traffic, so it says little about quality in deployment.

\section*{Ethics Statement}
\label{sec:ethics}

\paragraph{Data and licences.}
The headline suites reuse existing public datasets and gold labels under their licences (Appendix~\ref{app:data}).
For the exploratory interface-language ablation, an LLM translated the instruction and option texts (Appendix~\ref{app:native}).
We treat XNLI, AfriXNLI, AmericasNLI and RTP-LX as noncommercial.
Belebele, SIB-200, AmericasNLI, Global PIQA and parts of RTP-LX are ShareAlike.
Global PIQA forbids training, so we never use it for fitting.
InjongoIntent carries conflicting licence declarations, none of them noncommercial.
The benchmark as a whole is for research use.
\Afterreview{} will release manifests (identifiers and hashes), adapters, the scorer and aggregate results\releaseat.
We will not release source text, which users can obtain from the original distributors.
The dataset authors built the datasets with their own annotator populations.
We recruited no annotators, translators or human participants.
The English described-label texts are new, and we wrote them from the label names.

\paragraph{Harmful content.}
RTP-LX, Multilingual HateCheck and Aya red-teaming contain toxic, hateful or harmful text by design.
RTP-LX's terms forbid harmful or surveillance use.
We send this text only to Typed Decision Models for scoring and release no payloads.
The RTP-LX archive stays behind its upstream password.
A high moderation skill score does not show that a model is fit for content moderation in any language.

\paragraph{Model access and provider terms.}
We accessed hosted models through a commercial decision router under the terms of the providers.
We sent only public benchmark text and never sent gold labels.
We ran the local models (Clef, Clef-flash, Laya and the Qwen backbones) from pinned public checkpoints under their licences.

\paragraph{Intended use.}
Passing a benchmark does not replace validation of a model or a confidence policy on target-language deployment data.
On \coretwelve{} XNLI, a 5\% threshold fitted in English admits \aThreeXnliDeFrRiskLo{} to \aThreeXnliDeFrRiskHi\% errors in German and French.

\paragraph{Compute.}
Hosted API spend was \$\costApiTotal{} over all runs.
Local inference used \costHHundredHours{} H100 GPU-hours and an estimated \costRtxHours{} RTX 4090 hours.
The MediaPipe arms used an estimated \costCpuHours{} CPU wall-clock hours (Appendix~\ref{app:compute}, Table~\ref{tab:cost}).

\paragraph{Use of generative AI.}
Generative AI tools assisted at several stages of preparing this paper, in addition to the translations described in Appendix~\ref{app:native}.
The authors checked all outputs and take full responsibility for the content, including any errors.

\bibliography{references}

\begin{thebibliography}{67}
\providecommand{\natexlab}[1]{#1}

\bibitem[{Aakanksha et~al.(2024)Aakanksha, Ahmadian, Ermis, Goldfarb-Tarrant, Kreutzer, Fadaee, and Hooker}]{aakanksha2024prism}
Aakanksha, Arash Ahmadian, Beyza Ermis, Seraphina Goldfarb-Tarrant, Julia Kreutzer, Marzieh Fadaee, and Sara Hooker. 2024.
\newblock \href {https://doi.org/10.18653/v1/2024.emnlp-main.671} {The multilingual alignment prism: Aligning global and local preferences to reduce harm}.
\newblock In \emph{Proceedings of the 2024 Conference on Empirical Methods in Natural Language Processing}, pages 12027--12049, Miami, Florida, USA. Association for Computational Linguistics.

\bibitem[{Adelani et~al.(2024)Adelani, Liu, Shen, Vassilyev, Alabi, Mao, Gao, and Lee}]{adelani2024sib200}
David~Ifeoluwa Adelani, Hannah Liu, Xiaoyu Shen, Nikita Vassilyev, Jesujoba~O. Alabi, Yanke Mao, Haonan Gao, and En-Shiun~Annie Lee. 2024.
\newblock \href {https://doi.org/10.18653/v1/2024.eacl-long.14} {{SIB}-200: A simple, inclusive, and big evaluation dataset for topic classification in 200+ languages and dialects}.
\newblock In \emph{Proceedings of the 18th Conference of the European Chapter of the Association for Computational Linguistics (Volume 1: Long Papers)}, pages 226--245, St. Julian{'}s, Malta. Association for Computational Linguistics.

\bibitem[{Adelani et~al.(2025)Adelani, Ojo, Azime et~al.}]{adelani2025irokobench}
David~Ifeoluwa Adelani, Jessica Ojo, Israel~Abebe Azime, et~al. 2025.
\newblock \href {https://doi.org/10.18653/v1/2025.naacl-long.139} {{I}roko{B}ench: A new benchmark for {A}frican languages in the age of large language models}.
\newblock In \emph{Proceedings of the 2025 Conference of the Nations of the Americas Chapter of the Association for Computational Linguistics: Human Language Technologies (Volume 1: Long Papers)}, pages 2732--2757, Albuquerque, New Mexico. Association for Computational Linguistics.

\bibitem[{Ahuja et~al.(2022)Ahuja, Sitaram, Dandapat, and Choudhury}]{ahuja2022calibration}
Kabir Ahuja, Sunayana Sitaram, Sandipan Dandapat, and Monojit Choudhury. 2022.
\newblock \href {https://doi.org/10.18653/v1/2022.emnlp-main.290} {On the calibration of massively multilingual language models}.
\newblock In \emph{Proceedings of the 2022 Conference on Empirical Methods in Natural Language Processing}, pages 4310--4323, Abu Dhabi, United Arab Emirates. Association for Computational Linguistics.

\bibitem[{Angelopoulos and Bates(2023)}]{angelopoulos2023conformal}
Anastasios~N. Angelopoulos and Stephen Bates. 2023.
\newblock \href {https://doi.org/10.1561/2200000101} {Conformal prediction: A gentle introduction}.
\newblock \emph{Foundations and Trends in Machine Learning}, 16(4):494--591.

\bibitem[{{apolinario}(2026)}]{decisionindex2026}
{apolinario}. 2026.
\newblock Decision index: Reproduce the {Jev} decision-model benchmark suite locally or as one {HF} job.
\newblock \url{https://github.com/apolinario/decision-index}.
\newblock Edition 0.2.1; methodology at \url{https://huggingface.co/spaces/multimodalart/jev-decision-index}, revision cdbd1cab. Accessed 2026-10-08.

\bibitem[{Artetxe et~al.(2020)Artetxe, Labaka, and Agirre}]{artetxe2020translation}
Mikel Artetxe, Gorka Labaka, and Eneko Agirre. 2020.
\newblock \href {https://doi.org/10.18653/v1/2020.emnlp-main.618} {Translation artifacts in cross-lingual transfer learning}.
\newblock In \emph{Proceedings of the 2020 Conference on Empirical Methods in Natural Language Processing (EMNLP)}, pages 7674--7684, Online. Association for Computational Linguistics.

\bibitem[{Bandarkar et~al.(2024)Bandarkar, Liang, Muller, Artetxe, Shukla, Husa, Goyal, Krishnan, Zettlemoyer, and Khabsa}]{bandarkar2024belebele}
Lucas Bandarkar, Davis Liang, Benjamin Muller, Mikel Artetxe, Satya~Narayan Shukla, Donald Husa, Naman Goyal, Abhinandan Krishnan, Luke Zettlemoyer, and Madian Khabsa. 2024.
\newblock \href {https://doi.org/10.18653/v1/2024.acl-long.44} {The belebele benchmark: a parallel reading comprehension dataset in 122 language variants}.
\newblock In \emph{Proceedings of the 62nd Annual Meeting of the Association for Computational Linguistics (Volume 1: Long Papers)}, pages 749--775, Bangkok, Thailand. Association for Computational Linguistics.

\bibitem[{Bastianelli et~al.(2020)Bastianelli, Vanzo, Swietojanski, and Rieser}]{bastianelli2020slurp}
Emanuele Bastianelli, Andrea Vanzo, Pawel Swietojanski, and Verena Rieser. 2020.
\newblock \href {https://doi.org/10.18653/v1/2020.emnlp-main.588} {{SLURP}: A spoken language understanding resource package}.
\newblock In \emph{Proceedings of the 2020 Conference on Empirical Methods in Natural Language Processing (EMNLP)}, pages 7252--7262, Online. Association for Computational Linguistics.

\bibitem[{{Benchmark Heaven}(2026)}]{jevbench2026}
{Benchmark Heaven}. 2026.
\newblock {JevBench}: A benchmark for {Jev}-class typed decision models.
\newblock \url{https://github.com/fstandhartinger/jevbench}.
\newblock V1.5 frozen method: \url{https://github.com/fstandhartinger/jevbench/blob/main/docs/METHOD-v1.5.md}. Accessed 2026-10-08.

\bibitem[{Chang et~al.(2025)Chang, Arnett, Sadallah et~al.}]{chang2025globalpiqa}
Tyler~A. Chang, Catherine Arnett, Abdelrahman Sadallah, et~al. 2025.
\newblock \href {https://arxiv.org/abs/2510.24081} {Global {PIQA}: Evaluating commonsense reasoning across 100+ languages and cultures}.
\newblock \url{https://arxiv.org/abs/2510.24081}.
\newblock \emph{Preprint}, arXiv:2510.24081.

\bibitem[{{Cloudflare}(2026)}]{cloudflare2026clef}
{Cloudflare}. 2026.
\newblock Introducing {Clef}: {Cloudflare}'s first open-source decision models, now on {Workers AI}.
\newblock \url{https://developers.cloudflare.com/changelog/post/2026-10-01-clef-workers-ai/}.
\newblock Changelog, 1 October 2026; model cards \url{https://huggingface.co/Cloudflare/clef} and \url{https://huggingface.co/Cloudflare/clef-flash}. Accessed 2026-10-08.

\bibitem[{{Common Crawl}(2026)}]{commoncrawl2026languages}
{Common Crawl}. 2026.
\newblock Statistics of {Common Crawl} monthly archives: Distribution of languages.
\newblock \url{https://commoncrawl.github.io/cc-crawl-statistics/plots/languages}.
\newblock Crawl CC-MAIN-2026-39; primary document language by CLD2. Accessed 2026-10-08.

\bibitem[{Conneau et~al.(2022)Conneau, Bapna, Zhang et~al.}]{conneau2022xtremes}
Alexis Conneau, Ankur Bapna, Yu~Zhang, et~al. 2022.
\newblock \href {https://arxiv.org/abs/2203.10752} {{XTREME-S}: Evaluating cross-lingual speech representations}.
\newblock In \emph{Proc. Interspeech 2022}, pages 3248--3252.

\bibitem[{Conneau et~al.(2018)Conneau, Rinott, Lample, Williams, Bowman, Schwenk, and Stoyanov}]{conneau2018xnli}
Alexis Conneau, Ruty Rinott, Guillaume Lample, Adina Williams, Samuel~R. Bowman, Holger Schwenk, and Veselin Stoyanov. 2018.
\newblock \href {https://doi.org/10.18653/v1/D18-1269} {{XNLI}: Evaluating cross-lingual sentence representations}.
\newblock In \emph{Proceedings of the 2018 Conference on Empirical Methods in Natural Language Processing}, pages 2475--2485, Brussels, Belgium. Association for Computational Linguistics.

\bibitem[{{Convai Innovations}(2026)}]{convai2026laya}
{Convai Innovations}. 2026.
\newblock {Laya} model card.
\newblock \url{https://huggingface.co/convaiinnovations/laya}.
\newblock Benchmark report: \url{https://github.com/NandhaKishorM/laya/blob/main/BENCHMARKS.md}. Accessed 2026-10-08.

\bibitem[{de~Wynter et~al.(2025)de~Wynter, Watts, Wongsangaroonsri et~al.}]{dewynter2025rtplx}
Adrian de~Wynter, Ishaan Watts, Tua Wongsangaroonsri, et~al. 2025.
\newblock \href {https://doi.org/10.1609/aaai.v39i27.35011} {{RTP-LX}: Can {LLMs} evaluate toxicity in multilingual scenarios?}
\newblock In \emph{Proceedings of the AAAI Conference on Artificial Intelligence}.

\bibitem[{Deu{\ss}er et~al.(2026)Deu{\ss}er, Ramamurthy, Sparrenberg, and Sifa}]{deusser2026jev}
Tobias Deu{\ss}er, Rajkumar Ramamurthy, Lorenz Sparrenberg, and Rafet Sifa. 2026.
\newblock \href {https://arxiv.org/abs/2609.37647} {Evaluating and benchmarking the system one model {Jev}}.
\newblock \url{https://arxiv.org/abs/2609.37647}.
\newblock \emph{Preprint}, arXiv:2609.37647.
\newblock ArXiv preprint, v1, 29 September 2026.

\bibitem[{Ebrahimi et~al.(2022)Ebrahimi, Mager, Oncevay et~al.}]{ebrahimi2022americasnli}
Abteen Ebrahimi, Manuel Mager, Arturo Oncevay, et~al. 2022.
\newblock \href {https://doi.org/10.18653/v1/2022.acl-long.435} {{A}mericas{NLI}: Evaluating zero-shot natural language understanding of pretrained multilingual models in truly low-resource languages}.
\newblock In \emph{Proceedings of the 60th Annual Meeting of the Association for Computational Linguistics (Volume 1: Long Papers)}, pages 6279--6299, Dublin, Ireland. Association for Computational Linguistics.

\bibitem[{Enevoldsen et~al.(2025)Enevoldsen, Chung, Kerboua et~al.}]{enevoldsen2025mmteb}
Kenneth Enevoldsen, Isaac Chung, Imene Kerboua, et~al. 2025.
\newblock \href {https://arxiv.org/abs/2502.13595} {{MMTEB}: Massive multilingual text embedding benchmark}.
\newblock In \emph{International Conference on Learning Representations (ICLR)}.

\bibitem[{FitzGerald et~al.(2023)FitzGerald, Hench, Peris, Mackie, Rottmann, Sanchez, Nash, Urbach, Kakarala, Singh, Ranganath, Crist, Britan, Leeuwis, Tur, and Natarajan}]{fitzgerald2023massive}
Jack FitzGerald, Christopher Hench, Charith Peris, Scott Mackie, Kay Rottmann, Ana Sanchez, Aaron Nash, Liam Urbach, Vishesh Kakarala, Richa Singh, Swetha Ranganath, Laurie Crist, Misha Britan, Wouter Leeuwis, Gokhan Tur, and Prem Natarajan. 2023.
\newblock \href {https://doi.org/10.18653/v1/2023.acl-long.235} {{MASSIVE}: A 1{M}-example multilingual natural language understanding dataset with 51 typologically-diverse languages}.
\newblock In \emph{Proceedings of the 61st Annual Meeting of the Association for Computational Linguistics (Volume 1: Long Papers)}, pages 4277--4302, Toronto, Canada. Association for Computational Linguistics.

\bibitem[{Gan et~al.(2026)Gan, Lee, and Kang}]{gan2026sansi}
Shuyu Gan, Young-Jun Lee, and Dongyeop Kang. 2026.
\newblock \href {https://arxiv.org/abs/2610.07730} {{SanSi}: A looped typed decision model for system 1.5 thinking}.
\newblock \url{https://arxiv.org/abs/2610.07730}.
\newblock \emph{Preprint}, arXiv:2610.07730.
\newblock ArXiv preprint.

\bibitem[{Geifman and El-Yaniv(2017)}]{geifman2017selective}
Yonatan Geifman and Ran El-Yaniv. 2017.
\newblock \href {https://papers.nips.cc/paper/2017/hash/4a8423d5e91fda00bb7e46540e2b0cf1-Abstract.html} {Selective classification for deep neural networks}.
\newblock In \emph{Advances in Neural Information Processing Systems 30}, pages 4878--4887. Curran Associates, Inc.

\bibitem[{Gneiting and Raftery(2007)}]{gneiting2007strictly}
Tilmann Gneiting and Adrian~E. Raftery. 2007.
\newblock \href {https://doi.org/10.1198/016214506000001437} {Strictly proper scoring rules, prediction, and estimation}.
\newblock \emph{Journal of the American Statistical Association}, 102(477):359--378.

\bibitem[{Han et~al.(2026)Han, Zhang, Chen, Binbinliu, Pechenizkiy, Fang, and Zheng}]{han2025mubench}
Wenhan Han, Yifan Zhang, Zhixun Chen, Binbinliu, Mykola Pechenizkiy, Meng Fang, and Yin Zheng. 2026.
\newblock \href {https://doi.org/10.18653/v1/2026.findings-acl.794} {{M}u{B}ench: Assessment of multilingual capabilities of large language models across 61 languages}.
\newblock In \emph{Findings of the {A}ssociation for {C}omputational {L}inguistics: {ACL} 2026}, pages 16163--16192, San Diego, California, United States. Association for Computational Linguistics.

\bibitem[{Himmi et~al.(2023)Himmi, Irurozki, Noiry, Cl{\'e}men{\c{c}}on, and Colombo}]{himmi2023missing}
Anas Himmi, Ekhine Irurozki, Nathan Noiry, Stephan Cl{\'e}men{\c{c}}on, and Pierre Colombo. 2023.
\newblock \href {https://arxiv.org/abs/2305.10284} {Towards more robust {NLP} system evaluation: Handling missing scores in benchmarks}.
\newblock \url{https://arxiv.org/abs/2305.10284}.
\newblock \emph{Preprint}, arXiv:2305.10284.

\bibitem[{Holm(1979)}]{holm1979simple}
Sture Holm. 1979.
\newblock \href {https://doi.org/10.2307/4615733} {A simple sequentially rejective multiple test procedure}.
\newblock \emph{Scandinavian Journal of Statistics}, 6(2):65--70.

\bibitem[{Huang et~al.(2026)Huang, Lu, Zeng, Iwasawa, Matsuo, Chandar, Marrese-Taylor, and Li}]{huang2026eacl}
Jerry Huang, Peng Lu, Qiuhao Zeng, Yusuke Iwasawa, Yutaka Matsuo, Sarath Chandar, Edison Marrese-Taylor, and Irene Li. 2026.
\newblock \href {https://doi.org/10.18653/v1/2026.eacl-short.1} {Investigating the multilingual calibration effects of language model instruction tuning}.
\newblock In \emph{Proceedings of the 19th Conference of the {E}uropean Chapter of the {A}ssociation for {C}omputational {L}inguistics (Volume 2: Short Papers)}, pages 1--59, Rabat, Morocco. Association for Computational Linguistics.

\bibitem[{Huang et~al.(2025)Huang, Zhu, Hu, He, Li, Huang, and Yuan}]{huang2025benchmax}
Xu~Huang, Wenhao Zhu, Hanxu Hu, Conghui He, Lei Li, Shujian Huang, and Fei Yuan. 2025.
\newblock \href {https://doi.org/10.18653/v1/2025.findings-emnlp.909} {{BenchMAX}: A comprehensive multilingual evaluation suite for large language models}.
\newblock In \emph{Findings of the Association for Computational Linguistics: EMNLP 2025}, pages 16751--16774.

\bibitem[{Hupkes and Bogoychev(2025)}]{hupkes2025multiloko}
Dieuwke Hupkes and Nikolay Bogoychev. 2025.
\newblock \href {https://arxiv.org/abs/2504.10356} {{MultiLoKo}: a multilingual local knowledge benchmark for {LLMs} spanning 31 languages}.
\newblock \url{https://arxiv.org/abs/2504.10356}.
\newblock \emph{Preprint}, arXiv:2504.10356.

\bibitem[{{Inception}(2026)}]{inception2026mercury}
{Inception}. 2026.
\newblock Mercury decide.
\newblock \url{https://openrouter.ai/inception/mercury-decide}.
\newblock Structured decision model served as a System One endpoint; snapshot inception/mercury-decide-20260930. Accessed 2026-10-08.

\bibitem[{Jiang et~al.(2022)Jiang, Liu, and Van~Durme}]{jiang2022calibrating}
Zhengping Jiang, Anqi Liu, and Benjamin Van~Durme. 2022.
\newblock \href {https://doi.org/10.18653/v1/2022.emnlp-main.170} {Calibrating zero-shot cross-lingual (un-)structured predictions}.
\newblock In \emph{Proceedings of the 2022 Conference on Empirical Methods in Natural Language Processing}, pages 2648--2674, Abu Dhabi, United Arab Emirates. Association for Computational Linguistics.

\bibitem[{Joshi et~al.(2020)Joshi, Santy, Budhiraja, Bali, and Choudhury}]{joshi2020state}
Pratik Joshi, Sebastin Santy, Amar Budhiraja, Kalika Bali, and Monojit Choudhury. 2020.
\newblock \href {https://doi.org/10.18653/v1/2020.acl-main.560} {The state and fate of linguistic diversity and inclusion in the {NLP} world}.
\newblock In \emph{Proceedings of the 58th Annual Meeting of the Association for Computational Linguistics}, pages 6282--6293, Online. Association for Computational Linguistics.

\bibitem[{Li(2026)}]{li2026fastmodels}
Jiawei Li. 2026.
\newblock \href {https://arxiv.org/abs/2610.02267} {Fast models, slow evidence: A paired and self-audited evaluation of {System-1} decision models for {LLM} agent harnesses}.
\newblock \url{https://arxiv.org/abs/2610.02267}.
\newblock \emph{Preprint}, arXiv:2610.02267.
\newblock ArXiv preprint.

\bibitem[{Liang et~al.(2023)Liang, Bommasani, Lee et~al.}]{liang2023helm}
Percy Liang, Rishi Bommasani, Tony Lee, et~al. 2023.
\newblock \href {https://arxiv.org/abs/2211.09110} {Holistic evaluation of language models}.
\newblock \emph{Transactions on Machine Learning Research}.

\bibitem[{{Liquid AI}(2026)}]{liquid2026d1}
{Liquid AI}. 2026.
\newblock d1: {Liquid AI}'s structured decision model.
\newblock \url{https://openrouter.ai/liquid/d1}.
\newblock Served as a System One endpoint via OpenRouter; snapshot liquid/d1-20260930. Accessed 2026-10-08.

\bibitem[{Lu and Wu(2026)}]{lu2026rejection}
Jiawen Lu and Tongtong Wu. 2026.
\newblock \href {https://arxiv.org/abs/2610.03387} {Benchmarking candidate coverage and rejection policy transfer in typed decision models}.
\newblock \url{https://arxiv.org/abs/2610.03387}.
\newblock \emph{Preprint}, arXiv:2610.03387.
\newblock ArXiv preprint.

\bibitem[{{manjunathshiva}(2026)}]{opendecider2026}
{manjunathshiva}. 2026.
\newblock {OpenDecider}: Open, calibrated decision models that beat {Laya} and {Jev} on typed-decisions.
\newblock \url{https://huggingface.co/blog/manjunathshiva/opendecider-beats-laya-and-jev}.
\newblock Hugging Face community blog. Accessed 2026-10-08.

\bibitem[{Marone et~al.(2025)Marone, Weller, Fleshman, Yang, Lawrie, and Van~Durme}]{marone2025mmbert}
Marc Marone, Orion Weller, William Fleshman, Eugene Yang, Dawn Lawrie, and Benjamin Van~Durme. 2025.
\newblock \href {https://arxiv.org/abs/2509.06888} {{mmBERT}: A modern multilingual encoder with annealed language learning}.
\newblock \emph{Preprint}, arXiv:2509.06888.

\bibitem[{{Microsoft}(2026)}]{microsoft2026decision}
{Microsoft}. 2026.
\newblock Microsoft-decision-1: Our model for fast decision-making.
\newblock \url{https://commandline.microsoft.com/microsoft-decision-1-model-foundry/}.
\newblock Decision-scoring model in Microsoft Foundry and on OpenRouter; snapshot microsoft/microsoft-decision-1-20261009. Accessed 2026-10-10.

\bibitem[{Miller(2024)}]{miller2024errorbars}
Evan Miller. 2024.
\newblock \href {https://arxiv.org/abs/2411.00640} {Adding error bars to evals: A statistical approach to language model evaluations}.
\newblock \url{https://arxiv.org/abs/2411.00640}.
\newblock \emph{Preprint}, arXiv:2411.00640.

\bibitem[{Muhammad et~al.(2025)Muhammad, Ousidhoum, Abdulmumin et~al.}]{muhammad2025brighter}
Shamsuddeen~Hassan Muhammad, Nedjma Ousidhoum, Idris Abdulmumin, et~al. 2025.
\newblock \href {https://doi.org/10.18653/v1/2025.acl-long.436} {{BRIGHTER}: {BRI}dging the gap in human-annotated textual emotion recognition datasets for 28 languages}.
\newblock In \emph{Proceedings of the 63rd Annual Meeting of the Association for Computational Linguistics (Volume 1: Long Papers)}, pages 8895--8916, Vienna, Austria. Association for Computational Linguistics.

\bibitem[{Nixon et~al.(2019)Nixon, Dusenberry, Jerfel, Nguyen, Liu, Zhang, and Tran}]{nixon2019measuring}
Jeremy Nixon, Mike Dusenberry, Ghassen Jerfel, Timothy Nguyen, Jeremiah Liu, Linchuan Zhang, and Dustin Tran. 2019.
\newblock \href {https://arxiv.org/abs/1904.01685} {Measuring calibration in deep learning}.
\newblock \url{https://arxiv.org/abs/1904.01685}.
\newblock \emph{Preprint}, arXiv:1904.01685.
\newblock ArXiv preprint.

\bibitem[{Ojo et~al.(2025)Ojo, Ogundepo, Oladipo, Ogueji, Lin, Stenetorp, and Adelani}]{ojo2025afrobench}
Jessica Ojo, Odunayo Ogundepo, Akintunde Oladipo, Kelechi Ogueji, Jimmy Lin, Pontus Stenetorp, and David~Ifeoluwa Adelani. 2025.
\newblock \href {https://doi.org/10.18653/v1/2025.findings-acl.976} {{AfroBench}: How good are large language models on {African} languages?}
\newblock In \emph{Findings of the Association for Computational Linguistics: ACL 2025}, pages 19048--19095.

\bibitem[{{OpenAI}(2026)}]{openai2026decisions}
{OpenAI}. 2026.
\newblock Decisions {API} (public beta).
\newblock \url{https://developers.openai.com/api/docs/guides/decisions}.
\newblock Accessed 2026-10-08.

\bibitem[{Palmer(2026)}]{palmer2026kev}
Jared Palmer. 2026.
\newblock Kev: A {Jev}-like family of decision models built on top of {Qwen3.5/3.8}.
\newblock \url{https://github.com/jaredpalmer/kev}.
\newblock Kev 1.0 release; model cards at \url{https://huggingface.co/jaredpalmer/kev-0.8b}. Accessed 2026-10-08.

\bibitem[{{Perplexity AI}(2026)}]{perplexity2026pplx}
{Perplexity AI}. 2026.
\newblock Decider v1.1 27b (pplx-decider-v1.1-27b).
\newblock \url{https://openrouter.ai/perplexity/pplx-decider-v1.1-27b}.
\newblock Predecessor model card: \url{https://huggingface.co/perplexity-ai/pplx-decider-v1-27b}. Accessed 2026-10-08.

\bibitem[{Post(2018)}]{post2018sacrebleu}
Matt Post. 2018.
\newblock \href {https://doi.org/10.18653/v1/W18-6319} {A call for clarity in reporting {BLEU} scores}.
\newblock In \emph{Proceedings of the Third Conference on Machine Translation: Research Papers}, pages 186--191, Brussels, Belgium. Association for Computational Linguistics.

\bibitem[{{Qwen Team}(2026{\natexlab{a}})}]{qwen2026qwen35}
{Qwen Team}. 2026{\natexlab{a}}.
\newblock {Qwen3.5}: Towards native multimodal agents.
\newblock \url{https://qwen.ai/blog?id=qwen3.5}.
\newblock Model card: \url{https://huggingface.co/Qwen/Qwen3.5-9B}. Accessed 2026-10-08.

\bibitem[{{Qwen Team}(2026{\natexlab{b}})}]{qwen2026qwen38}
{Qwen Team}. 2026{\natexlab{b}}.
\newblock {Qwen3.8-Max}: A new bar for coding and cowork.
\newblock \url{https://qwen.ai/blog?id=qwen3.8}.
\newblock Model card: \url{https://huggingface.co/Qwen/Qwen3.8-27B}. Accessed 2026-10-08.

\bibitem[{Rafe and Das(2026)}]{rafe2026decision}
Amir Rafe and Subasish Das. 2026.
\newblock \href {https://arxiv.org/abs/2610.00346} {Benchmarking system one decision models against trained classifiers and language models for automated decision gates}.
\newblock \url{https://arxiv.org/abs/2610.00346}.
\newblock \emph{Preprint}, arXiv:2610.00346.
\newblock ArXiv preprint.

\bibitem[{Romanou et~al.(2024)Romanou, Foroutan, Sotnikova, Chen, Nelaturu, Singh, Maheshwary, Altomare, Haggag, A, Amayuelas, Amirudin, Aryabumi, Boiko, Chang, Chim, Cohen, Dalmia, Diress, Duwal, Dzenhaliou, Florez, Farestam, Imperial, Islam, Isotalo, Jabbarishiviari, Karlsson, Khalilov, Klamm, Koto, Krzemiński, de~Melo, Montariol, Nan, Niklaus, Novikova, Ceron, Paul, Ploeger, Purbey, Rajwal, Ravi, Rydell, Santhosh, Sharma, Skenduli, Moakhar, Moakhar, Tamir, Tarun, Wasi, Weerasinghe, Yilmaz, Zhang, Schlag, Fadaee, Hooker, and Bosselut}]{romanou2025include}
Angelika Romanou, Negar Foroutan, Anna Sotnikova, Zeming Chen, Sree~Harsha Nelaturu, Shivalika Singh, Rishabh Maheshwary, Micol Altomare, Mohamed~A. Haggag, Snegha A, Alfonso Amayuelas, Azril~Hafizi Amirudin, Viraat Aryabumi, Danylo Boiko, Michael Chang, Jenny Chim, Gal Cohen, Aditya~Kumar Dalmia, Abraham Diress, and 40 others. 2024.
\newblock \href {https://arxiv.org/abs/2411.19799} {{INCLUDE}: Evaluating multilingual language understanding with regional knowledge}.
\newblock \emph{Preprint}, arXiv:2411.19799.

\bibitem[{R{\"o}ttger et~al.(2022)R{\"o}ttger, Seelawi, Nozza, Talat, and Vidgen}]{rottger2022mhc}
Paul R{\"o}ttger, Haitham Seelawi, Debora Nozza, Zeerak Talat, and Bertie Vidgen. 2022.
\newblock \href {https://doi.org/10.18653/v1/2022.woah-1.15} {Multilingual {H}ate{C}heck: Functional tests for multilingual hate speech detection models}.
\newblock In \emph{Proceedings of the Sixth Workshop on Online Abuse and Harms (WOAH)}, pages 154--169, Seattle, Washington (Hybrid). Association for Computational Linguistics.

\bibitem[{Sadinle et~al.(2019)Sadinle, Lei, and Wasserman}]{sadinle2019least}
Mauricio Sadinle, Jing Lei, and Larry Wasserman. 2019.
\newblock \href {https://doi.org/10.1080/01621459.2017.1395341} {Least ambiguous set-valued classifiers with bounded error levels}.
\newblock \emph{Journal of the American Statistical Association}, 114(525):223--234.

\bibitem[{Sharma et~al.(2026)Sharma, Duncan, Marceau~Caron, and Vollmer}]{sharma2026typedbench}
Rahul Sharma, Andrew~B. Duncan, Ga{\'e}tan Marceau~Caron, and Sebastian~J. Vollmer. 2026.
\newblock \href {https://arxiv.org/abs/2610.11392} {{TypedBench}: A benchmark for calibration, framing sensitivity, and cost in system one decision models}.
\newblock \url{https://arxiv.org/abs/2610.11392}.
\newblock \emph{Preprint}, arXiv:2610.11392.
\newblock ArXiv preprint.

\bibitem[{Singh et~al.(2025)Singh, Romanou, Fourrier et~al.}]{singh2025globalmmlu}
Shivalika Singh, Angelika Romanou, Cl{\'e}mentine Fourrier, et~al. 2025.
\newblock \href {https://doi.org/10.18653/v1/2025.acl-long.919} {Global {MMLU}: Understanding and addressing cultural and linguistic biases in multilingual evaluation}.
\newblock In \emph{Proceedings of the 63rd Annual Meeting of the Association for Computational Linguistics (Volume 1: Long Papers)}, pages 18761--18799, Vienna, Austria. Association for Computational Linguistics.

\bibitem[{Susanto et~al.(2025)Susanto, Hulagadri, Montalan, Ngui, Yong, Leong, Rengarajan, Limkonchotiwat, Mai, and Tjhi}]{susanto2025seahelm}
Yosephine Susanto, Adithya~Venkatadri Hulagadri, Jann~Railey Montalan, Jian~Gang Ngui, Xianbin Yong, Wei~Qi Leong, Hamsawardhini Rengarajan, Peerat Limkonchotiwat, Yifan Mai, and William~Chandra Tjhi. 2025.
\newblock \href {https://doi.org/10.18653/v1/2025.findings-acl.636} {{SEA}-{HELM}: {S}outheast {A}sian holistic evaluation of language models}.
\newblock In \emph{Findings of the Association for Computational Linguistics: ACL 2025}, pages 12308--12336, Vienna, Austria. Association for Computational Linguistics.

\bibitem[{Tateno(2026)}]{tateno2026s1mb}
Tateno. 2026.
\newblock {S1MB}: Comparing system one decision models across 100+ benchmarks.
\newblock \url{https://huggingface.co/blog/hotchpotch/system-one-mosaic-benchmark}.
\newblock System One Mosaic Benchmark; code at \url{https://github.com/hotchpotch/S1MB}. Accessed 2026-10-08.

\bibitem[{{TypeSafe AI}(2026)}]{typesafe2026jev}
{TypeSafe AI}. 2026.
\newblock Introducing system one models \& {Jev}.
\newblock \url{https://typesafe.ai/blog/introducing-system-one-models-and-jev}.
\newblock TypeSafe AI Blog; API docs at \url{https://docs.typesafe.ai/api.md}. Accessed 2026-10-08.

\bibitem[{{Upstage}(2026)}]{upstage2026solar}
{Upstage}. 2026.
\newblock Solar decide.
\newblock \url{https://openrouter.ai/upstage/solar-decide}.
\newblock Structured decision model served as a System One endpoint; snapshot upstage/solar-decide-20260928. Accessed 2026-10-08.

\bibitem[{Williams et~al.(2018)Williams, Nangia, and Bowman}]{williams2018mnli}
Adina Williams, Nikita Nangia, and Samuel~R. Bowman. 2018.
\newblock \href {https://doi.org/10.18653/v1/N18-1101} {A broad-coverage challenge corpus for sentence understanding through inference}.
\newblock In \emph{Proceedings of the 2018 Conference of the North {A}merican Chapter of the Association for Computational Linguistics: Human Language Technologies, Volume 1 (Long Papers)}, pages 1112--1122, New Orleans, Louisiana. Association for Computational Linguistics.

\bibitem[{Xuan et~al.(2025)Xuan, Yang, Qi, Zeng, Xiao, Feng, Liu, Xing, Wang, Gao, Lu, Jiang, Li, Li, Yu, Dong, Gu, Li, Xie, Juefei-Xu, Khomh, Yoshie, Chen, Teodoro, Liu, Goebel, Ma, Marrese-Taylor, Lu, Iwasawa, Matsuo, and Li}]{xuan2025mmluprox}
Weihao Xuan, Rui Yang, Heli Qi, Qingcheng Zeng, Yunze Xiao, Aosong Feng, Dairui Liu, Yun Xing, Junjue Wang, Fan Gao, Jinghui Lu, Yuang Jiang, Huitao Li, Xin Li, Kunyu Yu, Ruihai Dong, Shangding Gu, Yuekang Li, Xiaofei Xie, and 13 others. 2025.
\newblock \href {https://arxiv.org/abs/2503.10497} {{MMLU-ProX}: A multilingual benchmark for advanced large language model evaluation}.
\newblock \emph{Preprint}, arXiv:2503.10497.

\bibitem[{Xue et~al.(2025)Xue, Wang, Wang, Wang, Wang, Du, Liang, Zhang, and Wong}]{xue2025mlingconf}
Boyang Xue, Hongru Wang, Rui Wang, Sheng Wang, Zezhong Wang, Yiming Du, Bin Liang, Wenxuan Zhang, and Kam-Fai Wong. 2025.
\newblock \href {https://doi.org/10.18653/v1/2025.findings-acl.129} {{MlingConf}: A comprehensive study of multilingual confidence estimation on large language models}.
\newblock In \emph{Findings of the Association for Computational Linguistics: ACL 2025}, pages 2535--2556.

\bibitem[{Yang et~al.(2024)Yang, Dan, Roth, and Lee}]{yang2024calibration}
Yahan Yang, Soham Dan, Dan Roth, and Insup Lee. 2024.
\newblock \href {https://arxiv.org/abs/2311.08669} {On the calibration of multilingual question answering {LLM}s}.
\newblock \url{https://arxiv.org/abs/2311.08669v2}.
\newblock \emph{Preprint}, arXiv:2311.08669.
\newblock ArXiv preprint, v2, 15 April 2024.

\bibitem[{{yanng981}(2026)}]{yanng2026systemone}
{yanng981}. 2026.
\newblock system-one-benchmark: Zero-shot accuracy, calibration and latency of system one decision models.
\newblock \url{https://github.com/yanng981/system-one-benchmark}.
\newblock GitHub repository. Accessed 2026-10-08.

\bibitem[{Yu et~al.(2025)Yu, Alabi, Bukula et~al.}]{yu2025injongo}
Hao Yu, Jesujoba~Oluwadara Alabi, Andiswa Bukula, et~al. 2025.
\newblock \href {https://doi.org/10.18653/v1/2025.acl-long.464} {{INJONGO}: A multicultural intent detection and slot-filling dataset for 16 {A}frican languages}.
\newblock In \emph{Proceedings of the 63rd Annual Meeting of the Association for Computational Linguistics (Volume 1: Long Papers)}, pages 9429--9452, Vienna, Austria. Association for Computational Linguistics.

\bibitem[{Zhou et~al.(2025)Zhou, Zhang, Hu, Li, Collier, Vuli{\'c}, and Korhonen}]{zhou2025multilingual}
Ej~Zhou, Caiqi Zhang, Tiancheng Hu, Chengzu Li, Nigel Collier, Ivan Vuli{\'c}, and Anna Korhonen. 2025.
\newblock \href {https://arxiv.org/abs/2510.03136} {Beyond the final layer: Intermediate representations for better multilingual calibration in large language models}.
\newblock \url{https://arxiv.org/abs/2510.03136}.
\newblock \emph{Preprint}, arXiv:2510.03136.
\newblock ArXiv preprint.

\end{thebibliography}

\appendix
\makeatletter
\setlength{\@fptop}{0pt}\setlength{\@fpsep}{16pt}\setlength{\@fpbot}{0pt plus 1fil}
\setlength{\@dblfptop}{0pt}\setlength{\@dblfpsep}{16pt}\setlength{\@dblfpbot}{0pt plus 1fil}
\makeatother
\renewcommand{\dblfloatpagefraction}{0.6}
\setcounter{dbltopnumber}{3}\renewcommand{\dbltopfraction}{0.9}
\setcounter{topnumber}{3}\renewcommand{\topfraction}{0.9}\renewcommand{\textfraction}{0.1}

\section*{Appendix Conventions}
In tables, n.c.\ means not comparable (the caption gives the reason), n.r.\ not run and inc.\ incomplete. Refusals are never excluded and count as failures.
Every number in the appendix comes from an analysis report that we will release with the code\afterreview.

\section{Data and Suites}
\label{app:data}

\subsection{Sources and Licences}

\begin{table*}[t]
  \centering
  \footnotesize
  \setlength{\tabcolsep}{4pt}
  \renewcommand{\arraystretch}{1.15}
  \begin{tabular}{@{}l l >{\raggedright\arraybackslash}p{4.3cm} >{\raggedright\arraybackslash}p{5.4cm}@{}}
    \toprule
    Source & Revision & Licence (XDecideBench treatment) & Provenance of non-English text \\
    \midrule
    MASSIVE 1.1 & \texttt{ff6bd8e} & CC-BY-4.0 & Localized from English SLURP utterances \\
    XNLI & \texttt{b8dd5d7} & CC-BY-NC-4.0 & Translated MultiNLI-style pairs; labels propagated \\
    Belebele & \texttt{7899cdf} & CC-BY-SA-4.0 & Questions on FLORES-200 passages, translated \\
    MMLU-ProX (full) & \texttt{8e6106a} & MIT (declared) & Machine translation of MMLU-Pro with expert review \\
    SIB-200 & \texttt{38977a6} & CC-BY-SA-4.0 & English topic labels propagated through FLORES-200 \\
    INCLUDE & \texttt{d2e1f60} & Apache-2.0 (card only) & Native regional exams, not translated \\
    AfriXNLI & \texttt{e3ca06b} & card Apache-2.0 (NC: XNLI-derived) & Translations of XNLI 1.0 test sentences \\
    AmericasNLI & \texttt{1f3f4fa} & card CC-BY-SA-4.0 (NC, SA) & Translations of XNLI 1.0 test sentences \\
    RTP-LX & \texttt{805763e} & MIT/Apache-2.0 + NOTICE (NC; SA for he, ko, pt-BR) & Aligned multilingual toxicity prompts with native components; extremely toxic \\
    BRIGHTER & \texttt{08a5d61} & CC-BY-4.0 & Natively authored texts with emotion-intensity labels \\
    Multilingual HateCheck & \texttt{a03f2ca} & CC-BY-4.0 & Functional tests authored separately per language; hateful by design \\
    Aya red-teaming & \texttt{5a16fee} & Apache-2.0 (card) & Natively written harmful prompts \\
    InjongoIntent & \texttt{fe4be38} & conflicting: CC-BY-4.0 / Apache-2.0 / GPL-3.0 & Natively collected intent utterances \\
    Global PIQA & \texttt{6777742}, \texttt{b0b1851} & CC-BY-SA-4.0, no training & Native split written per variety; parallel split translated \\
    \bottomrule
  \end{tabular}
  \tabnote{NC: noncommercial. SA: ShareAlike.}
  \caption{\textbf{Several sources have noncommercial or ShareAlike terms.} Sources of the evaluation suites and Global PIQA, with pinned revisions (abbreviated), licences as declared upstream and as XDecideBench treats them, and provenance. MultiEURLEX, PAWS-X, XCOPA and MIRACL enter only diagnostic suites. Arabic in \coretwelve{} is Modern Standard Arabic and Chinese is simplified script.}
  \label{tab:sources}
\end{table*}

Before sampling, we dropped two MMLU-ProX questions in all languages, one because its answer letter and answer index disagree and one because a translated distractor is empty.
Manifests record item identifiers, hashes, selection rules and licence flags. They grant no right to redistribute the texts, and noncommercial or ShareAlike terms carry over to any texts rebuilt from them.
Each dataset's original paper or card describes its speaker and annotator populations.

\subsection{Suite Construction}
\label{app:tracks}

Figure~\ref{fig:overview} counts each language once, even when a suite has several regional varieties of it. Counted by variety, MASSIVE has \suiteMassiveVarieties{} locales, RTP-LX \suiteRtplxVarieties{} and native moderation \suiteModerVarieties{}.

\paragraph{Aligned suites.}
The Belebele suite asks the same 100 questions in 106 language varieties. The questions come from 54 passage clusters, none of which appear in \coretwelve{} or the \fresh{}.
MASSIVE uses 117 utterance IDs in all 52 MASSIVE 1.1 locales instead of 120, because one intent has no eligible test row and another has only one. The IDs are disjoint from the \coretwelve{} and \fresh{} MASSIVE IDs.
The multilingual NLI suite gives equal weight to three datasets, each parallel across its own languages: \coretwelve{} XNLI with 12 languages $\times$ 500 items, AfriXNLI with 18 $\times$ 429 and AmericasNLI with 10 $\times$ 150 and no English.
AfriXNLI and AmericasNLI exclude every XNLI premise cluster used in \coretwelve{} or \fresh{} and share only 27 pair IDs with each other. We therefore compute gaps to English only within datasets that contain English.
RTP-LX toxicity asks eight \score{} questions about each of the same 200 source items in 38 aligned locales, one on toxicity with five levels and seven on harm categories with three.

\paragraph{Native suites.}
INCLUDE regional exams samples up to 500 exam questions per language in 44 languages and keeps all questions where fewer exist.
Injongo intent classification has three utterances for each of 40 intents in 16 African languages. Its English rendering of the African utterances is only diagnostic, because it is not natively authored English.
Native moderation combines HateCheck and Aya red-teaming.
HateCheck has 200 \noul{} cases per language in 10 varieties, stratified by functionality, and Aya red-teaming has 120 nine-way harm-category \choice{} cases per language in 8 languages. Each dataset is normalised against its own reference before we take the equal-weight mean.
HateCheck's functional tests are partly shared across languages, but its items are authored separately and have no shared IDs, so we treat it as native.
BRIGHTER emotion intensity asks one four-level intensity question per annotated emotion (six emotions, or five in English) on 250 texts per language.
For \score{} datasets we also report the mean absolute error of the expected level, quadratic-weighted kappa and the ranked probability score.

\paragraph{Diagnostic suites.}
SIB-205 (205 varieties $\times$ 50) is drawn from a training split and its overlap with Belebele was removed only by English exact substring \citep{adelani2024sib200}.
Global PIQA has two options in its native split (136 varieties), so well-resourced languages hit a ceiling. Its parallel split (131 varieties, four options) is never pooled with the native one \citep{chang2025globalpiqa}.
The remaining diagnostics include RTP-LX's two unaligned locales, hbs and sw (\suiteDiagRtplxTexts{} texts, \suiteDiagRtplxQuestions{} questions, no aligned English) and InjongoIntent English.
The others are the described-label variant of \coretwelve{} MASSIVE and SIB-200 with 16,896 cases, English dev with 1,000, MultiEURLEX, PAWS-X and XCOPA with 2,600, and MIRACL relevance in three languages with 614.

\subsection{RTP-LX Gold Labels Across Locales}
\label{app:rtplx}

\begin{table}[h]
  \centering
  \small
  \setlength{\tabcolsep}{5pt}
  \begin{tabular}{@{}l S[table-format=0.3] S[table-format=+1.3, print-zero-integer=true] S[table-format=0.3]@{}}
    \toprule
    Rubric (levels) & {Agree} & {Locale$\minus$en} & {Differ $\ge$2} \\
    \midrule
    toxicity (5)        & .457 & \minus0.269 & .167 \\
    insult (3)          & .655 & \plus0.076  & .149 \\
    bias (3)            & .729 & \plus0.074  & .130 \\
    identity attack (3) & .740 & \plus0.087  & .120 \\
    violence (3)        & .864 & \plus0.090  & .058 \\
    microaggression (3) & .896 & \plus0.152  & .048 \\
    sexual content (3)  & .897 & \plus0.068  & .039 \\
    self-harm (3)       & .976 & \plus0.031  & .013 \\
    \midrule
    all rubrics         & .777 & \plus0.039  & .090 \\
    \bottomrule
  \end{tabular}
  \tabnote{\emph{Agree}: exact agreement of the median gold levels. \emph{Locale$\minus$en}: mean signed level difference. \emph{Differ $\ge$2}: share of pairs two or more levels apart.}
  \caption{\textbf{Locale labels often differ from the English gold.} They differ most on toxicity. Each row pools the 7,400 aligned pairs of a non-English locale and the English item with the same source index (37 locales $\times$ 200).}
  \label{tab:rtplx}
\end{table}

RTP-LX gives each locale its own three annotators, and the gold is their median, so an aligned item can have different targets in different locales.
Table~\ref{tab:rtplx} compares each locale's gold with the English gold for the same source item.
Toxicity agreement per locale ranges from 0.23 (Slovak) to 0.58 (Brazilian Portuguese), and locale annotators rate toxicity lower than the English annotators on average.
We score against the median of the three released annotations, which differs from the released aggregate label in 92 rows.

\subsection{INCLUDE Sample}
\label{app:include}

\begin{table}[h]
  \centering
  \small
  \setlength{\tabcolsep}{6pt}
  \begin{tabular}{@{}l S[table-format=3.0] S[table-format=3.0] S[table-format=4.0]@{}}
    \toprule
    Lang & {Test rows} & {Eligible} & {Frozen} \\
    \midrule
    ar & 552 & 544 & 500 \\
    de & 139 & 138 & 138 \\
    es & 550 & 549 & 500 \\
    fr & 419 & 419 & 419 \\
    hi & 547 & 545 & 500 \\
    ru & 552 & 542 & 500 \\
    ur & 352 & 343 & 343 \\
    vi & 550 & 549 & 500 \\
    zh & 545 & 545 & 500 \\
    \midrule
    Total & {} & {} & 3900 \\
    \bottomrule
  \end{tabular}
  \caption{\textbf{The INCLUDE sample.} It caps the items frozen per language and keeps every eligible item where fewer exist. This nine-language subset of the INCLUDE suite serves the native-versus-translated comparison (Appendix~\ref{app:nat}).}
  \label{tab:include}
\end{table}

\section{Model Interfaces}
\label{app:interfaces}

\subsection{Backbone Prompts}
\label{app:prompt}

The local readout uses the frozen user template below, wrapped in each backbone's chat template with the generation prompt on and thinking disabled. \texttt{\{options\}} lists one option text per line, prefixed by a hyphen, in canonical order.

\begin{small}
\begin{verbatim}
Read the state and answer the question. Reply
with exactly one of the allowed options, copied
verbatim.

STATE:
{state}

QUESTION:
{instruction}

ALLOWED OPTIONS:
{options}
\end{verbatim}
\end{small}

\noindent The line breaks inside the first sentence are only typographic.
The hosted readout keeps the state, instruction and option sections verbatim but lists options as \texttt{<label>. <text>} and asks for the label.
Labels run A--Z, a--z and 0--9, which allows up to 62 options, and each label is a single token in both Qwen tokenizers.
The readout decodes one token at temperature 0 with reasoning disabled.
A label's probability is the sum of the returned top-20 probabilities of tokens that equal the label after leading whitespace is stripped.
Mass that is not returned is shared equally among labels outside the top 20, and the result is renormalised.
Both endpoints are pinned to one provider and quantisation without fallbacks.

\subsection{Precision Rules and Interface Limits}
\label{app:precision}

\begin{table}[h]
  \centering
  \footnotesize
  \setlength{\tabcolsep}{4pt}
  \begin{tabular}{@{}l S[table-format=1.3, print-zero-integer=true, table-number-alignment=left, table-text-alignment=left] >{\raggedright\arraybackslash}p{0.48\columnwidth}@{}}
    \toprule
    Rule & {$|\sum p - 1|\le$} & Models \\
    \midrule
    two decimals  & 0.01  & Jev, Luna \\
    four decimals & 0.005 & Kev-4B; Clef, Clef-flash (API) \\
    six decimals  & {$10^{-4}$} & Solar \\
    strict        & {$10^{-5}$} & pplx-decider, d1, Mercury, Decision-1 \\
    fp32          & {strict} & Clef, Clef-flash (local) \\
    \bottomrule
  \end{tabular}
  \caption{\textbf{Acceptance tolerances follow each interface's precision.} Accepted vectors are renormalised, and the native choice may trail the rounded maximum by one rounding unit.}
  \label{tab:precision}
\end{table}

Two-decimal APIs return exact zeros, which leave log loss undefined without a floor. Their ties flatten risk--coverage curves.
Solar Decide rejects questions with more than 26 options, which rules out the 60-intent \coretwelve{} MASSIVE cases and all of MASSIVE-52 and Injongo. The scorer flags those \solarCappedMassive{} and \solarCappedInjongo{} cells as capped and scores them as uniform.
Laya's rendered-option cap refuses 372 of 26,460 \coretwelve{} cases, all of them MMLU-ProX.
Laya also refuses 6,000 described-label MASSIVE cases, 189 Belebele cases in 48 of the 106 languages and 27 cases of the nine-language INCLUDE sample, 25 by the option cap and 2 by input length.
GLiNER2.5-Decide reads at most 512 tokens. Its adapter refuses any case whose full question and text exceed that window instead of scoring a silently truncated input.
It therefore answers 24,387 of 26,460 \coretwelve{} cases, 8,398 of 10,600 Belebele cases, 20,640 of 21,079 INCLUDE cases, 1,719 of 2,040 Injongo cases and 10,896 of 16,896 described-label cases. In \glinerCappedBelebele{} Belebele languages (Tibetan, Burmese and Shan) it refuses nearly every case, and the scorer flags those cells as capped.
It also refuses 156 of 26,569 Global PIQA cases and answers every other configuration in full.
The hosted Clef routes silently truncate inputs at 16,384 tokens, well above our longest case of about 1,960 tokens.
Local Clef and Clef-flash use the vendor's length-bucketed batching. On synthetic test inputs, its probabilities differ from those of single requests by up to 0.006, and no prediction changes.
Every other model has full valid-output coverage on every suite it ran, apart from the cells marked in Table~\ref{tab:matrix}.
Table~\ref{tab:failzero} counts the failed items and the share of valid answers with an exact-zero probability for every model and suite. The exact zeros come from the two-decimal APIs (Jev, GPT-6 Luna).

\section{Model Rankings and Label Regimes}
\label{app:full}

\subsection{\coretwelve{} Scores and Rankings}

\begin{table*}[t]
  \centering
  \small
  \setlength{\tabcolsep}{6pt}
  \begin{tabular}{@{}l S[table-format=0.3]@{\hspace{4pt}}l c l@{\hspace{3pt}}S[table-format=0.2] S[table-format=-0.3] S[table-format=1.3]@{}}
    \toprule
    Model & \multicolumn{2}{c}{Brier skill} & Rank & \multicolumn{2}{c}{Worst} & {Gap to \lang{en}} & {Cov.} \\
    \midrule
    \tgroup{8}{Hosted Typed Decision Models, by \coretwelve{} skill} \\
    pplx-decider & \skPplxCoreTwelve & \tci{\ctPplxCI} & \rkPplxCoreTwelve & \lang{\ctPplxWorstLang} & \ctPplxWorst & \ctPplxGap & \ctPplxCov \\
    Jev & \skJevCoreTwelve & \tci{\ctJevCI} & \rkJevCoreTwelve & \lang{\ctJevWorstLang} & \ctJevWorst & \ctJevGap & \ctJevCov \\
    Clef, API & \skClefCoreTwelve & \tci{\ctClefCI} & \rkClefCoreTwelve & \lang{\ctClefWorstLang} & \ctClefWorst & \ctClefGap & \ctClefCov \\
    d1 & \skLiquidCoreTwelve & \tci{\ctLiquidCI} & \rkLiquidCoreTwelve & \lang{\ctLiquidWorstLang} & \ctLiquidWorst & \ctLiquidGap & \ctLiquidCov \\
    Decision-1 & \skMsftCoreTwelve & \tci{\ctMsftCI} & \rkMsftCoreTwelve & \lang{\ctMsftWorstLang} & \ctMsftWorst & \ctMsftGap & \ctMsftCov \\
    Clef-flash, API & \skClefFlashCoreTwelve & \tci{\ctClefFlashCI} & \rkClefFlashCoreTwelve & \lang{\ctClefFlashWorstLang} & \ctClefFlashWorst & \ctClefFlashGap & \ctClefFlashCov \\
    Mercury & \skMercuryCoreTwelve & \tci{\ctMercuryCI} & \rkMercuryCoreTwelve & \lang{\ctMercuryWorstLang} & \ctMercuryWorst & \ctMercuryGap & \ctMercuryCov \\
    Luna & \skLunaCoreTwelve & \tci{\ctLunaCI} & \rkLunaCoreTwelve & \lang{\ctLunaWorstLang} & \ctLunaWorst & \ctLunaGap & \ctLunaCov \\
    Kev-4B & \skKevCoreTwelve & \tci{\ctKevCI} & \rkKevCoreTwelve & \lang{\ctKevWorstLang} & \ctKevWorst & \ctKevGap & \ctKevCov \\
    Solar & \skSolarCoreTwelve & \tci{\ctSolarCI} & \rkSolarCoreTwelve & \lang{\ctSolarWorstLang} & \ctSolarWorst & \ctSolarGap & \ctSolarCov \\
    \addlinespace
    \tgroup{8}{Local heads} \\
    Clef, local & \skClefLocalCoreTwelve & \tci{\ctClefLocalCI} & \tmark{--} & \lang{\ctClefLocalWorstLang} & \ctClefLocalWorst & \ctClefLocalGap & \ctClefLocalCov \\
    Clef-flash, local & \skClefFlashLocalCoreTwelve & \tci{\ctClefFlashLocalCI} & \tmark{--} & \lang{\ctClefFlashLocalWorstLang} & \ctClefFlashLocalWorst & \ctClefFlashLocalGap & \ctClefFlashLocalCov \\
    \addlinespace
    \tgroup{8}{Backbone readouts} \\
    Qwen3.8-27B$^\ast$ & \skQwenTwentySevenLocalCoreTwelve & \tci{\ctQwenTwentySevenLocalCI} & \tmark{--} & \lang{\ctQwenTwentySevenLocalWorstLang} & \ctQwenTwentySevenLocalWorst & \ctQwenTwentySevenLocalGap & \ctQwenTwentySevenLocalCov \\
    Qwen3.5-9B$^\ast$ & \skQwenNineLocalCoreTwelve & \tci{\ctQwenNineLocalCI} & \rkQwenNineLocalCoreTwelve & \lang{\ctQwenNineLocalWorstLang} & \ctQwenNineLocalWorst & \ctQwenNineLocalGap & \ctQwenNineLocalCov \\
    Qwen3.8-27B$^\circ$ & \skQwenTwentySevenHostedCoreTwelve & \tci{\ctQwenTwentySevenHostedCI} & \tmark{--} & \lang{\ctQwenTwentySevenHostedWorstLang} & \ctQwenTwentySevenHostedWorst & \ctQwenTwentySevenHostedGap & \ctQwenTwentySevenHostedCov \\
    Qwen3.5-9B$^\circ$ & \skQwenNineHostedCoreTwelve & \tci{\ctQwenNineHostedCI} & \tmark{--} & \lang{\ctQwenNineHostedWorstLang} & \ctQwenNineHostedWorst & \ctQwenNineHostedGap & \ctQwenNineHostedCov \\
    \bottomrule
  \end{tabular}
  \tabnote{Brier skill: estimate followed by its 95\% joint cluster-bootstrap interval \tci{[lo, hi]}. \emph{Rank}: hierarchical 95\% rank interval among the \nRankedCoreTwelve{} models ranked together, the ten hosted Typed Decision Models and the local Qwen3.5-9B backbone, with datasets and languages also resampled and bounds widened to whole ranks. \tmark{--}: not ranked. \emph{Worst}: lowest-scoring language and its skill. \emph{Gap}: mean language minus English. \emph{Cov.}: share of valid typed answers. $^\ast$Backbone, local readout. $^\circ$Backbone, hosted readout.}
  \caption{\textbf{pplx-decider leads \coretwelve{}.} Every model is weakest in Swahili or Hindi. Per-model \coretwelve{} Brier skill against uniform chance, rank interval, worst language, gap to English and coverage. Solar's \ctSolarCappedCells{} MASSIVE cells score as uniform (26-option cap), and without them it scores \skSolarCoreTwelveUncapped{}. Laya runs a separate \coretwelve{} manifest, and mmBERT has no Belebele or MMLU-ProX head, so neither has a suite score (\S\ref{sec:panel}).}
  \label{tab:core12}
\end{table*}

Leaving out any one dataset changes the \coretwelve{} ranking little (Kendall $\tau$ \lodoTauLo--\lodoTauHi). We report a view without Solar's capped MASSIVE cells but do not rank it.
Intervals on the other suites are wide. Clef-flash has the top MASSIVE-52 score (\skClefFlashMassive{}), for example, but its interval overlaps those of the next six models.

\paragraph{English selection under other criteria.}
The main text counts a cell as a loss for the English pick when the paired loss interval excludes zero (\regEnWrongPaired{} of \regEnUnits{} cells).
The two models' own intervals do not overlap in \regEnWrongSep{} cells, and the loss is at least \regEnThreshold{} Brier skill in \regEnGeTen{}.
Within English, the pick's interval never lies above the runner-up's, and a paired interval separates them in only one of the six English comparisons (pplx-decider against Mercury), so the English results do not settle the choice.
Of the \coretwelve{} pick's paired-significant losses on Belebele, INCLUDE and BRIGHTER, only the INCLUDE loss also has non-overlapping intervals.

\paragraph{Ordinal suites under RPS skill.}
The \ordModels{} hosted Typed Decision Models all completed RTP-LX and BRIGHTER. We rank them by ranked probability skill against uniform, macro-averaged over languages, and compare this ranking with the Brier-skill ranking of Table~\ref{tab:matrix}.
On BRIGHTER the rankings agree at Kendall $\tau_b=\tauRpsVsBrierBrighter$. They differ only in three adjacent pairs, where Clef and pplx-decider swap first and second place, Clef-flash and d1 third and fourth, and Kev-4B and Decision-1 seventh and eighth.
On RTP-LX $\tau_b=\tauRpsVsBrierRtplx$. Luna moves from fifth to second, d1 from second to fifth and Clef-flash from first to third, and \ordRtplxSwappedPairs{} of \ordRtplxPairs{} model pairs swap.
Because the references differ (base rate for Brier skill, uniform for RPS skill), we compare only the orderings.

\subsection{Native Versus Translated Exams and Described Labels}
\label{app:nat}

MMLU-ProX translates English exam questions, whereas INCLUDE collects exams written in each language \citep{xuan2025mmluprox,romanou2025include}.
On the nine languages both cover (Appendix~\ref{app:include}), the two sets differ in subject mix, option count (four against ten) and difficulty, and INCLUDE items are not parallel. We therefore compare chance-referenced skills within each arm and language and never compare raw accuracy across the two sets.
Table~\ref{tab:natdesc} gives each arm's INCLUDE minus MMLU-ProX Brier skill, averaged over the nine languages (3,900 INCLUDE items and 500 MMLU-ProX items per language), with a 95\% bootstrap interval. The bootstrap resamples MMLU-ProX parallel groups and INCLUDE items within each language, and all arms share one draw.
Only the differences between arms are therefore meaningful.
Solar has the lowest gap at $-.155$ and Luna the highest at $+.210$, so Luna's lead on INCLUDE in Table~\ref{tab:matrix} is specific to native exams. Jev does worse on native exams than on translated ones, at $-.082$.
German and Urdu have a negative gap for most arms.

The described-label diagnostic repeats the 8,448 \coretwelve{} MASSIVE and SIB-200 cases with a one-line English description per label, keeping state, gold and option identities fixed.
We pair each described case with its original from the same run and report the change in Brier skill, averaged over the twelve languages, with a joint 10,000-replicate bootstrap.
On MASSIVE, descriptions raise skill for Kev-4B, Jev, Mercury and Luna by .035 to .053 and lower it slightly for both Clef heads. On SIB-200 every change lies between $-.048$ and 0, and only the intervals of Decision-1 ($-.048$) and Laya ($-.035$) exclude zero.
Laya rejects all 6,000 described MASSIVE cases because its option-description cap would truncate them, but answers all 6,000 originals. Solar's option cap rules out MASSIVE in both variants.
The local backbones lose .43 to .61, but their readout scores the whole description sentence in place of the label text, so this loss reflects the changed readout input rather than the label names.

\begin{table*}[t]
  \centering
  \footnotesize
  \setlength{\tabcolsep}{3pt}
  \begin{tabular}{@{}l
      S[table-format=-0.3]>{\raggedright\arraybackslash}p{2.6cm}
      S[table-format=-0.3]>{\raggedright\arraybackslash}p{2.6cm}
      S[table-format=-0.3]>{\raggedright\arraybackslash}p{2.6cm}@{}}
    \toprule
    & \multicolumn{2}{c}{NAT: INCLUDE $-$} & \multicolumn{4}{c@{}}{DESC: described $-$ bare} \\
    \cmidrule(l){4-7}
    Arm & \multicolumn{2}{c}{MMLU-ProX} & \multicolumn{2}{c}{MASSIVE} & \multicolumn{2}{c@{}}{SIB-200} \\
    \midrule
    \tgroup{7}{Typed Decision Models} \\
    pplx-decider & .052 & \tci{[\minus.005, .109]} & .009 & \tci{[\minus.001, .019]} & \minus.011 & \tci{[\minus.049, .023]} \\
    Jev & \minus.082 & \tci{[\minus.136, \minus.028]} & .043 & \tci{[.020, .066]} & .000 & \tci{[\minus.028, .025]} \\
    Clef, API & .095 & \tci{[.045, .144]} & \minus.021 & \tci{[\minus.037, \minus.006]} & \minus.008 & \tci{[\minus.051, .032]} \\
    Clef, local & .094 & \tci{[.045, .143]} & \minus.021 & \tci{[\minus.037, \minus.005]} & \minus.008 & \tci{[\minus.051, .031]} \\
    d1 & .102 & \tci{[.046, .157]} & .007 & \tci{[\minus.010, .025]} & \minus.022 & \tci{[\minus.058, .011]} \\
    Clef-flash, API & .072 & \tci{[.024, .120]} & \minus.013 & \tci{[\minus.030, .003]} & \minus.006 & \tci{[\minus.039, .026]} \\
    Clef-flash, local & .072 & \tci{[.024, .120]} & \minus.013 & \tci{[\minus.030, .003]} & \minus.006 & \tci{[\minus.039, .026]} \\
    Mercury & .147 & \tci{[.081, .211]} & .039 & \tci{[.018, .061]} & \minus.021 & \tci{[\minus.064, .015]} \\
    Luna & .210 & \tci{[.158, .259]} & .035 & \tci{[.014, .056]} & \minus.012 & \tci{[\minus.064, .036]} \\
    Kev-4B & .096 & \tci{[.059, .132]} & .053 & \tci{[.039, .067]} & \minus.012 & \tci{[\minus.034, .010]} \\
    Solar & \minus.155 & \tci{[\minus.218, \minus.087]} & \multicolumn{2}{c}{\tmark{--}} & \minus.035 & \tci{[\minus.089, .016]} \\
    Decision-1 & .109 & \tci{[.048, .171]} & \minus.013 & \tci{[\minus.028, .002]} & \minus.048 & \tci{[\minus.084, \minus.015]} \\
    Laya & .002 & \tci{[\minus.018, .023]} & \multicolumn{2}{c}{\tmark{refused}} & \minus.035 & \tci{[\minus.069, \minus.0005]} \\
    \addlinespace
    \tgroup{7}{Backbone readouts} \\
    Qwen3.8-27B, hosted & .172 & \tci{[.114, .229]} & \multicolumn{2}{c}{\tmark{n.r.}} & \multicolumn{2}{c@{}}{\tmark{n.r.}} \\
    Qwen3.5-9B, hosted & .182 & \tci{[.128, .236]} & \multicolumn{2}{c}{\tmark{n.r.}} & \multicolumn{2}{c@{}}{\tmark{n.r.}} \\
    Qwen3.8-27B, local & .152 & \tci{[.128, .176]} & \multicolumn{2}{c}{\tmark{n.c.}} & \multicolumn{2}{c@{}}{\tmark{n.c.}} \\
    Qwen3.5-9B, local & .129 & \tci{[.107, .150]} & \multicolumn{2}{c}{\tmark{n.c.}} & \multicolumn{2}{c@{}}{\tmark{n.c.}} \\
    \bottomrule
  \end{tabular}
  \tabnote{Estimate followed by its 95\% bootstrap interval \tci{[lo, hi]}. NAT: native INCLUDE minus translated MMLU-ProX, mean over nine languages. DESC: described labels minus bare label identifiers on the same \coretwelve{} cases, mean over twelve languages. \tmark{--}: Solar's 26-option cap rejects MASSIVE. \tmark{refused}: Laya rejects every described MASSIVE case, scored as a failure. \tmark{n.r.}: not run. \tmark{n.c.}: not comparable, because the local readout scores the description text itself rather than the label.}
  \caption{\textbf{The native-exam gap changes sign across arms.} Described labels move the skill of Typed Decision Models much less. Change in Brier skill against chance for native against translated exams (NAT) and described against bare labels (DESC).}
  \label{tab:natdesc}
\end{table*}

\subsection{Interface Language (Exploratory)}
\label{app:native}

\begin{table}[h]
  \centering
  \footnotesize
  \setlength{\tabcolsep}{3pt}
  \sisetup{table-format=-0.3, table-space-text-post=$^{*}$}
  \begin{tabular}{@{}l SS SS@{}}
    \toprule
    & \multicolumn{2}{c}{Native instr.} & \multicolumn{2}{c@{}}{Fully native} \\
    \cmidrule(lr){2-3}\cmidrule(l){4-5}
    Model & {XNLI} & {MASSIVE} & {XNLI} & {MASSIVE} \\
    \midrule
    \tgroup{5}{Hosted Typed Decision Models} \\
    Jev & \niJevXnliNat & \niJevMassiveNat & \niJevXnliFull & \niJevMassiveFull \\
    pplx-decider & \niPplxXnliNat & \niPplxMassiveNat & \niPplxXnliFull & \niPplxMassiveFull \\
    d1 & \niLiquidXnliNat & \niLiquidMassiveNat & \niLiquidXnliFull & \niLiquidMassiveFull \\
    Luna & \niLunaXnliNat & \niLunaMassiveNat & \niLunaXnliFull & \niLunaMassiveFull \\
    Mercury & \niMercuryXnliNat & \niMercuryMassiveNat & \niMercuryXnliFull & \niMercuryMassiveFull \\
    Kev-4B & \niKevXnliNat & \niKevMassiveNat & {\tmark{n.r.}} & {\tmark{n.r.}} \\
    Solar & \niSolarXnliNat & {\tmark{--}} & {\tmark{n.r.}} & {\tmark{n.r.}} \\
    Clef & \niClefXnliNat & \niClefMassiveNat & {\tmark{n.r.}} & {\tmark{n.r.}} \\
    Clef-flash & \niClefFlashXnliNat & \niClefFlashMassiveNat & \niClefFlashXnliFull & \niClefFlashMassiveFull \\
    Decision-1 & \niMsftXnliNat & \niMsftMassiveNat & {\tmark{n.r.}} & {\tmark{n.r.}} \\
    \addlinespace
    \tgroup{5}{Backbone readout} \\
    Qwen3.5-9B$^\circ$ & \niQwenNineXnliNat & \niQwenNineMassiveNat & {\tmark{n.c.}} & {\tmark{n.c.}} \\
    \bottomrule
  \end{tabular}
  \tabnote{\emph{Native instr.}: translated instruction and option texts, English option IDs. \emph{Fully native}: translated texts with neutral IDs. Clef and Clef-flash through the API. $^\circ$hosted backbone readout. $^{*}$: significant (95\% paired cluster-bootstrap interval excludes zero). \tmark{n.r.}: not run. \tmark{--}: Solar's 26-option cap rejects MASSIVE. \tmark{n.c.}: the hosted readout never shows option IDs.}
  \caption{\textbf{Native instructions cost the Qwen-based heads most on XNLI.} Neutral option IDs cost most on MASSIVE. Change in Brier skill against chance relative to the English-schema \coretwelve{} run on the same items, mean over \niLangs{} languages.}
  \label{tab:native}
\end{table}

We registered this ablation before its first request, separately from the \fresh{} pre-registration.
On \coretwelve{} XNLI and MASSIVE, all 500 items per language in Arabic, German, French, Hindi, Swahili, Thai, Urdu and Chinese get a copy with translated instruction and option texts. Option IDs, states and gold answers are unchanged, and the paired baseline is each model's existing \coretwelve{} run.
A second condition replaces the English option IDs with neutral ones, in English and in translated form, for \niFullModels{} models. We left out Kev-4B, Solar and Decision-1 to stay within budget.
\texttt{claude-opus-5.5} translated the 520 texts from a written brief, and \texttt{gpt-6.1-sol}, blind to the translator, back-translated them and mapped every translated option back to an English ID. All 504 options were recovered, and no review changed a label's meaning.
No native speaker reviewed the translations, and the XNLI state keeps its English premise and hypothesis headers.

Pooled over the Typed Decision Models, the low-minus-high difference on XNLI is \niXnliLowHigh{} \niXnliLowHighCI{} points of Brier skill in favour of the high-web-share group.
Removing the English IDs alone lowers MASSIVE skill by \niMassiveIds{} \niMassiveIdsCI{}, and a synthetic probe confirms that nine of the ten hosted Typed Decision Models follow the IDs when option texts are blank.
Only GPT-6 Luna and, slightly, Jev gain on XNLI under native instructions.
Clef-flash's Brier-skill advantage over the hosted Qwen3.5-9B readout shrinks by \niHeadShrink{} \niHeadShrinkCI{} on XNLI and by \niHeadShrinkMassive{} on MASSIVE.
The English pick on MASSIVE is wrong in \niPickWrong{} of \niLangs{} languages under either schema. Under native instructions its mean loss rises from \niRegEn{} to \niRegNative{}.
The model ranking by language mean agrees with the English-schema ranking at Kendall $\tau=\niTauXnli$ on XNLI and \niTauMassive{} on MASSIVE.
On MASSIVE, the best model changes in \niBestChanges{} languages, and the English pick's largest loss, in Swahili, grows from \niRegMaxEn{} to \niRegMaxNative{}.

\section{Heads Versus Backbones}
\label{app:hb}

\subsection{Gains per Language and Dataset}

\begin{figure}[h]
  \centering
  \includegraphics[width=\columnwidth]{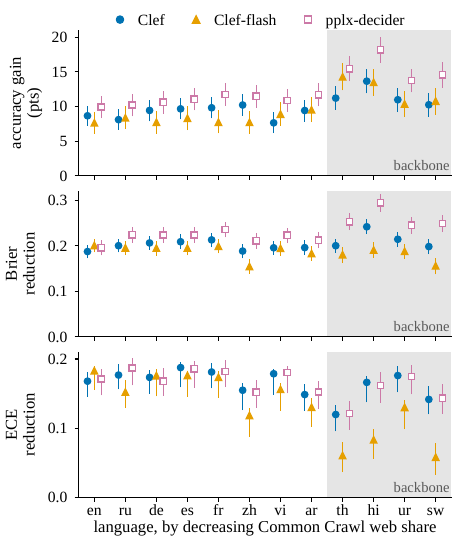}
  \caption{\textbf{Each head beats its backbone in all \coretwelve{} languages and metrics.} Head minus backbone for accuracy and backbone minus head for Brier score and ECE under the local readout, so higher is better and 0 is the backbone. Bars show 95\% paired source-cluster bootstrap intervals, shading marks the low-web-share group, and open squares mark pplx-decider, the vendor-stated pair.}
  \label{fig:a1-metrics}
\end{figure}

\paragraph{Web share and the local-readout accuracy gain.}
Across the twelve \coretwelve{} languages, the head's accuracy gain over its backbone falls as web share rises. Spearman's $\rho$ is $-0.69$ for Clef, $-0.82$ for Clef-flash and $-0.87$ for pplx-decider (permutation $p=.017$, .002 and $<.001$), and the OLS slopes are $-0.8$, $-1.4$ and $-1.7$ accuracy points per tenfold increase in web share.
These correlations are not independent, since Clef and pplx-decider share a backbone and all three pairs share items.

\paragraph{Head--backbone interactions by dataset.}
Under the local readout, Belebele and MASSIVE have the largest accuracy interactions.
On Belebele they are $+7.2$ points for Clef, $+7.1$ for Clef-flash and $+8.6$ for pplx-decider, and on MASSIVE they are $+4.3$, $+9.5$ and $+6.6$ in the same order.
On XNLI the sign depends on the pair, with Clef's gain smaller in the low-web-share group ($-2.9$ [$-5.5$, $-0.3$]) and Clef-flash's larger ($+6.2$ [3.8, 8.6]).
For both Clef heads the Brier contrasts on XNLI and MMLU-ProX favour the high-web-share group, and for Clef-flash they outweigh Belebele and MASSIVE.

\subsection{English Temperature Scaling}
\label{app:temp}

\begin{table*}[t]
  \centering
  \footnotesize
  \setlength{\tabcolsep}{5pt}
  \sisetup{print-zero-integer=true, table-format=1.2}
  \begin{tabular}{@{}l SS SS
      S[table-format=-0.3, print-zero-integer=false]@{\ to\ }S[table-format=+0.3, print-zero-integer=false]
      S[table-format=-0.3, print-zero-integer=false]@{\ to\ }S[table-format=-0.3, print-zero-integer=false]@{}}
    \toprule
    & \multicolumn{2}{c}{$T_h$} & \multicolumn{2}{c}{$T_b$} & \multicolumn{2}{c}{$\Delta$ECE raw} & \multicolumn{2}{c@{}}{$\Delta$ECE scaled} \\
    \cmidrule(lr){2-3}\cmidrule(lr){4-5}\cmidrule(lr){6-7}\cmidrule(l){8-9}
    Pair & {MASSIVE} & {XNLI} & {MASSIVE} & {XNLI} & {lowest} & {highest} & {lowest} & {highest} \\
    \midrule
    Clef         & 0.97 & 0.86 & 0.65 & 1.24 & \minus.218 & \minus.118 & \minus.121 & \minus.019 \\
    Clef-flash   & 0.97 & 0.78 & 0.68 & 0.22 & \minus.236 & \plus.005  & \minus.077 & \minus.003 \\
    pplx-decider & 1.20 & 0.92 & 0.65 & 1.24 & \minus.214 & \minus.121 & \minus.130 & \minus.050 \\
    \bottomrule
  \end{tabular}
  \tabnote{$T_h$, $T_b$: temperatures of the head and of its backbone's local readout. $\Delta$ECE: head minus backbone, lowest to highest over the languages.}
  \caption{\textbf{English temperatures shrink the heads' ECE advantage.} Clef-flash loses the most. One temperature per arm and dataset, MASSIVE/XNLI, fitted by negative log-likelihood on the English development half used for temperatures, for each head ($T_h$) and its backbone's local readout ($T_b$). ECE differences are head minus backbone per language, as the mean over MASSIVE and XNLI, ranged over the twelve \coretwelve{} languages. On these two datasets raw Clef-flash ECE is higher than its backbone's in Swahili (\plus.005).}
  \label{tab:temp}
\end{table*}

\begin{figure}[h]
  \centering
  \includegraphics[width=\columnwidth]{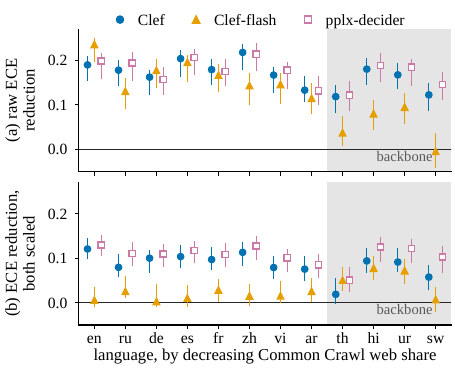}
  \caption{\textbf{Heads keep part of their ECE advantage after English temperature scaling.} Clef-flash keeps the least. ECE reduction, backbone minus head, on \coretwelve{} MASSIVE and XNLI under the local readout, (a) raw and (b) after applying one temperature per dataset, fitted on English development data, to both the head and its backbone. Bars show 95\% paired source-cluster intervals, and shading marks the low-web-share group.}
  \label{fig:temp}
\end{figure}

Scaling lowers backbone ECE on MASSIVE (Table~\ref{tab:temp}) from 0.132--0.199 to 0.029--0.088 for Qwen3.8-27B and from 0.124--0.254 to 0.027--0.087 for Qwen3.5-9B.
On XNLI it helps much less (0.134--0.347 to 0.116--0.299, and 0.059--0.329 to 0.035--0.283).
After scaling, the head's Brier advantage ranges from $-0.278$ to $-0.213$ for Clef, $-0.296$ to $-0.142$ for Clef-flash and $-0.331$ to $-0.253$ for pplx-decider, and the paired interval excludes zero in all 36 language cells.
For ECE, 29 of the 36 intervals lie below zero. The exceptions are Thai for Clef and German, English, Spanish, Swahili, Vietnamese and Chinese for Clef-flash.
After scaling, the low-minus-high web-share contrast in the ECE difference is $+0.037$ [0.017, 0.050] for Clef, $+0.017$ [0.003, 0.035] for pplx-decider and $-0.037$ [$-0.049$, $-0.014$] for Clef-flash.
Intervals use 10,000 paired source-cluster replicates.

\subsection{Wide-Belebele Breadth Check (Exploratory)}
\label{app:bb}

\begin{table*}[t]
  \centering
  \footnotesize
  \setlength{\tabcolsep}{6pt}
  \begin{tabular}{@{}l S[table-format=3.0] S[table-format=2.0] S[table-format=1.1, print-zero-integer=true] S[table-format=0.2]
      S[table-format=-0.2]@{\ }l S[table-format=-1.1, print-zero-integer=true]@{}}
    \toprule
    Head & {Gain${>}0$} & {CI${>}0$} & {Median gain} & {Median share} & \multicolumn{2}{c}{$\rho$ with web share ($p$)} & {Slope} \\
    \midrule
    Clef         & \bbPosClef      & \bbCiPosClef      & \bbAccMedianClef      & \bbEsMedianClef      & \bbRhoClef      & (\bbPClef)      & \bbSlopeClef \\
    Clef-flash   & \bbPosClefFlash & \bbCiPosClefFlash & \bbAccMedianClefFlash & \bbEsMedianClefFlash & \bbRhoClefFlash & (\bbPClefFlash) & \bbSlopeClefFlash \\
    pplx-decider & \bbPosPplx      & \bbCiPosPplx      & \bbAccMedianPplx      & \bbEsMedianPplx      & \bbRhoPplx      & (\bbPPplx)      & \bbSlopePplx \\
    \bottomrule
  \end{tabular}
  \tabnote{\emph{Gain${>}0$}: languages where the head's accuracy is higher. \emph{CI${>}0$}: languages whose 95\% paired source-cluster interval lies above zero. \emph{Median gain}: in accuracy points. \emph{Median share}: share of backbone errors removed. $\rho$: Spearman correlation of the gain with log web share over \bbWebN{} languages, with permutation $p$. \emph{Slope}: OLS, accuracy points per tenfold web share.}
  \caption{\textbf{Every head beats its backbone in most Wide-Belebele languages.} Only Clef's and pplx-decider's gains fall with web share. Wide-Belebele breadth check (exploratory, local readout): each head against its own backbone in \bbLanguages{} languages.}
  \label{tab:bb106}
\end{table*}

We repeat the head--backbone comparison on all \bbLanguages{} Wide-Belebele languages under the local readout, with the same paired bootstrap (Table~\ref{tab:bb106}).
The low- and high-web-share groups exist only for the twelve \coretwelve{} languages, so here web share is the only language variable.
The accuracy gain is above zero in \bbPosClef{} languages for Clef, \bbPosClefFlash{} for Clef-flash and \bbPosPplx{} for pplx-decider, and no language has an interval below zero.
All three heads beat their backbones in \bbAllBeat{} languages, and in \bbAllBeatCI{} of them all three intervals lie above zero.
The Brier score improves significantly (interval below zero) in \bbBrierCiNegClef{}, \bbBrierCiNegClefFlash{} and \bbBrierCiNegPplx{} languages and worsens significantly (interval above zero) in \bbBrierCiPosClef{}, \bbBrierCiPosClefFlash{} and \bbBrierCiPosPplx{}.
Over the \bbWebN{} languages with a usable web share, the accuracy gain falls as web share rises for Clef ($\rho=\bbRhoClef$, \bbPClef) and pplx-decider ($\rho=\bbRhoPplx$, \bbPPplx), with OLS slopes of \bbSlopeClef{} and \bbSlopePplx{} points per tenfold share.
Clef-flash shows no such trend ($\rho=\bbRhoClefFlash$, \bbPClefFlash), unlike on \coretwelve{}. Instead, its Brier difference tracks web share ($\rho=\bbBrierRhoClefFlash$, \bbBrierPClefFlash), with larger Brier gains where web share is higher.
For Clef and pplx-decider the Brier difference shows no clear trend ($\rho=\bbBrierRhoClef$ and $\bbBrierRhoPplx$, \bbBrierPClef{} and \bbBrierPPplx).
With the merged-variety languages added back ($n=\bbWebNSens$), the accuracy correlations are \bbRhoSensClef{}, \bbRhoSensClefFlash{} and \bbRhoSensPplx{}.
Each language has only a hundred questions, so single-language intervals are wide.

Web shares come from the same crawl as above, recomputed from page counts.
Each FLORES code maps to its own crawl (CLD2) code or, if the crawl lists only the macrolanguage, to its ISO 639-3 macrolanguage (\texttt{swh\_Latn} to Swahili, \texttt{nob\_Latn} to Norwegian).
Bambara, Fula, Ilocano, Jingpho and Shan (\bbUnmapped{} languages) have neither and get no web share.
We also leave out \bbMerged{} languages whose crawl code also covers a FLORES+ variety without its own code, either a different language or the same language in another script. These are Arabic, Azerbaijani, Chinese, Kurdish, Latvian, Malay, Mongolian, Persian, Sindhi and Uzbek.
The full mapping will be released with the code\afterreview.

\subsection{Second Head--Backbone Pair (Exploratory)}
\label{app:second}

Microsoft states that Decision-1 is post-trained from Qwen3.5-9B, which we cannot verify. We added the comparison after the \fresh{} pre-registration, so it is exploratory.
Among the remaining hosted models, only Decision-1 has a declared backbone available with log-probabilities through our provider. We did not run Qwen3.5-4B, the public base of Kev-4B.
We analyse the pair with the estimands and bootstrap of the three main pairs. Unlike them, its registration names the hosted readout of Qwen3.5-9B as primary and the local readout as sensitivity, and we report both. Clef-flash shares the backbone, so the two pairs are not independent.
On \coretwelve{}, the low-minus-high accuracy interaction $\bar D_1$ is \srLocalDOne{} \srLocalDOneCI{} accuracy points with the local backbone readout and \srHostedDOne{} \srHostedDOneCI{} with the hosted one, and the gain's Spearman correlation with log web share is \srLocalRho{} for both.
The low-minus-high difference in the share of backbone errors removed, $\bar D_2$, is \srLocalDTwo{} \srLocalDTwoCI{} locally and \srHostedDTwo{} \srHostedDTwoCI{} through the API.
On the \fresh{} the local and hosted $\bar D_1$ are \srFreshLocalDOne{} \srFreshLocalDOneCI{} and \srFreshHostedDOne{} \srFreshHostedDOneCI{}, with $\bar D_2$ of \srFreshLocalDTwo{} \srFreshLocalDTwoCI{} and \srFreshHostedDTwo{} \srFreshHostedDTwoCI{}.
On Wide-Belebele, the head's accuracy gain has an interval above zero in \srBbLocalAbove{} of \srBbLanguages{} languages under the local readout (\srBbHostedAbove{} hosted) and below zero in none, with a median gain of \srBbLocalMedian{} points.

\section{Confidence-Policy Transfer}
\label{app:a3}

Table~\ref{tab:a3} applies the English-fitted policies of every hosted model with English development runs to \coretwelve{} XNLI and MASSIVE.
For models with exact zeros, we floor probabilities before temperature fitting, at $10^{-6}$ in the primary analysis.
Across Jev, d1 and pplx-decider, both datasets and twelve languages, a floor of $10^{-4}$ moves the realised risk at the 5\% threshold by at most \aThreeFloorFourRiskMax{} points and conformal coverage by at most \aThreeFloorFourCovMax{}. With $10^{-8}$ the largest shifts are \aThreeFloorEightRiskMax{} and \aThreeFloorEightCovMax{} points.
Rounding every model to two decimals before fitting (mass-preserving, available for Jev and d1) changes risk and coverage by no more than \aThreeRoundedRiskMax{} and \aThreeRoundedCovMax{} points.
The XNLI conformal sets also become empty for up to \aThreeXnliEmptySetOtherMax\% of decisions in a non-English language, against at most \aThreeXnliEmptySetEnMax\% in English.

\begin{table*}[t]
  \centering
  \footnotesize
  \setlength{\tabcolsep}{3pt}
  \begin{tabular}{@{}l
      S[table-format=1.3]@{\ (}S[table-format=2.0, print-zero-integer=true]@{)\hspace{6pt}}
      S[table-format=1.3]@{\ (}S[table-format=2.0, print-zero-integer=true]@{)\hspace{6pt}}
      S[table-format=1.3]@{\ (}S[table-format=2.0, print-zero-integer=true]@{)\hspace{10pt}}
      S[table-format=0.2]@{/}S[table-format=1.2, print-zero-integer=true]
      S[table-format=0.2]@{/}S[table-format=1.2, print-zero-integer=true]
      S[table-format=0.2]@{/}S[table-format=1.2, print-zero-integer=true]@{}}
    \toprule
    & \multicolumn{6}{c}{Risk at 5\% (admitted \%)} & \multicolumn{6}{c@{}}{90\% set: coverage/size} \\
    \cmidrule(r{10pt}){2-7}\cmidrule(l){8-13}
    Model & \multicolumn{2}{c}{En} & \multicolumn{2}{c}{High} & \multicolumn{2}{c}{Low} & \multicolumn{2}{c}{En} & \multicolumn{2}{c}{High} & \multicolumn{2}{c@{}}{Low} \\
    \midrule
    \tgroup{13}{XNLI} \\
    Jev          & \atJevXnliEnRisk & \atJevXnliEnAdm & \atJevXnliHighRisk & \atJevXnliHighAdm & \atJevXnliLowRisk & \atJevXnliLowAdm & \atJevXnliEnCov & \atJevXnliEnSize & \atJevXnliHighCov & \atJevXnliHighSize & \atJevXnliLowCov & \atJevXnliLowSize \\
    pplx-decider & \atPplxXnliEnRisk & \atPplxXnliEnAdm & \atPplxXnliHighRisk & \atPplxXnliHighAdm & \atPplxXnliLowRisk & \atPplxXnliLowAdm & \atPplxXnliEnCov & \atPplxXnliEnSize & \atPplxXnliHighCov & \atPplxXnliHighSize & \atPplxXnliLowCov & \atPplxXnliLowSize \\
    d1           & \atLiquidXnliEnRisk & \atLiquidXnliEnAdm & \atLiquidXnliHighRisk & \atLiquidXnliHighAdm & \atLiquidXnliLowRisk & \atLiquidXnliLowAdm & \atLiquidXnliEnCov & \atLiquidXnliEnSize & \atLiquidXnliHighCov & \atLiquidXnliHighSize & \atLiquidXnliLowCov & \atLiquidXnliLowSize \\
    Mercury      & \atMercuryXnliEnRisk & \atMercuryXnliEnAdm & \atMercuryXnliHighRisk & \atMercuryXnliHighAdm & \atMercuryXnliLowRisk & \atMercuryXnliLowAdm & \atMercuryXnliEnCov & \atMercuryXnliEnSize & \atMercuryXnliHighCov & \atMercuryXnliHighSize & \atMercuryXnliLowCov & \atMercuryXnliLowSize \\
    Kev-4B       & \atKevXnliEnRisk & \atKevXnliEnAdm & \atKevXnliHighRisk & \atKevXnliHighAdm & \atKevXnliLowRisk & \atKevXnliLowAdm & \atKevXnliEnCov & \atKevXnliEnSize & \atKevXnliHighCov & \atKevXnliHighSize & \atKevXnliLowCov & \atKevXnliLowSize \\
    Luna         & {\tmark{--}} & \atLunaXnliEnAdm & {\tmark{--}} & \atLunaXnliHighAdm & {\tmark{--}} & \atLunaXnliLowAdm & \atLunaXnliEnCov & \atLunaXnliEnSize & \atLunaXnliHighCov & \atLunaXnliHighSize & \atLunaXnliLowCov & \atLunaXnliLowSize \\
    Decision-1   & \atMsftXnliEnRisk & \atMsftXnliEnAdm & \atMsftXnliHighRisk & \atMsftXnliHighAdm & \atMsftXnliLowRisk & \atMsftXnliLowAdm & \atMsftXnliEnCov & \atMsftXnliEnSize & \atMsftXnliHighCov & \atMsftXnliHighSize & \atMsftXnliLowCov & \atMsftXnliLowSize \\
    Solar        & \atSolarXnliEnRisk & \atSolarXnliEnAdm & \atSolarXnliHighRisk & \atSolarXnliHighAdm & \atSolarXnliLowRisk & \atSolarXnliLowAdm & \atSolarXnliEnCov & \atSolarXnliEnSize & \atSolarXnliHighCov & \atSolarXnliHighSize & \atSolarXnliLowCov & \atSolarXnliLowSize \\
    \addlinespace
    \tgroup{13}{MASSIVE} \\
    Jev          & \atJevMassiveEnRisk & \atJevMassiveEnAdm & \atJevMassiveHighRisk & \atJevMassiveHighAdm & \atJevMassiveLowRisk & \atJevMassiveLowAdm & \atJevMassiveEnCov & \atJevMassiveEnSize & \atJevMassiveHighCov & \atJevMassiveHighSize & \atJevMassiveLowCov & \atJevMassiveLowSize \\
    pplx-decider & \atPplxMassiveEnRisk & \atPplxMassiveEnAdm & \atPplxMassiveHighRisk & \atPplxMassiveHighAdm & \atPplxMassiveLowRisk & \atPplxMassiveLowAdm & \atPplxMassiveEnCov & \atPplxMassiveEnSize & \atPplxMassiveHighCov & \atPplxMassiveHighSize & \atPplxMassiveLowCov & \atPplxMassiveLowSize \\
    d1           & \atLiquidMassiveEnRisk & \atLiquidMassiveEnAdm & \atLiquidMassiveHighRisk & \atLiquidMassiveHighAdm & \atLiquidMassiveLowRisk & \atLiquidMassiveLowAdm & \atLiquidMassiveEnCov & \atLiquidMassiveEnSize & \atLiquidMassiveHighCov & \atLiquidMassiveHighSize & \atLiquidMassiveLowCov & \atLiquidMassiveLowSize \\
    Mercury      & \atMercuryMassiveEnRisk & \atMercuryMassiveEnAdm & \atMercuryMassiveHighRisk & \atMercuryMassiveHighAdm & \atMercuryMassiveLowRisk & \atMercuryMassiveLowAdm & \atMercuryMassiveEnCov & \atMercuryMassiveEnSize & \atMercuryMassiveHighCov & \atMercuryMassiveHighSize & \atMercuryMassiveLowCov & \atMercuryMassiveLowSize \\
    Kev-4B       & \atKevMassiveEnRisk & \atKevMassiveEnAdm & \atKevMassiveHighRisk & \atKevMassiveHighAdm & \atKevMassiveLowRisk & \atKevMassiveLowAdm & \atKevMassiveEnCov & \atKevMassiveEnSize & \atKevMassiveHighCov & \atKevMassiveHighSize & \atKevMassiveLowCov & \atKevMassiveLowSize \\
    Decision-1   & \atMsftMassiveEnRisk & \atMsftMassiveEnAdm & \atMsftMassiveHighRisk & \atMsftMassiveHighAdm & \atMsftMassiveLowRisk & \atMsftMassiveLowAdm & \atMsftMassiveEnCov & \atMsftMassiveEnSize & \atMsftMassiveHighCov & \atMsftMassiveHighSize & \atMsftMassiveLowCov & \atMsftMassiveLowSize \\
    Luna         & \atLunaMassiveEnRisk & \atLunaMassiveEnAdm & \atLunaMassiveHighRisk & \atLunaMassiveHighAdm & \atLunaMassiveLowRisk & \atLunaMassiveLowAdm & \atLunaMassiveEnCov & \atLunaMassiveEnSize & \atLunaMassiveHighCov & \atLunaMassiveHighSize & \atLunaMassiveLowCov & \atLunaMassiveLowSize \\
    \bottomrule
  \end{tabular}
  \tabnote{\emph{High}: mean over German, Spanish, French, Russian and Chinese. \emph{Low}: mean over Hindi, Swahili, Thai and Urdu. Risk: error rate among admitted decisions, with the admitted share of the 500 decisions per language in parentheses. \tmark{--}: no decision admitted. Coverage counts all scheduled decisions, and size is the mean prediction-set size.}
  \caption{\textbf{English-fitted policies miss their targets on XNLI outside English.} Policies fitted on English development data and applied unchanged on \coretwelve{}, for every hosted model with English development runs, with a floor of $10^{-6}$. Solar cannot be fitted on MASSIVE, where its option cap leaves no valid English development decision.}
  \label{tab:a3}
\end{table*}

Figure~\ref{fig:a3} omits the cells that cannot be assessed. Luna's English XNLI threshold admits no decision and Solar's admits 1 to 4\% per language, so neither is drawn on XNLI, and Solar has no MASSIVE fit.

\section{Held-out Cohort Results}
\label{app:fresh}

\paragraph{Confirmatory tests.}
The paired stratified bootstrap draws \freshReps{} replicates over \freshClusters{} parent groups.
Under the local readout, pooled $\bar D_1$ is \freshDOne{} accuracy points, with a one-sided bound of \freshDOneBound{} at the Holm level ($p=\freshHOneP$, tested at \freshHolmFirstLevel{}).
Pooled $\bar D_2$ is \freshDTwo{}, with a \freshDTwoLevel\% interval of \freshDTwoCI{} (TOST $p=\freshHTwoP$, tested at \freshHolmSecondLevel{}).
Under the hosted readout, which is exploratory, $\bar D_1$ is \freshHostedDOne{} points (bound \freshHostedDOneBound{}, $p=\freshHostedHOneP$) and $\bar D_2$ is \freshHostedDTwo{} \freshHostedDTwoCI{}, with TOST $p=\freshHostedHTwoP$, so equivalence is not shown.

\paragraph{Per pair (exploratory, local readout).}
The accuracy interaction is \freshDOneClef{} \freshDOneClefCI{} points for Clef, \freshDOneClefFlash{} \freshDOneClefFlashCI{} for Clef-flash and \freshDOnePplx{} \freshDOnePplxCI{} for pplx-decider.
The difference in the share of backbone errors removed is \freshDTwoClef{} \freshDTwoClefCI{}, \freshDTwoClefFlash{} \freshDTwoClefFlashCI{} and \freshDTwoPplx{} \freshDTwoPplxCI{}.
For the Brier gain, the low-minus-high interaction is \freshBrierGainInteractionClef{} \freshBrierGainInteractionClefCI{}, \freshBrierGainInteractionClefFlash{} \freshBrierGainInteractionClefFlashCI{} and \freshBrierGainInteractionPplx{} \freshBrierGainInteractionPplxCI{}.
The ECE gain is smaller in the low-web-share group for all three pairs.
The interactions are \freshEceGainInteractionClef{} \freshEceGainInteractionClefCI{} for Clef, \freshEceGainInteractionClefFlash{} \freshEceGainInteractionClefFlashCI{} for Clef-flash and \freshEceGainInteractionPplx{} \freshEceGainInteractionPplxCI{} for pplx-decider.
Across the twelve languages, the Spearman correlation of the accuracy gain with log web share is \freshWebRhoClef{} for Clef, \freshWebRhoClefFlash{} for Clef-flash and \freshWebRhoPplx{} for pplx-decider, with permutation $p$ values of \freshWebPClef{}, \freshWebPClefFlash{} and \freshWebPPplx{}.

\paragraph{Skill ranking and policy transfer (exploratory).}
On the held-out cohort, Brier skill against chance runs from \skFreshLo{} to \skFreshHi{} over all complete arms, including backbones and the two MediaPipe arms. The MediaPipe embedding arm scores \skMediaPipeFresh{}, and GLiNER scores \skGlinerFresh{} at \covGlinerFresh\% coverage, missing only cases refused by the 512-token window.
Over the \nFreshRanked{} Typed Decision Models, the held-out ranking agrees with \coretwelve{} at Kendall $\tau=\tauFreshVsCore$, with pplx-decider first on both. Decision-1 was registered after the pre-registration, so its held-out rows are exploratory.
We also apply the English-fitted 5\% thresholds unchanged to held-out XNLI. For the six assessable models, risk is \freshAThreeXnliEnRiskLo--\freshAThreeXnliEnRiskHi\% in English and \freshAThreeXnliDeFrRiskLo--\freshAThreeXnliDeFrRiskHi\% in German and French. Conformal coverage in the low-web-share group is \freshAThreeXnliLowCovLo--\freshAThreeXnliLowCovHi{} against a target of .90.

\section{Reproducibility and Compute}
\label{app:compute}

\paragraph{Fixed artefacts.}
The statistical protocol, manifests and model registrations are identified by SHA-256 hashes in the release, and Table~\ref{tab:chrono} gives the order in which they were fixed. Every manifest rebuilds identical case bytes from pinned source revisions (Table~\ref{tab:sources}).
Each suite score comes with a signature that records the suite version and hash, manifest hashes, scorer commit, reference, floor, refusal and bootstrap settings \citep{post2018sacrebleu}.
Hosted models are reached through one decision router and one generic adapter, and a response is accepted only if it names the registered dated snapshot.

\begin{table}[t]
  \centering
  \footnotesize
  \setlength{\tabcolsep}{3pt}
  \renewcommand{\arraystretch}{1.15}
  \begin{tabular}{@{}l>{\raggedright\arraybackslash}p{0.55\columnwidth}@{}}
    \toprule
    Time (UTC, 2026) & Event \\
    \midrule
    10-04 00:56 & Statistical protocol frozen, with accuracy as the primary endpoint \\
    10-04 01:23--03:58 & \coretwelve{} and English development manifests \\
    10-06 20:25 & \Fresh{} manifest frozen \\
    10-07 14:42 & Described-label manifest \\
    10-07 14:43--10-08 20:41 & Model registrations (Laya's on 10-04), each before that model's first benchmark request \\
    10-07 14:44--10-08 20:38 & \coretwelve{} inference \\
    10-07 20:14 & Head--backbone analysis code committed \\
    10-08 01:23 & First pair with both \coretwelve{} runs complete \\
    10-08 16:04--17:04 & Manifests of the other eight suites \\
    10-08 17:10 & Suite scoring contract: Brier skill primary, breadth manifest hashes pinned \\
    10-08 19:18--20:38 & Hosted Clef, Clef-flash and Qwen \coretwelve{} arms \\
    10-08 19:18--10-09 00:09 & Inference on the other eight suites \\
    10-08 22:58--23:00 & Final English temperature and transfer fits \\
    10-09 12:00 & \Fresh{} pre-registration, before the first \fresh{} request \\
    10-10 10:20 & Decision-1 registered, after the \fresh{} pre-registration \\
    10-10 10:20--15:34 & Decision-1 inference on every suite and the \fresh{} \\
    \bottomrule
  \end{tabular}
  \caption{\textbf{Protocol, manifests and registrations were fixed before the requests they cover.} Chronology from git commits, manifest creation times and run logs. Laya's \coretwelve{} manifest (10-08 08:20) has the same case identifiers and input hashes as the 10-04 manifest, and Laya's breadth manifests date from 10-08 19:27. Some registrations followed synthetic probe and smoke requests.}
  \label{tab:chrono}
\end{table}

\paragraph{Repeat runs and hosted--local agreement.}
The described-label runs re-send the 8,448 \coretwelve{} MASSIVE and SIB-200 cases with unchanged case identifiers and input hashes on the same day.
Against the headline answers, top-1 agreement is 1.000 for pplx-decider and Luna, .9995 for Clef, .9996 for Clef-flash, .997 for Mercury, .995 for Kev-4B, .990 for d1, .989 for Decision-1, .980 for Jev and .965 for Solar (SIB-200 only).
Dataset Brier skill changes by at most .0024 (Solar).
Decision-1's API is not deterministic. Twelve identical probe requests spread by up to \msftRepeatDp{} on one option, and only \retestIdenticalMsft\% of its repeated probability vectors are identical, yet it keeps \retestTopOneMsft\% of its top answers and moves dataset skill by at most \retestBrierMsft{}.
The 99th percentile of the largest per-case probability difference is .005 for both hosted Clef heads, at most .148 for the other models and .973 for Solar.
Hosted and local Clef agree on the top answer for 99.14\% of the 26,460 \coretwelve{} cases and Clef-flash for 99.48\%, with median largest probability difference .0014 for both.
Eight fixed probe requests sent at the start and end of the hosted runs differ by at most \retestSentinelMaxDp{}, for Decision-1, whose probes bracket all its suites. Solar's differ by up to .016. The hosted Clef routes have start probes only, pplx-decider's end probes precede its last \coretwelve{} attempt, and no probes bracket the other models' runs on the other eight suites.

\paragraph{Bootstrap settings.}
The head--backbone analysis uses 10,000 paired replicates, and suite scores use 1,000 joint replicates at the 95\% level, both with seed 20261004.
Transfer intervals use the same number of replicates and seed. Per dataset, one draw of source clusters (500 MASSIVE utterances, 455 XNLI premises) is shared by English, every target language and every model, and the English fit is held fixed.
Resampling each language independently would understate the standard error of Jev's \coretwelve{} Brier skill about threefold (0.0037 against 0.0110).
SIB-200 and Belebele share text. Of the 204 SIB English test sentences, 176 occur verbatim in a Belebele passage, and 107 of them appear in 87 of the 276 passages of the \coretwelve{} Belebele sample.
The shipped bootstrap resamples the two datasets independently.
Resampling each linked passage together with its SIB sentences as one parent, with 2,000 replicates, changes the standard error of \coretwelve{} Brier skill by a factor of \sibLinkSeRatioLo{} to \sibLinkSeRatioHi{} across \sibLinkModels{} models.
These are local Clef and Clef-flash, the two Qwen backbones and every hosted Typed Decision Model except the two hosted Clef routes.
For Jev the interval moves from [.6664, .7102] to [.6655, .7076].

\paragraph{Power of the held-out tests.}
Our power simulation (10,000 draws) takes, per task, as many \coretwelve{} source clusters as the \fresh{} has parent groups, \freshParentGroups{} in total, and shares one draw among the three pairs.
Two of the pairs share the Qwen3.8-27B backbone and all share items, so the pooled estimate is one correlated test, with standard error \hTwoStandardError{} for H2.
H2 uses two one-sided tests at $\alpha=\hTwoAlphaPerSide$ per side, which needs the 90\% interval inside $\pm\hTwoMargin$. Its power is \hTwoPower\% at a true difference of zero and \hTwoPowerAtCoreEstimate\% at the \coretwelve{} estimate. A margin of $\pm\hTwoMarginEightyPower$ would give 80\% power, and a margin of $\pm .05$ almost none.
Under Holm, H2 is tested at \hTwoAlphaPerSide{} once H1 has been rejected. If H2 has the smaller $p$-value and is tested first at .025, its power at zero is \hTwoPowerHolmFirst\%.
H1 has \hOnePower\% power at the \coretwelve{} effect, or \hOnePowerHolm\% at .025.
Per pair, H2 power at $\pm\hTwoMargin$ is only \hTwoPowerPerPairLo{} to \hTwoPowerPerPairHi\%, so we report per-pair values only descriptively.
At the low-web-share group's \coretwelve{} backbone error rates, \hTwoBackboneErrorTwentySeven{} for Qwen3.8-27B and \hTwoBackboneErrorNine{} for Qwen3.5-9B, the margin corresponds to about \hTwoMarginPointsTwentySeven{} and \hTwoMarginPointsNine{} accuracy points.

\paragraph{Scoring a new model.}
One command, \texttt{xdecidebench eval -{}-model} \textit{model}, scores a new model on the benchmark.
It rebuilds the cases of every suite the chosen tracks need from the pinned sources, runs the model, audits the runs and writes a report per suite with Brier skill, valid-output coverage and a 95\% interval under a signature that fixes how it was computed.
It then compares the scores with the published models.
The model can be a hosted route, an OpenAI-compatible server, a local Hugging Face checkpoint or a custom adapter that returns a full probability vector for each scheduled question. Every case is scored, including refusals.
Guides for scoring a new model and extending the benchmark will be released with the code\afterreview\releaseat.

\paragraph{Compute and cost.}
Hosted API spend over every run in the run stores is \$\costApiTotal{}: \$\costApiBenchmark{} for the benchmark suites, \$\costApiFresh{} for the \fresh{}, \$\costApiExploratory{} for the exploratory interface-language ablation and \$\costApiPilot{} for pilots (Table~\ref{tab:cost}).
The ten hosted Typed Decision Models account for \$\costApiDecision{}, at most \$\costApiDecisionMax{} each, and the two hosted Qwen readouts for \$\costApiBackbones{}.
Per attempt, the cost is the billed total of the provider's generation records where these were fetched, and otherwise the cost the response reported. The providers reported \costTokensInM{}M input and \costTokensOutM{}M output tokens.
Probe requests are not in the run stores and are not counted.
Local Clef, Clef-flash and the Qwen3.8-27B backbone took \costHHundredHours{} H100 GPU-hours, summed from per-run wall-clock times in the run receipts. For one Qwen3.8-27B run, this is end-to-end time including an interruption.
On one RTX 4090, Laya, the Qwen3.5-9B backbone and mmBERT inference took an estimated \costRtxHours{} hours and mmBERT fine-tuning \costMmbertTrainHours{} hours, and the two MediaPipe arms took an estimated \costCpuHours{} wall-clock hours on CPU.
The RTX 4090 inference and CPU figures are estimated from attempt start times with gaps over ten minutes removed, since usage was not metered.

\begin{table*}[t]
\centering
\footnotesize
\setlength{\tabcolsep}{4pt}
\begin{tabular}{@{}lS[table-format=2.2, print-zero-integer]S[table-format=1.2, print-zero-integer]S[table-format=1.2, print-zero-integer]S[table-format=1.2, print-zero-integer]S[table-format=2.2, print-zero-integer]S[table-format=3.1, print-zero-integer]S[table-format=2.1, print-zero-integer]@{}}
  \toprule
   & \multicolumn{5}{c}{Spend (US\$)} & \multicolumn{2}{c}{Tokens (M)} \\
  \cmidrule(lr){2-6}\cmidrule(lr){7-8}
  Hosted arm & {Bench.} & {\Fresh{}} & {Explor.} & {Pilot} & {Total} & {In} & {Out} \\
  \midrule
  Qwen3.8-27B, hosted & 40.52 & 1.43 & 0.00 & 0.00 & 41.95 & 42.1 & 0.2 \\
  Luna & 5.75 & 0.19 & 0.72 & 0.00 & 6.66 & 66.6 & 0.0 \\
  Clef & 3.34 & 0.14 & 2.90 & 0.00 & 6.38 & 85.1 & 0.0 \\
  Jev & 3.97 & 0.16 & 0.58 & 0.35 & 5.06 & 120.5 & 26.5 \\
  Solar & 4.73 & 0.10 & 0.09 & 0.00 & 4.92 & 98.5 & 0.2 \\
  Qwen3.5-9B, hosted & 4.11 & 0.14 & 0.24 & 0.00 & 4.49 & 44.4 & 0.2 \\
  Clef-flash & 1.72 & 0.07 & 0.48 & 0.00 & 2.27 & 85.7 & 0.0 \\
  Kev-4B & 1.76 & 0.06 & 0.11 & 0.00 & 1.93 & 46.0 & 45.6 \\
  d1 & 1.49 & 0.05 & 0.16 & 0.00 & 1.70 & 42.6 & 0.0 \\
  Decision-1 & 1.47 & 0.06 & 0.11 & 0.00 & 1.64 & 39.1 & 0.2 \\
  pplx-decider & 1.08 & 0.04 & 0.14 & 0.00 & 1.25 & 62.6 & 0.2 \\
  Mercury & 0.00 & 0.00 & 0.00 & 0.00 & 0.00 & 63.7 & 0.3 \\
  \midrule
  \textit{Total} & 69.94 & 2.44 & 5.53 & 0.35 & 78.26 & 796.8 & 73.5 \\
  \bottomrule
\end{tabular}
\par\medskip
\begin{tabular}{@{}llS[table-format=2.1, print-zero-integer]@{}}
  \toprule
  Local arm & Device & {Hours} \\
  \midrule
  Clef, local & H100 & 3.9 \\
  Clef-flash, local & H100 & 2.2 \\
  Qwen3.8-27B, local & H100 & 35.3 \\
  Laya & RTX 4090 & 0.9 \\
  Qwen3.5-9B, local & RTX 4090 & 13.2 \\
  mmBERT (inference) & RTX 4090 & 0.0 \\
  MP-EG2 & CPU & 11.4 \\
  MP-GLiNER & CPU & 45.8 \\
  \bottomrule
\end{tabular}
\tabnote{H100 hours from run receipts; RTX 4090 inference and CPU hours estimated from attempt start times (gaps over ten minutes removed). Mercury was served at no charge.}
\caption{\textbf{Hosted API spend, tokens and local compute hours.} Spend in US dollars by run group, and the input and output tokens the providers reported, in millions (top); local compute hours (bottom).}
\label{tab:cost}
\end{table*}

\section{Additional Result Tables}
\label{app:tables}

\paragraph{Reading the full-results tables.}
Tables~\ref{tab:failzero} to~\ref{tab:perdataset} report failures, suite scores and per-dataset scores for fourteen arms of Table~\ref{tab:matrix}: the ten hosted Typed Decision Models, the hosted readouts of the two backbones and the two MediaPipe comparators. They use the groups, order and short names of that table, with hosted Typed Decision Models ordered by \coretwelve{} Brier skill. Failures are scored as uniform. Bold marks the best hosted Typed Decision Model in each column and underline the runner-up, by point estimate. Uncertainty is printed in grey at the size of the estimate: an interval in brackets is a 95\% interval, never a half-width. Cells without a score are grey (--: no value).
Table~\ref{tab:fullsuite} gives each suite score under both refusal views, with the constant-predictor baseline, and Table~\ref{tab:perdataset} the per-dataset skills of the multi-dataset suites. Per-language scores of every arm on every suite, the per-language head--backbone differences and the per-language policy-transfer results will be released with the code\afterreview.

\begin{table*}[tp]
\centering
\footnotesize
\setlength{\tabcolsep}{5pt}
\begin{tabular}{@{}lS[table-format=4, print-zero-integer]@{\hspace{6pt}}S[table-format=2.1, print-zero-integer]S[table-format=4, print-zero-integer]@{\hspace{6pt}}S[table-format=2.1, print-zero-integer]S[table-format=4, print-zero-integer]@{\hspace{6pt}}S[table-format=3.1, print-zero-integer]S[table-format=1, print-zero-integer]@{\hspace{6pt}}S[table-format=2.1, print-zero-integer]S[table-format=3, print-zero-integer]@{\hspace{6pt}}S[table-format=2.1, print-zero-integer]@{}}
  \toprule
  (a) & \multicolumn{8}{c}{Aligned, shared labels} & \multicolumn{2}{c}{Locale labels} \\
  \cmidrule(lr){2-9}\cmidrule(lr){10-11}
   & \multicolumn{2}{c}{\coretwelve{}} & \multicolumn{2}{c}{Belebele} & \multicolumn{2}{c}{MASSIVE} & \multicolumn{2}{c}{NLI} & \multicolumn{2}{c}{RTP-LX} \\
  Model & {Fail.} & {Zero} & {Fail.} & {Zero} & {Fail.} & {Zero} & {Fail.} & {Zero} & {Fail.} & {Zero} \\
  \midrule
  \tgroup{11}{Hosted Typed Decision Models} \\
  pplx-decider & 0 & 0.0 & 0 & 0.0 & 0 & 0.0 & 0 & 0.0 & 0 & 0.0 \\
  Jev & 0 & 86.5 & 0 & 75.6 & 0 & 100.0 & 0 & 72.7 & 0 & 93.1 \\
  Clef & 0 & 0.0 & 0 & 0.0 & 0 & 0.0 & 0 & 0.0 & 8 & 0.0 \\
  d1 & 0 & 0.0 & 0 & 0.0 & 0 & 0.0 & 0 & 0.0 & 0 & 0.0 \\
  Decision-1 & 0 & 0.0 & 0 & 0.0 & 0 & 0.6 & 0 & 0.0 & 0 & 0.0 \\
  Clef-flash & 0 & 0.0 & 0 & 0.0 & 0 & 0.0 & 0 & 0.0 & 8 & 0.0 \\
  Mercury & 0 & 0.0 & 0 & 0.0 & 0 & 0.0 & 0 & 0.0 & 0 & 0.0 \\
  Luna & 64 & 67.4 & 1 & 71.0 & 46 & 99.7 & 0 & 56.3 & 328 & 96.2 \\
  Kev-4B & 0 & 0.2 & 0 & 0.0 & 0 & 0.3 & 0 & 0.2 & 0 & 0.0 \\
  Solar & 6000 & 0.6 & 0 & 0.0 & 6084 & {\tmark{--}} & 0 & 0.4 & 0 & 0.0 \\
  \addlinespace[2pt]
  \tgroup{11}{Backbone readouts, hosted} \\
  Qwen3.8-27B & 0 & 0.0 & 0 & 0.0 & 0 & 0.0 & 0 & 0.0 & 0 & 0.0 \\
  Qwen3.5-9B & 0 & 0.0 & 0 & 0.0 & 0 & 0.0 & 0 & 0.0 & 0 & 0.0 \\
  \addlinespace[2pt]
  \tgroup{11}{Comparators} \\
  MP-EG2 & 0 & 0.0 & 0 & 0.0 & 0 & 0.0 & 0 & 0.0 & 0 & 0.0 \\
  MP-GLiNER & 2073 & 0.0 & 2202 & 0.0 & 0 & 0.0 & 0 & 0.0 & 0 & 0.0 \\
  \bottomrule
\end{tabular}
\par\medskip
\begin{tabular}{@{}lS[table-format=3, print-zero-integer]@{\hspace{6pt}}S[table-format=2.1, print-zero-integer]S[table-format=4, print-zero-integer]@{\hspace{6pt}}S[table-format=3.1, print-zero-integer]S[table-format=2, print-zero-integer]@{\hspace{6pt}}S[table-format=2.1, print-zero-integer]S[table-format=1, print-zero-integer]@{\hspace{6pt}}S[table-format=2.1, print-zero-integer]@{}}
  \toprule
  (b) & \multicolumn{8}{c}{Native-authored} \\
  \cmidrule(lr){2-9}
   & \multicolumn{2}{c}{INCLUDE} & \multicolumn{2}{c}{Injongo} & \multicolumn{2}{c}{Moder.} & \multicolumn{2}{c}{BRIGHTER} \\
  Model & {Fail.} & {Zero} & {Fail.} & {Zero} & {Fail.} & {Zero} & {Fail.} & {Zero} \\
  \midrule
  \tgroup{9}{Hosted Typed Decision Models} \\
  pplx-decider & 0 & 0.0 & 0 & 0.0 & 0 & 0.0 & 0 & 0.0 \\
  Jev & 0 & 51.2 & 0 & 100.0 & 0 & 31.8 & 0 & 77.6 \\
  Clef & 0 & 0.0 & 0 & 0.0 & 0 & 0.0 & 0 & 0.0 \\
  d1 & 0 & 0.0 & 0 & 0.0 & 0 & 0.0 & 0 & 0.0 \\
  Decision-1 & 0 & 0.0 & 0 & 0.0 & 0 & 0.0 & 0 & 0.0 \\
  Clef-flash & 0 & 0.0 & 0 & 0.0 & 0 & 0.0 & 0 & 0.0 \\
  Mercury & 0 & 0.0 & 0 & 0.0 & 0 & 0.0 & 0 & 0.0 \\
  Luna & 122 & 55.9 & 12 & 84.2 & 48 & 60.5 & 0 & 96.5 \\
  Kev-4B & 0 & 0.0 & 0 & 4.0 & 0 & 0.0 & 0 & 0.0 \\
  Solar & 0 & 0.1 & 1920 & {\tmark{--}} & 0 & 0.9 & 0 & 0.0 \\
  \addlinespace[2pt]
  \tgroup{9}{Backbone readouts, hosted} \\
  Qwen3.8-27B & 0 & 0.0 & 0 & 0.0 & 0 & 0.0 & 0 & 0.0 \\
  Qwen3.5-9B & 0 & 0.0 & 0 & 0.0 & 0 & 0.0 & 0 & 0.0 \\
  \addlinespace[2pt]
  \tgroup{9}{Comparators} \\
  MP-EG2 & 0 & 0.0 & 0 & 0.0 & 0 & 0.0 & 0 & 0.0 \\
  MP-GLiNER & 439 & 0.0 & 321 & 0.0 & 0 & 0.0 & 0 & 0.0 \\
  \bottomrule
\end{tabular}
\tabnote{\emph{Fail.}: items whose status is not a valid typed answer (refusals, provider errors and interface caps). \emph{Zero}: share of valid answers, in percent, with an exact-zero probability.}
\caption{\textbf{Failures and exact zeros per model and suite.} Failed items (scored as uniform) and the share of valid answers whose probability vector contains an exact zero: (a)~aligned suites; (b)~native suites.}
\label{tab:failzero}
\end{table*}

\begin{table*}[tp]
\centering
\footnotesize
\setlength{\tabcolsep}{3pt}
\begin{tabular}{@{}lS[table-format=-0.3]@{\,}lS[table-format=-0.3]@{\,}lS[table-format=0.3]S[table-format=-0.3]S[table-format=-1.3]S[table-format=-0.3]S[table-format=0.3]S[table-format=-0.3]S[table-format=-1.3]@{}}
  \toprule
   & \multicolumn{6}{c}{Aligned, shared labels} & \multicolumn{1}{c}{Locale labels} & \multicolumn{4}{c}{Native-authored} \\
  \cmidrule(lr){2-7}\cmidrule(lr){8-8}\cmidrule(lr){9-12}
  Model & \multicolumn{2}{c}{\coretwelve{}} & \multicolumn{2}{c}{Belebele} & {MASSIVE} & {NLI} & {RTP-LX} & {INCLUDE} & {Injongo} & {Moder.} & {BRIGHTER} \\
  \midrule
  \tgroup{12}{Hosted Typed Decision Models} \\
  pplx-decider & \bfseries .712 &  & .756\tunder{.756} &  & .718 & \bfseries .397 & .051 & .536\tunder{.536} & .578\tunder{.578} & \bfseries .673 & .144\tunder{.144} \\
  Jev & .688\tunder{.688} &  & \bfseries .785 &  & .619 & .128 & \minus.101 & .531 & .499 & .561\tunder{.561} & \minus.428 \\
  Clef & .674 &  & .701 &  & .714 & .254 & .045 & .459 & .492 & .274 & \bfseries .167 \\
  d1 & .663 &  & .711 &  & .634 & .315\tunder{.315} & .070\tunder{.070} & .488 & .481 & .449 & .118 \\
  Decision-1 & .651 &  & .716 &  & .721\tunder{.721} & .108 & \minus.229 & .400 & .417 & .500 & \minus.375 \\
  Clef-flash & .631 &  & .597 &  & \bfseries .745 & .127 & \bfseries .071 & .385 & .415 & .379 & .123 \\
  Mercury & .611 &  & .677 &  & .651 & .188 & \minus.022 & .411 & .416 & .449 & .013 \\
  Luna & .568 &  & .743 &  & .602 & \minus.058 & \minus.018 & \bfseries .600 & \bfseries .580 & .541 & \minus.096 \\
  Kev-4B & .449 &  & .389 &  & .371 & .070 & \minus.169 & .273 & .225 & .278 & \minus.275 \\
  Solar & .398 & /\,.497 & .334 &  & .000 & \minus.236 & \minus.568 & .251 & .000 & \minus.046 & \minus1.396 \\
  \addlinespace[2pt]
  \tgroup{12}{Backbone readouts, hosted} \\
  Qwen3.8-27B & .584 &  & .678 &  & .523 & .043 & \minus.336 & .426 & .405 & .397 & \minus.056 \\
  Qwen3.5-9B & .477 &  & .525 &  & .313 & \minus.036 & \minus.562 & .310 & .167 & .347 & \minus.034 \\
  \addlinespace[2pt]
  \tgroup{12}{Comparators} \\
  MP-EG2 & \minus.084 &  & \minus.284 &  & .347 & \minus.997 & \minus1.363 & \minus.242 & .054 & \minus.004 & \minus1.166 \\
  MP-GLiNER & .057 &  & \minus.066 & /\,\minus.068 & .029 & .010 & \minus.984 & \minus.034 & .032 & .076 & \minus.807 \\
  \midrule
  \textit{Baseline} & .013 &  & .024 &  & .000 & .001 & .000 & .007 & .000 & .001 & .000 \\
  \bottomrule
\end{tabular}
\tabnote{After a slash: the sensitivity view that drops capped cells, where any are dropped. \textit{Baseline}: the constant predictor that answers every item with its cell's label prior.}
\caption{\textbf{Suite Brier skill under both refusal views.} Each suite's skill against its declared reference with failures scored as uniform, and the constant-predictor baseline.}
\label{tab:fullsuite}
\end{table*}

\begin{table*}[tp]
\centering
\footnotesize
\setlength{\tabcolsep}{2pt}
\begin{tabular}{@{}lS[table-format=-0.3]@{\,}lS[table-format=-0.3]@{\,}lS[table-format=-0.3]@{\,}lS[table-format=-0.3]@{\,}lS[table-format=0.3]@{\,}l@{}}
  \toprule
  (a) & \multicolumn{10}{c}{\coretwelve{}} \\
  \cmidrule(lr){2-11}
  Model & \multicolumn{2}{c}{MASSIVE} & \multicolumn{2}{c}{XNLI} & \multicolumn{2}{c}{Belebele} & \multicolumn{2}{c}{MMLU-ProX} & \multicolumn{2}{c}{SIB-200} \\
  \midrule
  \tgroup{11}{Hosted Typed Decision Models} \\
  pplx-decider & .766 & \tci{[.729, .803]} & \bfseries .663 & \tci{[.619, .700]} & .887\tunder{.887} & \tci{[.861, .910]} & .450\tunder{.450} & \tci{[.402, .498]} & .794 & \tci{[.735, .844]} \\
  Jev & .651 & \tci{[.608, .693]} & .482 & \tci{[.424, .536]} & .883 & \tci{[.851, .911]} & \bfseries .584 & \tci{[.535, .634]} & .839\tunder{.839} & \tci{[.783, .888]} \\
  Clef & \bfseries .776 & \tci{[.739, .811]} & .572 & \tci{[.525, .615]} & .853 & \tci{[.819, .880]} & .363 & \tci{[.318, .404]} & .808 & \tci{[.749, .868]} \\
  d1 & .712 & \tci{[.676, .748]} & .601\tunder{.601} & \tci{[.550, .645]} & .856 & \tci{[.824, .883]} & .358 & \tci{[.313, .406]} & .786 & \tci{[.733, .834]} \\
  Decision-1 & .766 & \tci{[.731, .800]} & .447 & \tci{[.389, .500]} & \bfseries .922 & \tci{[.903, .938]} & .263 & \tci{[.213, .312]} & \bfseries .857 & \tci{[.816, .895]} \\
  Clef-flash & .774\tunder{.774} & \tci{[.735, .811]} & .493 & \tci{[.443, .538]} & .786 & \tci{[.751, .816]} & .301 & \tci{[.261, .340]} & .801 & \tci{[.748, .853]} \\
  Mercury & .684 & \tci{[.640, .728]} & .501 & \tci{[.441, .556]} & .855 & \tci{[.821, .885]} & .254 & \tci{[.194, .312]} & .764 & \tci{[.689, .829]} \\
  Luna & .608 & \tci{[.570, .643]} & .243 & \tci{[.168, .310]} & .854 & \tci{[.820, .884]} & .326 & \tci{[.282, .372]} & .809 & \tci{[.748, .864]} \\
  Kev-4B & .434 & \tci{[.407, .458]} & .397 & \tci{[.337, .459]} & .471 & \tci{[.416, .526]} & .182 & \tci{[.154, .212]} & .763 & \tci{[.716, .808]} \\
  Solar & .000 & \tci{[.000, .000]} & .164 & \tci{[.082, .244]} & .683 & \tci{[.638, .721]} & .392 & \tci{[.339, .445]} & .750 & \tci{[.684, .812]} \\
  \addlinespace[2pt]
  \tgroup{11}{Backbone readouts, hosted} \\
  Qwen3.8-27B & .592 & \tci{[.545, .637]} & .435 & \tci{[.370, .493]} & .840 & \tci{[.805, .870]} & .238 & \tci{[.190, .288]} & .817 & \tci{[.757, .872]} \\
  Qwen3.5-9B & .441 & \tci{[.395, .486]} & .278 & \tci{[.211, .339]} & .725 & \tci{[.686, .759]} & .125 & \tci{[.081, .167]} & .814 & \tci{[.759, .865]} \\
  \addlinespace[2pt]
  \tgroup{11}{Comparators} \\
  MP-EG2 & .519 & \tci{[.486, .555]} & \minus.996 & \tci{[\minus1.112, \minus.876]} & \minus.280 & \tci{[\minus.327, \minus.233]} & \minus.044 & \tci{[\minus.060, \minus.029]} & .379 & \tci{[.321, .442]} \\
  MP-GLiNER & .132 & \tci{[.110, .154]} & .033 & \tci{[.015, .052]} & \minus.060 & \tci{[\minus.097, \minus.020]} & \minus.068 & \tci{[\minus.086, \minus.051]} & .248 & \tci{[.191, .308]} \\
  \bottomrule
\end{tabular}
\par\medskip
\begin{tabular}{@{}lS[table-format=-0.3]@{\,}lS[table-format=-0.3]@{\,}lS[table-format=-0.3]@{\,}lS[table-format=-0.3]@{\,}lS[table-format=0.3]@{\,}l@{}}
  \toprule
  (b) & \multicolumn{6}{c}{NLI} & \multicolumn{4}{c}{Moder.} \\
  \cmidrule(lr){2-7}\cmidrule(lr){8-11}
  Model & \multicolumn{2}{c}{XNLI} & \multicolumn{2}{c}{AfriXNLI} & \multicolumn{2}{c}{AmericasNLI} & \multicolumn{2}{c}{HateCheck} & \multicolumn{2}{c}{Aya red-teaming} \\
  \midrule
  \tgroup{11}{Hosted Typed Decision Models} \\
  pplx-decider & \bfseries .663 & \tci{[.619, .700]} & \bfseries .436 & \tci{[.395, .476]} & \bfseries .094 & \tci{[.057, .132]} & \bfseries .797 & \tci{[.767, .823]} & \bfseries .549 & \tci{[.513, .585]} \\
  Jev & .482 & \tci{[.424, .536]} & .254 & \tci{[.204, .303]} & \minus.351 & \tci{[\minus.401, \minus.295]} & .596 & \tci{[.561, .629]} & .527 & \tci{[.487, .570]} \\
  Clef & .572 & \tci{[.525, .615]} & .298 & \tci{[.252, .341]} & \minus.109 & \tci{[\minus.143, \minus.078]} & .018 & \tci{[\minus.030, .073]} & .531\tunder{.531} & \tci{[.498, .570]} \\
  d1 & .601\tunder{.601} & \tci{[.550, .645]} & .374\tunder{.374} & \tci{[.332, .414]} & \minus.030\tunder{\minus.030} & \tci{[\minus.068, .005]} & .413 & \tci{[.373, .452]} & .486 & \tci{[.451, .524]} \\
  Decision-1 & .447 & \tci{[.389, .500]} & .127 & \tci{[.081, .171]} & \minus.251 & \tci{[\minus.296, \minus.205]} & .509 & \tci{[.462, .552]} & .492 & \tci{[.458, .532]} \\
  Clef-flash & .493 & \tci{[.443, .538]} & .140 & \tci{[.095, .183]} & \minus.252 & \tci{[\minus.285, \minus.218]} & .271 & \tci{[.218, .323]} & .487 & \tci{[.452, .526]} \\
  Mercury & .501 & \tci{[.441, .556]} & .234 & \tci{[.182, .282]} & \minus.171 & \tci{[\minus.220, \minus.124]} & .418 & \tci{[.373, .468]} & .479 & \tci{[.434, .524]} \\
  Luna & .243 & \tci{[.168, .310]} & .153 & \tci{[.090, .206]} & \minus.569 & \tci{[\minus.628, \minus.501]} & .605\tunder{.605} & \tci{[.555, .648]} & .478 & \tci{[.438, .520]} \\
  Kev-4B & .397 & \tci{[.337, .459]} & .063 & \tci{[.011, .110]} & \minus.251 & \tci{[\minus.297, \minus.206]} & .162 & \tci{[.117, .200]} & .393 & \tci{[.372, .417]} \\
  Solar & .164 & \tci{[.082, .244]} & \minus.345 & \tci{[\minus.381, \minus.307]} & \minus.527 & \tci{[\minus.554, \minus.498]} & \minus.406 & \tci{[\minus.518, \minus.307]} & .315 & \tci{[.262, .373]} \\
  \addlinespace[2pt]
  \tgroup{11}{Backbone readouts, hosted} \\
  Qwen3.8-27B & .435 & \tci{[.370, .493]} & .156 & \tci{[.101, .206]} & \minus.461 & \tci{[\minus.521, \minus.396]} & .293 & \tci{[.245, .347]} & .500 & \tci{[.461, .540]} \\
  Qwen3.5-9B & .278 & \tci{[.211, .339]} & .000 & \tci{[\minus.046, .045]} & \minus.387 & \tci{[\minus.429, \minus.344]} & .250 & \tci{[.210, .296]} & .444 & \tci{[.404, .484]} \\
  \addlinespace[2pt]
  \tgroup{11}{Comparators} \\
  MP-EG2 & \minus.996 & \tci{[\minus1.112, \minus.876]} & \minus.998 & \tci{[\minus.998, \minus.998]} & \minus.998 & \tci{[\minus.998, \minus.998]} & \minus.160 & \tci{[\minus.205, \minus.124]} & .151 & \tci{[.139, .165]} \\
  MP-GLiNER & .033 & \tci{[.015, .052]} & .002 & \tci{[\minus.003, .006]} & \minus.005 & \tci{[\minus.013, .003]} & .002 & \tci{[\minus.035, .037]} & .150 & \tci{[.126, .175]} \\
  \bottomrule
\end{tabular}
\caption{\textbf{Per-dataset Brier skill in the multi-dataset suites.} (a)~The five \coretwelve{} datasets; (b)~the datasets of the NLI and native moderation (Moder.) suites, each against its own reference.}
\label{tab:perdataset}
\end{table*}

\end{document}